\documentclass[lettersize,journal,compsoc]{IEEEtran}
\usepackage{amsmath,amsfonts}
\usepackage{algorithmic}
\usepackage{algorithm}
\usepackage{array}
\usepackage[caption=false,font=normalsize,labelfont=sf,textfont=sf]{subfig}
\usepackage{textcomp}
\usepackage{stfloats}
\usepackage{url}
\usepackage{verbatim}
\usepackage{graphicx}
\usepackage{cite}
\usepackage{array}
\usepackage{tabularx}
\usepackage{ragged2e} % \justfying

\newcommand{\bB}{\mathbf{B}}
\newcommand{\bc}{\mathbf{c}}
\newcommand{\bD}{\mathbf{D}}

\newcommand{\bg}{\mathbf{g}}\newcommand{\bG}{\mathbf{G}}

\newcommand{\bI}{\mathbf{I}}

\newcommand{\bM}{\mathbf{M}}

\newcommand{\bp}{\mathbf{p}}
\newcommand{\bq}{\mathbf{q}}\newcommand{\bQ}{\mathbf{Q}}
\newcommand{\br}{\mathbf{r}}\newcommand{\bR}{\mathbf{R}}
\newcommand{\bS}{\mathbf{S}}
\newcommand{\bt}{\mathbf{t}}\newcommand{\bT}{\mathbf{T}}

\newcommand{\bx}{\mathbf{x}}\newcommand{\bX}{\mathbf{X}}

\newcommand{\bmu}{\boldsymbol{\mu}}

\newcommand{\nR}{\mathbb{R}}

\newcommand{\figref}[1]{Fig.~\ref{#1}}

\newcommand{\eqnref}[1]{Eq.~\eqref{#1}}
\newcommand{\tabnref}[1]{Table~\ref{#1}}

\DeclareMathOperator*{\argmin}{argmin~}

\makeatletter
\DeclareRobustCommand\onedot{\futurelet\@let@token\@onedot}
\def\@onedot{\ifx\@let@token.\else.\null\fi\xspace}
\def\eg{e.g\onedot} 
\def\ie{i.e\onedot}

\makeatother

\newcommand{\PAR}[1]{\vspace{0.1cm}\noindent{\bf #1} }

\newcommand{\norm}[1]{\left\lVert#1\right\rVert}

\def\mbavo2{MBA-SLAM}
\def\ie{i.e.}
\def\eg{e.g.}

\usepackage{color}
\usepackage{booktabs}
\usepackage[table,dvipsnames,xcdraw]{xcolor} % used by tables
\usepackage{makecell}

\usepackage{multirow}
\usepackage{pifont} % for red cross
\usepackage{utfsym}

\usepackage{float}
\usepackage[export]{adjustbox}

\colorlet{colorFst}{Green!30}
\colorlet{colorSnd}{Orange!30}              % second
\colorlet{colorTrd}{Yellow!30}
\newcommand{\fs}{\cellcolor{colorFst}\bf}
\newcommand{\snd}{\cellcolor{colorSnd}}      % second

\newcommand{\add}[1]{#1}

\usepackage[pagebackref,breaklinks,colorlinks]{hyperref}

\begin{document}

\title{MBA-SLAM: Motion Blur Aware Gaussian Splatting SLAM}

\author{Peng Wang\IEEEauthorrefmark{1}, Lingzhe Zhao\IEEEauthorrefmark{1}, Yin Zhang, Shiyu Zhao, Peidong Liu\IEEEauthorrefmark{2}%~\IEEEmembership{Staff,~IEEE,}
 % <-this % stops a space

\thanks{Peng Wang is with the College of Computer Science and Technology at Zhejiang University and the School of Engineering at Westlake University, Hangzhou, China. (Email:\emph{wangpeng}@westlake.edu.cn)

All the other authors are with the School of Engineering, Westlake University, Hangzhou, Zhejiang, China. Email: ({\emph{zhaolingzhe}, \emph{zhangyin}, \emph{zhaoshiyu}, \emph{liupeidong}}@westlake.edu.cn). }
\thanks{Peng Wang and Lingzhe Zhao contributed equally.}
\thanks{Peidong Liu is the corresponding author.}
}
\markboth{Journal of \LaTeX\ Class Files,~Vol.~14, No.~8, August~2021}%
{Shell \MakeLowercase{\textit{et al.}}: A Sample Article Using IEEEtran.cls for IEEE Journals}

\IEEEpubid{0000--0000/00\$00.00~\copyright~2021 IEEE}

\IEEEtitleabstractindextext{
\begin{abstract}
\justifying
% brief intro & motivation of our work
Emerging 3D scene representations, such as Neural Radiance Fields (NeRF) and 3D Gaussian Splatting (3DGS), have demonstrated their effectiveness in Simultaneous Localization and Mapping (SLAM) for photo-realistic rendering, particularly when using high-quality video sequences as input. However, existing methods struggle with motion-blurred frames, which are common in real-world scenarios like low-light or long-exposure conditions. This often results in a significant reduction in both camera localization accuracy and map reconstruction quality.
%
% what we do & how we do
To address this challenge, we propose a dense visual deblur SLAM pipeline (i.e. \textit{\mbavo2}) to handle severe motion-blurred inputs and enhance image deblurring. Our approach integrates an efficient motion blur-aware tracker with either neural radiance fields or Gaussian Splatting based mapper. By accurately modeling the physical image formation process of motion-blurred images, our method simultaneously learns 3D scene representation and estimates the cameras' local trajectory during exposure time, enabling proactive compensation for motion blur caused by camera movement.
%
% what results do we deliver
In our experiments, we demonstrate that \textit{\mbavo2} surpasses previous state-of-the-art methods in both camera localization and map reconstruction, showcasing superior performance across a range of datasets, including synthetic and real datasets featuring sharp images as well as those affected by motion blur, highlighting the versatility and robustness of our approach. Code is available at \href{https://github.com/WU-CVGL/MBA-SLAM}{https://github.com/WU-CVGL/MBA-SLAM.}

\end{abstract}
}

\maketitle

%%%%%%%%% BODY TEXT
\section{Introduction}
\label{sec:intro}
%
%% Background 
Simultaneous Localization and Mapping (SLAM) is a fundamental problem in 3D vision with broad applications, including autonomous driving, robotic navigation, and virtual reality. While traditional sparse SLAM methods~\cite{mur2017orb, engel2017direct} use sparse point clouds for map reconstruction, recent learning-based dense SLAM systems~\cite{bloesch2018codeslam, czarnowski2020deepfactors, zhi2019scenecode, mccormac2017semanticfusion} focus on generating dense maps, which are essential for downstream applications.
%

%
%% Motivation
% motion blur images
Due to their ability to enable photo-realistic 3D scene representations, Neural Radiance Fields (NeRF)~\cite{nerf} and 3D Gaussian Splatting (3DGS)~\cite{kerbl3Dgaussians} have been explored in conjunction with SLAM systems \cite{sucar2021imap, zhu2022nice, wang2023co, johari2023eslam, rosinol2023nerf, yan2024gs, keetha2024splatam, Matsuki:Murai:etal:CVPR2024, hhuang2024photoslam}, demonstrating significant improvements in map representation and high-fidelity surface reconstruction.
However, existing methods heavily rely on high-quality, sharp RGB-D inputs, which poses challenges when dealing with motion-blurred frames, often encountered in low-light or long-exposure conditions. Such conditions can significantly degrade the localization and mapping performance of these methods. The difficulties that motion-blurred images present to dense visual SLAM systems stem from two primary factors: \textbf{1) inaccurate pose estimation during tracking:} current photo-realistic dense visual SLAM algorithms depend on sharp images to estimate camera poses by maximizing photometric consistency. However, motion-blurred images, commonly occurring in real-world scenarios, violate this assumption, making it difficult to accurately recover poses from blurred frames. These inaccurately tracked poses, in turn, affect the mapping process, leading to inconsistent multi-view geometry. \textbf{2) inconsistent multi-view geometry in mapping:} the mismatched features between multi-view blurry images introduce erroneous 3D geometry information, resulting in poor 3D map reconstruction. This will degrade map reconstruction quality, which subsequently affects the tracking process. Combined these two factors, existing dense virtual SLAM systems would usually exhibit performance degradation when handling motion-blurred images.

\IEEEpubidadjcol

%% What we do
To address these challenges, we introduce \textit{\mbavo2}, a photo-realistic dense RGB-D SLAM pipeline designed to handle motion-blurred inputs effectively. Our approach integrates the physical motion blur imaging process into both the tracking and mapping stages. Specifically, we employ a continuous motion model within the $\mathbf{SE}(3)$ space to characterize the camera motion trajectory within exposure time. 
Given the typically short exposure duration, the trajectory of each motion-blurred image is represented by its initial and final poses at the start and end of the exposure time respectively.  
% 
% In the tracking stage of \textit{\mbavo2}, we estimate the local camera motion trajectory within the exposure time for each frame. Subsequently, in the mapping stage, we optimize the local camera motion trajectories of selected keyframes jointly with the 3D scene representation.
% %
% 
During tracking, we firstly render a reference sharp image corresponding to the latest keyframe, from our learned 3D scene representation. The rendered image can then be reblurred to match the current captured blurry image based on the predicted motion trajectory from previous optimization iteration. We enforce the photo-metric consistency between the tracked blurry image and the reblurred image to further refine the camera motion trajectory within exposure time.
In the mapping stage, we jointly optimize the trajectories of a set of sparsely selected frames (i.e. keyframes) and the 3D scene representation by minimizing the photo-metric consistency loss. Two commonly used scene representations are explored in our implementation, i.e. implicit neural radiance fields ~\cite{johari2023eslam} and explicit 3D Gaussian Splatting ~\cite{kerbl3Dgaussians}. Both representations exhibit different advantages and disadvantages. In particular, NeRF-based implementation is able to achieve higher frame rates (FPS) but exhibits lower rendering quality than 3D-GS based implementation. In contrary, 3D-GS based implementation delivers better rendering quality at the expense of lower FPS. We present both implementations to satisfy the requirements of different usage scenarios.

% \begin{figure}
% 	\centering
% 	\setlength\tabcolsep{0pt}
% 	\begin{tabular}{lr}
% 		\includegraphics[width=0.5\columnwidth]{figs/gfx/ref_image2.jpg}   	\includegraphics[width=0.5\columnwidth]{figs/gfx/tracked_image2.jpg}
% 	\end{tabular}
	
% 	\caption{\textbf{Motion Blur Aware Visual Odometry}. \textbf{Left:} rendered virtual sharp keyframe image with detected feature points; \textbf{Right:} current input blurred image with tracked trajectory during exposure time. By explicitly modelling the image formation process during tracking and mapping, we can actively compensate for motion blur in the direct image alignment.}
% 	\label{fig:tracker}
% 	\vspace{-1.5em}
% \end{figure}

%% What we achieve 
% 1: photometric bundle adjustment 
% 2: achieve superior deblur performance (comparison with a multi-view deblur method)

We evaluate the performance of \textit{\mbavo2} thoroughly by using both the sharp and blurry datasets, against prior state-of-the-art methods. In particularly, we conducted evaluations with both a public synthetic blurry dataset \cite{mba-vo} and a self-captured blurry dataset. The real dataset is collected with a RealSense RGB-D camera under low-lighting conditions. To further evaluate the performance of \textit{\mbavo2} on sharp images, we exploit the commonly used public datasets from Replica~\cite{replica19arxiv}, ScanNet~\cite{dai2017scannet} and TUM RGB-D~\cite{sturm2012benchmark}. The experimental results demonstrate that \textit{\mbavo2} not only delivers more robust performance with blurry images, but also has superior performance with sharp images, than prior state-of-the-art methods. 

%
%% Our main contributions 
In summary, our {\bf{contributions}} are as follows:
%\vspace{-0.3em}
\begin{itemize}
	\item We present a novel photometric bundle adjustment formulation specifically designed for motion blurred images, establishing an RGB-D 3DGS/NeRF-based SLAM pipeline that demonstrates robustness against motion blur.
	\item Our SLAM pipeline is enhanced by integrating a motion blur-aware tracker, resulting in improved tracking accuracy, which in turn leads to superior mapping performance.
	\item We illustrate how this formulation enables the acquisition of precise camera trajectories and high-fidelity 3D scene maps from motion-blurred inputs.
	\item Our experimental results demonstrate the superior tracking and mapping performance of \mbavo2 across various datasets, outperforming previous state-of-the-art NeRF-based and 3DGS-based SLAM methods, including both synthetic and real motion blurred datasets.
        \item Our method also performs well and surpasses previous state-of-the-art dense visual SLAM pipelines on commonly used standard  datasets with sharp images.
\end{itemize}

%%% previous work
\textit{\mbavo2} is based on three preliminary seminar papers of the authors, \ie, MBA-VO~\cite{mba-vo}, BAD-NeRF~\cite{wang2023badnerf}, and BAD-Gaussians~\cite{zhao2024bad}, which have been accepted by ICCV 2021 (oral), CVPR 2023, and ECCV 2024, respectively. 
In this paper, we extend these works in several significant ways: 
1) we integrate them into a comprehensive SLAM pipeline, by exploiting the motion blur aware tracker from MBA-VO~\cite{mba-vo} and the motion blur aware bundle adjustment algorithm from either BAD-NeRF~\cite{wang2023badnerf} or BAD-Gaussians~\cite{zhao2024bad};
2) we replace the vanilla NeRF representation of BAD-NeRF~\cite{wang2023badnerf} with a more efficient tri-plane based representation, significantly improving the training efficiency by a factor of 100 times;
3) all the experimental evaluations are newly conducted to thoroughly verify the effectiveness of the pipeline, against prior state-of-the-art methods.
% 
% 4) our pipeline achieves superior performance without relying on a pre-trained deblurring network such as~\cite{Tao2018CVPR, kupyn2019deblurgan}, which was necessary for MBA-VO~\cite{mba-vo}. It also does not require any pose initialization from COLMAP~\cite{colmap}, which was essential for BAD-NeRF~\cite{wang2023badnerf} and BAD-Gaussian~\cite{zhao2024bad}.
% 
% 
% Thorough experimental evaluations have been conducted to verify the effectiveness of the extended system. Second, our SLAM system achieves commendable performance without relying on pre-trained deblurring networks such as~\cite{Tao2018CVPR, kupyn2019deblurgan}, which were necessary for MBA-VO~\cite{mba-vo}, and without requiring pose initialization from COLMAP~\cite{colmap}, which was essential for BAD-NeRF~\cite{wang2023badnerf} and BAD-GS~\cite{zhao2024bad}. Third, we enhance the NeRF framework introduced in BAD-NeRF~\cite{wang2023badnerf}, significantly improving training efficiency by a factor of 100 times. 
% The previous codes are avaliable at \href{https://github.com/ethliup/MBA-VO}{MBA-VO}, \href{https://github.com/WU-CVGL/BAD-NeRF}{BAD-NeRF} and \href{https://github.com/WU-CVGL/BAD-Gaussians}{BAD-GS}, which have received over 400 stars combined.

\section{Related Work}
\label{sec:related}
In the subsequent section, our attention will be primarily directed towards a comprehensive review of methodologies closely aligned with our work.
% \subsection{3D Scene Reconstruction}
% NeRF \cite{nerf}, which leverages implicit MLPs, demonstrates exceptional proficiency in generating high-fidelity novel view images and accurately representing 3D scenes. A plethora of extension studies have emerged to enhance NeRF's capabilities, including improvements in training and rendering efficiency \cite{muller2022instant, Chen2022ECCV} and the well-known 3DGS~\cite{kerbl3Dgaussians}. Furthermore, some approaches aim to enhance the robustness of NeRF and 3DGS against imperfect inputs, such as inaccurate camera poses~\cite{Lin2021, wang2021nerfmm, Fu_2024_CVPR} and low-quality captured images~\cite{mildenhall2022rawnerf, deblur-nerf, li2023usb, peng2024bags}.

\subsection{Radiance Fields based Photo-realistic Dense Visual SLAM.}
Numerous approaches have emerged aiming to integrate NeRF into SLAM frameworks \cite{sucar2021imap, zhu2022nice, wang2023co, johari2023eslam, zhang2023go, chung2023orbeez, rosinol2023nerf} to achieve dense map reconstruction. The pioneering work iMAP \cite{sucar2021imap} initially proposed employing a single MLP to represent the entire scene, while NICE-SLAM \cite{zhu2022nice} extended iMAP \cite{sucar2021imap} by encoding the scene with hierarchical, grid-based features and decoding them using pre-trained MLPs. Subsequent works such as Point-SLAM \cite{sandstrom2023point}, CoSLAM \cite{wang2023co}, and ESLAM \cite{johari2023eslam} have further enhanced representation accuracy and efficiency by incorporating neural points, hash grids, and feature planes, respectively. Additionally, approaches like Orbeez-SLAM \cite{chung2023orbeez} and GO-SLAM \cite{zhang2023go} amalgamate the tracking components from ORB-SLAM2 \cite{mur2017orb} and DROID-SLAM \cite{teed2021droid} to mitigate tracking drift errors.
With the emergence of 3DGS~\cite{kerbl3Dgaussians}, several works have deployed the 3DGS map representation for SLAM ~\cite{yan2024gs,keetha2024splatam, Matsuki:Murai:etal:CVPR2024, hhuang2024photoslam}, SplaTAM~\cite{keetha2024splatam}.
While these methods excel in processing high-quality, sharp input data, their performance is significantly compromised when dealing with imperfect inputs such as motion blur. Additionally, visual odometries, like those used in Orbeez-SLAM~\cite{chung2023orbeez} and GO-SLAM~\cite{zhang2023go}, often struggle to track frames affected by motion blur. In contrast, we leverage the motion blur-aware visual odometry introduced in our preliminary work~\cite{mba-vo}, which estimates local motion trajectories within the exposure time rather than relying on instantaneous poses. This adaptation significantly enhances the robustness of our full SLAM system to motion blur.  

\subsection{NeRF/3DGS for Deblurring.}
Several scene deblurring methods based on NeRF~\cite{nerf} and 3DGS~\cite{kerbl3Dgaussians} have emerged, such as Deblur-NeRF~\cite{deblur-nerf}, DP-NeRF~\cite{Lee_2023_CVPR}, Deblur-GS~\cite{lee2024deblurring} and BAGS~\cite{peng2024bags}. These methods aim to reconstruct sharp scene representations from sets of motion-blurred images by estimating the blur kernel while fixing the inaccurate camera poses recovered from blurred images during training NeRF and 3DGS. BAD-NeRF~\cite{wang2023badnerf} and its extension, ExBlurF~\cite{lee2023exblurf} and BAD-Gaussians~\cite{zhao2024bad}, employ joint learning of camera motion trajectories within exposure time and radiance fields, adhering to the physical blur process. Despite these advancements, the accurate initialization of camera poses from COLMAP~\cite{colmap} remains a prerequisite, and challenges pertaining to low convergence efficiency persist, hindering their seamless integration into SLAM systems.

\subsection{Classic SLAM Algorithms.}
Visual odometry is the process of determining the relative motion of a camera based on captured images. Direct methods, such as LSD-SLAM~\cite{engel2014lsd} and DSO~\cite{engel2017direct}, along with their numerous variants~\cite{gao2018ldso, liu2018towards, liu2017direct, schubert2019rolling}, optimize the camera pose simultaneously with the 3D scene by minimizing the photometric loss across multi-view images. In contrast, feature-based visual odometries~\cite{davison2003real, mur2017orb, nister2004visual} estimate pose and scene structure by enforcing consistency between the locations of keypoints extracted from raw images and their corresponding projections.
While some recent learning-based approaches~\cite{Ummenhofer2017CVPR, zhou2018deeptam, Zhou2017CVPR} treat the joint optimization of poses and scenes as an end-to-end regression problem, they remain relatively nascent compared to the more established direct and feature-based geometric methods in terms of scalability and performance. Nonetheless, the performance of these visual odometry systems can be significantly compromised by motion-blurred images, as they typically assume that the captured images are sharp. Previous work~\cite{lee2011simultaneous} attempted to address motion blur in visual odometry by assuming smooth motion between neighboring frames and trying to linearly interpolate the motion within the exposure time. However, this method heavily relies on initial motion predictions to make critical decisions regarding correspondences. In contrast, MBA-VO~\cite{mba-vo} adopts a different approach by directly optimizing the local camera motion trajectory to re-blur patches, implicitly solving the data association problem using a direct image alignment strategy. 
More thorough reviews on classic VO/SLAM algorithms can refer to \cite{Cadena2016TOR}.

% Nevertheless, MBA-VO~\cite{mba-vo} requires pre-trained deblurring models \cite{Tao2018CVPR} to recover sharp reference keyframes from motion-blurred frames.
%
% In contrast, our SLAM system is capable of directly rendering virtual sharp reference keyframes from neural scene representations without the need for additional models.
%
\begin{figure*}[t]
	\centering
	\includegraphics[width=0.95\textwidth]{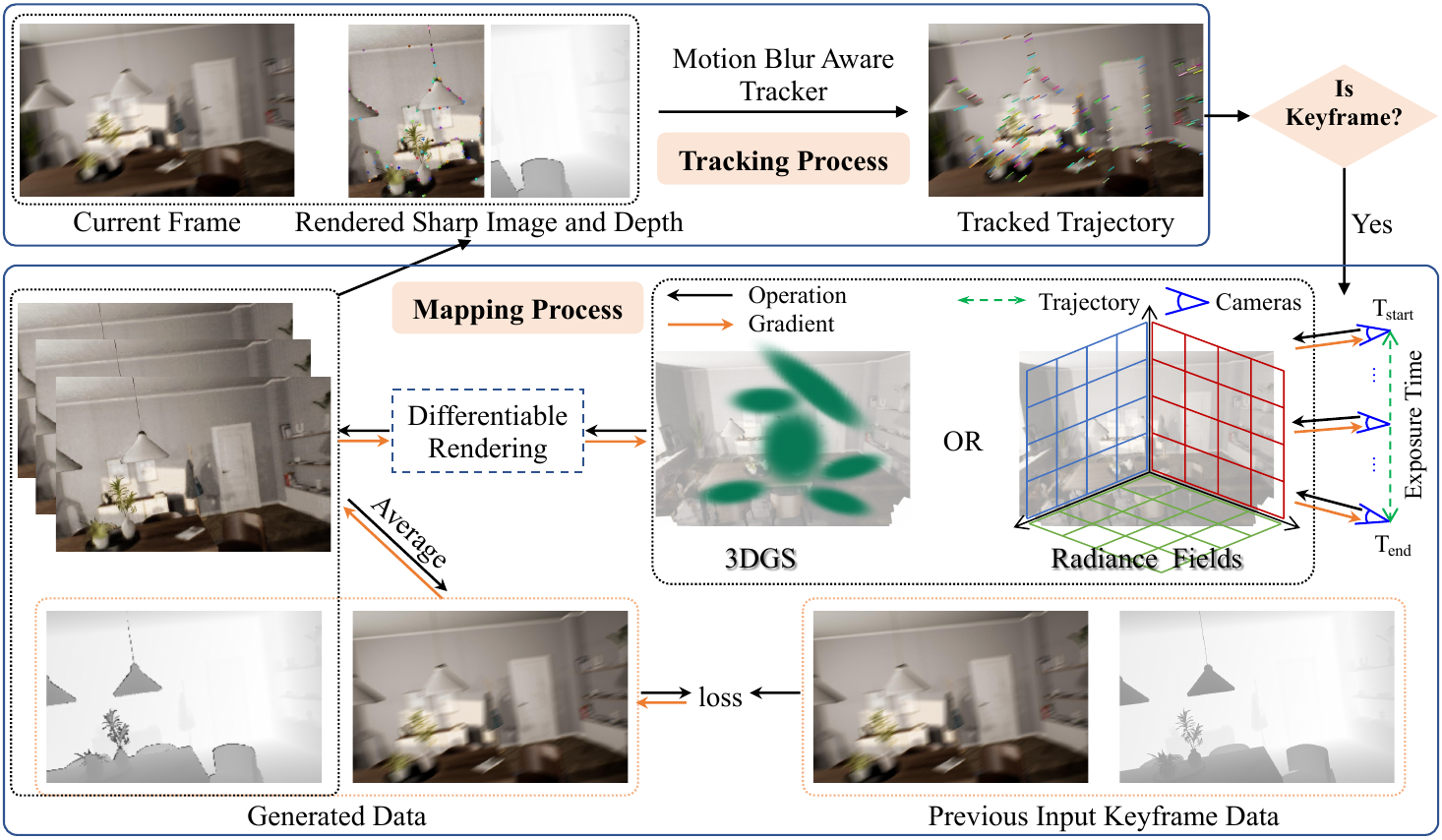}
        \vspace{-1ex}
	\caption{\textbf{The pipeline of \mbavo2.} Our framework consists of blur aware tracking process and bundle adjustment deblurring mapping process. \textbf{\textit{Tracking:}} Given the current blurry frame, the mapper first renders a virtual sharp image of the lastest blurry keyframe from the 3D scene. Our motion blur-aware tracker directly estimates the camera motion trajectory during the exposure time, represented by the camera positions at the start and end of the exposure (\ie $\bT_\mathrm{start}$ and $\bT_\mathrm{end}$). Intermediate camera poses can be interpolated in $\mathbf{SE}(3)$ space. \textbf{\textit{Mapping:}} Our mapper generates virtual sharp images along the camera trajectory, following the standard rendering procedures of Radiance Fields or Gaussian Splatting. The blurry image can then be synthesized by averaging these virtual images, adhering to the physical image formation model of motion-blurred images. Finally, the scene representation and camera trajectory are jointly optimized by minimizing the loss between the synthesized images and the input data.}
    \label{fig_pipeline}
    \vspace{-1.0em}
\end{figure*}
\section{Method}\label{sec:method}
In this section, we detail our approach, Motion Blur-Aware Dense Visual SLAM (\mbavo2), designed to process streams of motion-blurred RGB images and corresponding depth data. The primary objective of \mbavo2 is to reconstruct high-quality dense 3D scenes while accurately estimating camera motion trajectories. This is achieved by integrating two key components: a motion blur-aware tracker and a bundle-adjusted deblur mapper based on either NeRF~\cite{nerf} or 3D Gaussian Splatting~\cite{kerbl3Dgaussians}.

The front-end tracker estimates the local camera motion trajectory within the exposure time of the current blurry frame, relative to the latest virtual sharp keyframe image rendered from the learned 3D scene representations. The back-end mapper operates by jointly learning the implicit or explicit scene representation and estimating camera trajectories. Sec.~\ref{sec:blur_formation_model} introduces the physical formation process of motion blurry images, while Sec.~\ref{sec:mba_vo} elaborates on how the tracker estimates camera motion trajectories from motion-blurred images through direct image alignment. In Sec.~\ref{sec:bad_mapping}, the camera trajectories and scene representation are estimated by maximizing the photometric consistency between synthesized and real captured blurry images. Each component will be detailed in the following sections.

\vspace{-1ex}
\subsection{Motion Blur Image Formation Model}\label{sec:blur_formation_model}
The process of physical motion blur image formation encompasses the digital camera's acquisition of photons over the duration of exposure, converting them into quantifiable electric charges. Mathematically, this intricate process necessitates the integration across a sequence of virtual sharp images: 
\begin{equation}
	\bB(\bx) = \phi \int_{0}^{\tau} \bI_\mathrm{t}(\bx) \mathrm{dt},
\end{equation}
where $\bB(\bx) \in \nR^{\mathrm{W} \times \mathrm{H} \times 3}$ denotes the captured image, $\mathrm{W}$ and $\mathrm{H}$ represent its width and height, $\bx \in \nR^2$ is the pixel location, $\phi$ serves as a normalization factor, $\tau$ is the camera exposure time, $\bI_\mathrm{t}(\bx) \in \nR^{\mathrm{W} \times \mathrm{H} \times 3}$ is the virtual sharp image captured at timestamp $\mathrm{t}$ within the exposure time. The blurred image $\bB(\bx)$ caused by camera motion within the exposure time, is a composite of virtual images $\bI_\mathrm{t}(\bx)$ at each $t$. The  discrete approximation of this model is described as follows: 
\begin{equation}\label{eq_blur_im_formation}
	\bB(\bx)  \approx \frac{1}{n} \sum_{i=0}^{n-1} \bI_\mathrm{i}(\bx), 
   \end{equation}
where $n$ is the number of discrete samples. 

The level of motion blur in an image is directly influenced by the motion of the camera during the exposure duration. For instance, a camera moving swiftly would entail minimal relative motion, particularly with shorter exposure times, whereas a slow-moving camera could result in motion-blurred images, especially noticeable in low-light scenarios with long exposure time. Consequently, it can be inferred that $\bB(\bx)$ exhibits differentiability with respect to each virtual sharp image $\bI_i(\bx)$.

\subsection{Motion Blur Aware Tracker}\label{sec:mba_vo}
\subsubsection{Direct Image Alignment with Sharp Images}\label{sec:mba_vo_sharp_alignment}
Prior to presenting our direct image alignment algorithm designed for blurry images, we provide an overview of the original algorithm tailored for sharp images. This direct image alignment algorithm forms the fundamental component of direct visual odometry approaches.
It estimates the camera pose of current tracked frame by minimizing the photometric error between the latest keyframe and current frame.
This process is formally defined as follows:
\begin{equation}\label{eq_photoconsistency}
\bT^{*} = \argmin_{\bT} \sum_{i=0}^{m-1} \norm{\bI_{\mathrm{ref}}(\bx_i) - \bI_{\mathrm{cur}}(\hat{\bx}_i)}_2^2,
\end{equation}
where $\mathrm{\bT} \in \mathbf{SE}(3)$ is the transformation matrix from the reference image $\bI_{\mathrm{ref}}$ to the current image $\bI_{\mathrm{cur}}$, $m$ denotes the number of sampled pixels used for motion estimation, $\mathrm{\bx}_i \in \nR^2$ is the location of the $i^{th}$ pixel, $\hat{\bx}_i \in \nR^2$ is the pixel location corresponding to pixel $\mathrm{\bx}_i$ in current image $\bI_{\mathrm{cur}}$. Robust loss function (\eg~huber loss) is typically also applied to the error residuals to ensure robust pose estimation. The image points $\mathrm{\bx}_i$ and $\hat{\bx}_i$ are interconnected by the camera pose $\mathrm{\bT}$ and the depth $d_i$ as
\begin{equation} \label{eq_transfer}
\hat{\bx}_i = \mathrm{\pi}(\bT \cdot \pi^{-1}(\bx_i, d_i)),
\end{equation}
where $\pi: \nR^3 \rightarrow \nR^2$ is the camera projection function, responsible for projecting points in 3D space onto the image plane; conversely, $\pi^{-1}: \nR^2 \times \nR \rightarrow \nR^3$ is the inverse projection function, which facilitates the transformation of a 2D point from the image plane back into 3D space by incorporating the corresponding depth $d_i$. This formulation can be extended seamlessly to multi-frame scenarios, enabling the joint optimization of camera poses, 3D scene structures, and camera intrinsic parameters (\ie~commonly referred to as photometric bundle adjustment). 

Direct VO methods assume photoconsistency (\ie~\eqnref{eq_photoconsistency}) for the correct transformation $\bT$. However, varying motion blur in images $\bI_{\mathrm{ref}}$ and $\bI_{\mathrm{cur}}$ invalidates the photoconsistency loss, as local appearance differs for corresponding points. This is common in settings with non-linear trajectories, like augmented/mixed/virtual reality applications, leading to varying levels of motion blur in images.

\subsubsection{Camera Motion Trajectory modeling}\label{sec:mba_vo_trajectory}
Accurate compensation for motion blur necessitates modeling the local camera trajectory over the exposure period. One strategy involves parameterizing solely the final camera pose and then linearly interpolating between the previous frame and the new estimate. This interpolation enables the creation of virtual images to depict the motion blur, as outlined in~\eqnref{eq_blur_im_formation}. However, this method might fail when faced with camera trajectories exhibiting sudden directional changes, a common occurrence with hand-held and head-mounted cameras.

For robustness, we opt to parameterize the local camera trajectory independently of the previous frame.
Specifically, we parameterize two camera poses: one at the beginning of the exposure $\bT_\mathrm{start} \in \mathbf{SE}(3)$ and another at the end $\bT_\mathrm{end} \in \mathbf{SE}(3)$. Between the two poses we employ linear interpolation of poses in the Lie-algebra of $\mathbf{SE}(3)$. Thus, The virtual camera pose at time $t \in [0,\tau]$ is represented as 
\begin{equation} \label{eq_trajectory}
\bT_t = \bT_\mathrm{start} \cdot \exp(\frac{t}{\tau} \cdot \log(\bT_\mathrm{start}^{-1} \cdot \bT_\mathrm{end})),
\end{equation} 
where $\tau$ is the exposure time. For efficiency, we decompose \eqnref{eq_trajectory} as 
\begin{align}
\label{eq_mtraj_R}
\bar{\bq}_t &= \bar{\bq}_\mathrm{start} \otimes \exp(\frac{t}{\tau} \cdot \log((\bar{\bq}_\mathrm{start})^{-1} \otimes \bar{\bq}_\mathrm{end})), \\
\label{eq_mtraj_t}
\bt_t &= \bt_\mathrm{start} + \frac{t}{\tau} (\bt_\mathrm{end} - \bt_\mathrm{start}),
\end{align}
where $\otimes$ is the product operator for quaternions, $\bT_* = [\bR_* | \bt_*] \in \mathbf{SE}(3)$, $\bR_* \in \mathbf{SO}(3)$ and $\bt_* \in \nR^3$. We represent the rotation matrix $\bR_*$ with unit quaternion $\bar{\bq}_* = \begin{bmatrix} {q_x}_* & {q_y}_* & {q_z}_* & {q_w}_* \end{bmatrix}^T$.

Our motion blur-aware tracker aims to estimate both $\bT_\mathrm{start}$ and $\bT_\mathrm{end}$ for each frame. While our current approach employs linear interpolation between the two poses, higher-order splines could potentially capture more intricate camera motions. Nevertheless, our experiments revealed that the linear model worked well enough, since the exposure time is usually relatively short.

Here are some more detailed explanations regarding the interpolation and derivations of the related Jacobian.

\textit{Local parameterization of rotation}: For the real implementation, we use the local parameterization for the update of the rotation. The plus operation for unit quaternion $\bar{\bq}$ is defined as
\begin{align}
	\bar{\bq}^\prime = \bar{\bq} \otimes \Delta \bar{\bq},
\end{align}
where $\Delta \bar{\bq} = \exp(\Delta \br)$, $\Delta \br = \begin{bmatrix}
	\Delta r_x & \Delta r_y & \Delta r_z
\end{bmatrix}^T$ and $\Delta r_x \rightarrow 0$,  $\Delta r_y \rightarrow 0$,  $\Delta r_z \rightarrow 0$. 
The Jacobian with respect to $\Delta \br$ can thus be derived as\footnote{$\bQ(\bar{\bq})$ and $\hat{\bQ}(\bar{\bq})$ are the matrix forms of quaternion multiplication.}
\begin{align}
	\frac{\partial \bar{\bq}^\prime}{\partial \Delta \br} = \bQ(\bar{\bq}) \cdot \begin{bmatrix}
		0.5 & 0 & 0 \\
		0 & 0.5 & 0 \\
		0 & 0 & 0.5 \\
		0 & 0 & 0
	\end{bmatrix}.
\end{align}

\textit{Jacobian related to translation}: We can simplify \eqnref{eq_mtraj_t} as 
\begin{align}
	\bt_t &= \frac{\tau-t}{\tau}\bt_\mathrm{start} + \frac{t}{\tau} \bt_\mathrm{end}.
\end{align}
The Jacobians, \ie $\frac{\partial \bt_t}{\partial \bt_\mathrm{start}} \in \nR^{3\times3}$ and $\frac{\partial \bt_t}{\partial \bt_\mathrm{end}} \in \nR^{3\times3}$ can thus be derived as 
\begin{align}
	\frac{\partial \bt_t}{\partial \bt_\mathrm{start}} &= \begin{bmatrix}
		\frac{\tau-t}{\tau} & 0 & 0 \\
		0 & \frac{\tau-t}{\tau} & 0 \\
		0 & 0 & \frac{\tau-t}{\tau}
	\end{bmatrix},
\end{align}
\begin{align}
	\frac{\partial \bt_t}{\partial \bt_\mathrm{end}} &= \begin{bmatrix}
		\frac{t}{\tau} & 0 & 0 \\
		0 & \frac{t}{\tau} & 0 \\
		0 & 0 & \frac{t}{\tau}
	\end{bmatrix}. 
\end{align}

\textit{Jacobian related to rotation}: We decompose \eqnref{eq_mtraj_R} as 
\begin{align}
	\label{eq_mtraj_0}
	\bar{\bq}_\mathrm{end}^\mathrm{start} &= (\bar{\bq}_\mathrm{start})^{-1} \otimes \bar{\bq}_\mathrm{end}, \\
	\br &= \frac{t}{\tau} \cdot \log(\bar{\bq}_\mathrm{end}^\mathrm{start}), \\
	\bar{\bq}_t^\mathrm{start} &= \exp(\br), \\
	\label{eq_mtraj_3}
	\bar{\bq}_t &= \bar{\bq}_\mathrm{start} \otimes \bar{\bq}_t^\mathrm{start}.
\end{align}
We can rewrite both \eqnref{eq_mtraj_0} and \eqnref{eq_mtraj_3} as 
\begin{align}
	\bar{\bq}_\mathrm{end}^\mathrm{start} &= \bQ((\bar{\bq}_\mathrm{start})^{-1}) \cdot \bar{\bq}_\mathrm{end} = \hat{\bQ}(\bar{\bq}_\mathrm{end}) \cdot (\bar{\bq}_\mathrm{start})^{-1}, \\
	\bar{\bq}_t &= \bQ(\bar{\bq}_\mathrm{start}) \cdot \bar{\bq}_t^\mathrm{start} = \hat{\bQ}(\bar{\bq}_t^\mathrm{start}) \cdot \bar{\bq}_\mathrm{start}.
\end{align}
The Jacobian $\frac{\partial \bar{\bq}_t}{\partial \bar{\bq}_\mathrm{start}} \in \nR^{4\times4}$ can thus be derived as
\begin{align}
	\frac{\partial \bar{\bq}_t}{\partial \bar{\bq}_\mathrm{start}} &= \hat{\bQ}(\bar{\bq}_t^\mathrm{start}) +  \bQ(\bar{\bq}_\mathrm{start}) \cdot \frac{\partial \bar{\bq}_t^\mathrm{start}}{\partial \bar{\bq}_\mathrm{start}}, \\
	\frac{\partial \bar{\bq}_t^\mathrm{start}}{\partial \bar{\bq}_\mathrm{start}} &= \frac{\partial \bar{\bq}_t^\mathrm{start}}{\partial \br} \cdot \frac{\partial \br}{\partial \bar{\bq}_\mathrm{end}^\mathrm{start}} \cdot \hat{\bQ}(\bar{\bq}_\mathrm{end}) \cdot \frac{\partial (\bar{\bq}_\mathrm{start})^{-1}}{\partial \bar{\bq}_\mathrm{start}}.
\end{align}
Similarly for the Jacobian $\frac{\partial \bar{\bq}_t}{\partial \bar{\bq}_\mathrm{end}} \in \nR^{4\times4}$, we can derive it as 
\begin{align}
	\frac{\partial \bar{\bq}_t}{\partial \bar{\bq}_\mathrm{end}} = \bQ(\bar{\bq}_\mathrm{start}) \cdot \frac{\partial \bar{\bq}_t^\mathrm{start}}{\partial \br} \cdot \frac{\partial \br}{\partial \bar{\bq}_\mathrm{end}^\mathrm{start}} \cdot \bQ((\bar{\bq}_\mathrm{start})^{-1}).
\end{align}
Note that both $\frac{\partial \bar{\bq}_t^0}{\partial \br}$ and $\frac{\partial \br}{\partial \bar{\bq}_\tau^0}$ are the Jacobians related to the exponential mapping and logarithm mapping respectively. $\frac{\partial (\bar{\bq}_\mathrm{start})^{-1}}{\partial \bar{\bq}_\mathrm{start}}$ is the Jacobian related to the inverse of quaternion.

The Jacobians with respect to the local parameterization can then be computed as 
\begin{align}
	\frac{\partial \bar{\bq}_t}{\partial \Delta \br_\mathrm{start}} &= \frac{\partial \bar{\bq}_t}{\partial \bar{\bq}_\mathrm{start}} \cdot \bQ(\bar{\bq}_\mathrm{start}) \cdot \begin{bmatrix}
		0.5 & 0 & 0 \\
		0 & 0.5 & 0 \\
		0 & 0 & 0.5 \\
		0 & 0 & 0
	\end{bmatrix},
\end{align}

\begin{align}
	\frac{\partial \bar{\bq}_t}{\partial \Delta \br_\mathrm{end}} &= \frac{\partial \bar{\bq}_t}{\partial \bar{\bq}_\mathrm{end}} \cdot \bQ(\bar{\bq}_\mathrm{end}) \cdot \begin{bmatrix}
		0.5 & 0 & 0 \\
		0 & 0.5 & 0 \\
		0 & 0 & 0.5 \\
		0 & 0 & 0
	\end{bmatrix}.
\end{align}

\subsubsection{Direct Image Alignment with Blurry Images}\label{sec:mba_vo_blur_alignment}
Our motion blur-aware tracker operates by directly aligning the keyframe, assumed to be sharp, with the current frame, which may suffer from motion blur. To exploit photometric consistency during alignment, we must either de-blur the current frame or re-blur the keyframe. In our approach, we opt for the latter option as re-blurring is generally simpler and more robust compared to motion deblurring, particularly for images severely affected by motion blur. 

\begin{figure}
	\centering
	\begin{tabular}{cc}
		\includegraphics[width=0.45\columnwidth]{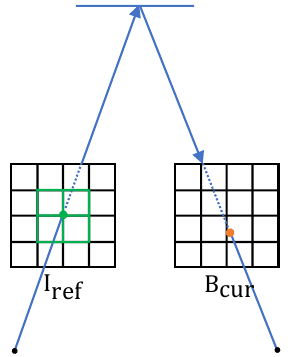} &
		\includegraphics[width=0.45\columnwidth]{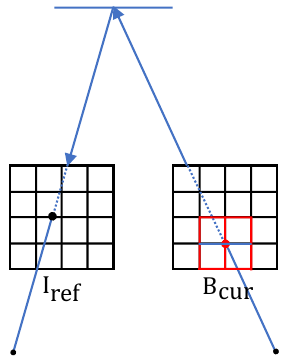} \\
	\end{tabular}
	\caption{\textbf{Pixel point transfer strategies.} Note that we assume the pixel center lies at the grid intersection, \eg the green grid is considered as a 3$\times$3 patch.}
	\label{fig_image_warping}
	% \vspace{-1em}
\end{figure}

Each sampled pixel in $\bI_\mathrm{ref}$ with a known depth is transfered to the current (blurry) image $\bB_\mathrm{cur}$ using~\eqref{eq_transfer}. We then identify the nearest integer position pixel in the current blurry image for each projected point. Assuming the 3D point lies on a fronto-parallel plane (with respect to $\bI_\mathrm{ref}$), we leverage this plane to transfer the selected pixel back into the reference view. Further details are provided in \figref{fig_image_warping}. To synthesize the re-blurred pixel from the reference view (facilitating comparison with the real captured pixel intensity), we interpolate between $\bT_\mathrm{start}$ and $\bT_\mathrm{end}$. For each virtual view $\bT_t$ sampled uniformly within $[0,\tau]$, the pixel coordinate (\ie the red pixel in \figref{fig_image_warping}) is transferred back into the reference image and image intensity values are obtained by bi-linear interpolation. 
The re-blurred pixel intensity is then computed by averaging over the intensity values (as in~\eqref{eq_blur_im_formation}):
\begin{equation}\label{eq_point_transfer}
\hat{\bB}_\mathrm{cur}(\bx) = \frac{1}{n}\sum_{i=0}^{n-1} \bI_\mathrm{ref}(\bx_{\frac{i\tau}{n-1}}),
\end{equation}
where $\bx_{\frac{i\tau}{n-1}} \in \nR^2$ corresponds to the transferred point at time $t = \frac{i\tau}{n-1}$ in the sharp reference frame, $n$ is the number of virtual frames used to synthesize the blurry image\footnote{In our experiments, we employ a fixed number of virtual frames. Nonetheless, this quantity can be dynamically adjusted based on the blur level to optimize computational resources.}.
The tracker then optimizes over the $\bT_\mathrm{start}$ and $\bT_\mathrm{end}$ to minimize the photoconsistency loss between the real captured intensities in current frame and the synthesized pixel intensities from the reference image via re-blurring:
\begin{equation}\label{eq_photoconsistency2}
% \small
{\bT_\mathrm{start}^*,~\bT_\mathrm{end}^*} = \argmin_{\bT_\mathrm{start},~\bT_\mathrm{end}}  \sum_{i=0}^{m-1} \norm{\bB_\mathrm{cur}(\bx_i) - \hat{\bB}_{\mathrm{cur}}(\bx_i)}.
\end{equation}

In practice, many direct image alignment methods use local patches to facilitate convergence. In contrast to direct image alignment algorithm for sharp images, which typically selects the local patch from the reference image (\eg the green $3\times3$ grid on the left of \figref{fig_image_warping}), we instead select the local patch from the current blurry image (\eg the red $3\times3$ grid on the right of \figref{fig_image_warping}) since this simplifies the re-blurring step of our pipeline.

\begin{figure}
	\centering
	\includegraphics[width=0.45\columnwidth]{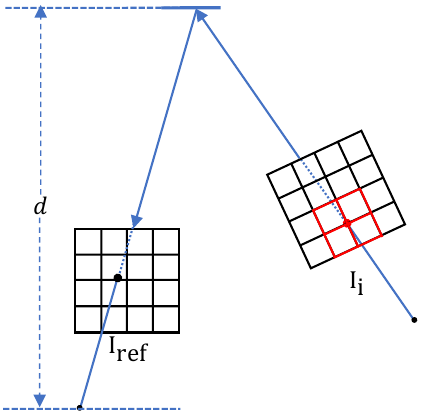}
	\includegraphics[width=0.5\columnwidth]{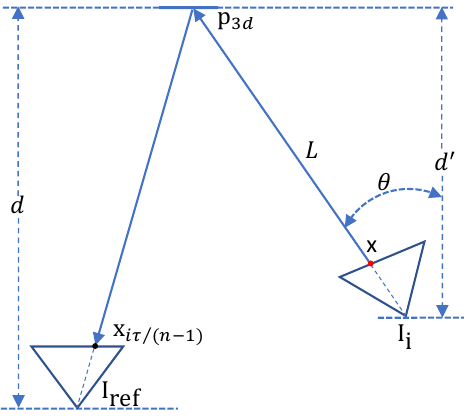}
	\caption{\textbf{Geometric relationship} between $\bx \in \nR^2$ (\ie the red pixel) of the virtual sharp image $\bI_i$ and $\bx_{\frac{i\tau}{n-1}} \in \nR^2$ (\ie the black pixel) of the reference image $\bI_\mathrm{ref}$. The right figure is its simplified 2D top-down view of the left figure.}
	\label{fig_geom_rela}
	% \vspace{-1em}
\end{figure}

\subsubsection{More Details on the Transfer}\label{sec:mba_vo_details} 
We further demonstrate the relationship between $\bx \in \nR^2$ and $\bx_{\frac{i\tau}{n-1}} \in \nR^2$ from \eqnref{eq_point_transfer} in the following notations. We denote the depth of the fronto-parallel plane as $d$, representing the estimated depth of the corresponding sampled keypoint from $\bI_\mathrm{ref}$ (\ie the green pixel in \figref{fig_image_warping}); additionally, we denote the camera pose of the virtual frame $\bI_i$ captured at timestamp $\frac{i\tau}{n-1}$ relative to the reference keyframe $\bI_{\mathrm{ref}}$ as $\bT_i \in \mathbf{SE}(3)$, which can be computed from \eqnref{eq_trajectory} as 
\begin{equation}
\bT_i = \bT_\mathrm{start} \cdot \exp(\frac{i}{n-1}\tau \cdot \log(\bT_\mathrm{start}^{-1} \cdot \bT_\mathrm{end})),
\end{equation}
where $\bT_\mathrm{start} \in \mathbf{SE}(3)$ and $\bT_\mathrm{end} \in \mathbf{SE}(3)$ are the relative camera poses, defined from the current camera coordinate frame to the reference camera coordinate frame, corresponding to the blurry image at the beginning and end of the image capture process respectively, $\tau$ is the camera exposure time.
Note that while the fronto-parallel plane is defined in the reference camera frame, it may not remain fronto-parallel with respect to the $i^{th}$ virtual camera frame.
%
%To mitigate potential confusion, we illustrate the relationship in \figref{fig_geom_Ii}. By appropriate algebraic manipulations, we can derive $\bx_{\frac{i\tau}{n-1}}$ as follows:
%\begin{align}
%&\bx_{\frac{i\tau}{n-1}} = \pi (\bT_i \cdot \bp_\mathrm{3d}), \\
%%
%&\bp_\mathrm{3d} = \frac{d-p_z}{\lambda}
%                   \begin{bmatrix}
%                     x, &
%                     y, &
%                     z
%                   \end{bmatrix}^\mathrm{T}, \\
%%
%&\lambda = 2x \cdot q_0 + 2y \cdot q_1 + z \cdot q_2 \\
%%
%&q_0 = q_x q_z - q_w q_y,\\
%&q_1 = q_x q_w + q_y q_z,\\
%&q_2 = q_w^2 - q_x^2 - q_y^2 + q_z^2, \\
%%
%&\begin{bmatrix}
%  x, & y, & z
% \end{bmatrix}^\mathrm{T} = \pi^{-1}(\bx),
%%
%\end{align}
%where $\pi: \nR^3 \rightarrow \nR^2$ is the camera projection function, $(q_w, q_x, q_y, q_z)$ is the quaternion representation of the rotation matrix of $\bT_i$ and $(p_x,p_y,p_z)$ is the translation vector of $\bT_i$, $d$ is the depth of the plane with respect to the reference key-frame, $\pi^{-1}: \nR^2 \rightarrow \nR^3$ is the camera back projection function such that $x^2 + y^2 + z^2 = 1$. Detailed algebraic derivations as well as related Jacobian can be found in our \textcolor[RGB]{255, 0, 0}{supplementary material}.
%

%
We will illustrate the relationship in \figref{fig_geom_rela} to mitigate potential confusion.
We denote the translation vector $\bp_i^{ref}$ of $\bT_i^{ref}$ with $[p_x, p_y, p_z]^T$ and represent the rotation $\bR_i^{ref}$ with unit quaternion, \ie $\bar{\bq}$ = $[q_x$, $q_y$, $q_z$, $q_w]^T$. We denote $d$ as the depth of the frontal-parallel plane with respect to the reference key-frame. We can then compute the distance $d^\prime$ between the camera center of the $i^{th}$ virtual camera to the frontal-parallel plane as 
\begin{equation}
	d^\prime = d - p_z.
\end{equation}
The unitary ray of pixel $\bx \in \nR^2$ in the $i^{th}$ image $\bI_i$ can be computed by the back-projection function $\pi^{-1}: \nR^2 \rightarrow \nR^3$ as
\begin{equation}
	\begin{bmatrix}
		x, &
            y, &
            z
	\end{bmatrix}^\mathrm{T} = \pi^{-1}(\bx),
\end{equation}
where $x^2 + y^2 + z^2 = 1$. We can then compute cosine function of the angle (\ie $\theta$) between the unitary ray and the plane normal of the frontal parallel plane as 
\begin{align}
	\lambda = \cos(\theta) = (\bR_i^{ref} \cdot \pi^{-1}(\bx))^T \cdot \begin{bmatrix}
		0, &
            0, &
            1
	\end{bmatrix}^\mathrm{T},
\end{align}
from which we can further simply it as 
\begin{equation}
	\lambda = 2x (q_x q_z - q_w q_y) + 2 y (q_x q_w + q_y q_z) + z (q_w^2 - q_x^2 - q_y^2 + q_z^2).
\end{equation}
The length of the line segment $L$, which goes through pixel point $\bx$ from camera center of the $i^{th}$ camera and intersects with the frontal-parallel plane, can then be simply computed as 
\begin{equation}
	|L| = \frac{d^\prime}{\lambda} = \frac{d-p_z}{\lambda}.
\end{equation}
The 3D intersection point $\bp_{3d}$ between the line segment $L$ and the frontal parallel plane can thus be computed as 
\begin{equation}
	\bp_{3d} = |L| \begin{bmatrix}
		x, &
            y, &
            z
	\end{bmatrix}^\mathrm{T} = \frac{d - p_z}{\lambda} \begin{bmatrix}
		x, &
            y, &
            z
	\end{bmatrix}^\mathrm{T},
\end{equation}
where the $\bp_{3d}$ is represented in the coordinate frame of the $i^{th}$ camera. To compute the corresponding pixel point $\bx_\frac{i \tau}{n-1}$ in the reference image $\bI_\mathrm{ref}$, we need transform the 3D point $\bp_{3d}$ to the reference camera coordinate frame and then project it to the image plane. It can be formally defined as 
\begin{align}
	\bp_{3d}^\prime &= \bT_i^{ref} \cdot \bp_{3d},\\
	\bx_\frac{i \tau}{n-1} &= \pi(\bp_{3d}^\prime),
\end{align}
where $\bp_{3d}^\prime$ is the 3D point $\bp_{3d}$ represented in the reference camera, $\pi: \nR^3 \rightarrow \nR^2$ is the camera projection function.

\textit{Jacobian derivations}: The pose of the $i^{th}$ virtual camera, \ie $\bT_i^{ref}$, relates to $\bT_0^{ref}$ and $\bT_\tau^{ref}$ via \eqnref{eq_trajectory}. To estimate both $\bT_0^{ref}$ and $\bT_\tau^{ref}$, we need to have the Jacobian of $\bx_\frac{i \tau}{n-1}$ with respect to $\bT_i^{ref}$. Since the relationship between $\bx_\frac{i \tau}{n-1}$ and $\bT_i^{ref}$ is complex, as derived above, we use the Mathematica Symbolic Toolbox\footnote{https://www.wolfram.com/mathematica/} for the ease of Jacobian derivations.
The details are as follows.

\begin{align}
	\alpha_0 &= q_x x + q_y y + q_z z, &
	\alpha_1 &= q_y x - q_w z - q_x y, \\
	\alpha_2 &= q_w y - q_x z + q_z x, &
	\alpha_3 &= q_w x + q_y z - q_z y, \\
	\alpha_4 &= q_w z + q_x y - q_y x, &
	\beta_0 &= -2 (q_w q_z - q_x q_y), \\
	\beta_1 &= 2 (q_w q_y + q_x q_z), &
	\beta_2 &= 2 (q_w q_z + q_x q_y),\\
	\beta_3 &= -2 (q_w q_x - q_y q_z),&
	\beta_4 &= -2 (q_w q_y - q_x q_z),\\
	\beta_5 &= 2 (q_w q_x + q_y q_z),&
\end{align}
\begin{align}
	\gamma_0 &= x (q_w^2 + q_x^2 - q_y^2 - q_z^2) + y \beta_0 + z \beta_1,\\
	\gamma_1 &= x \beta_2 + y (q_w^2 - q_x^2 + q_y^2 - q_z^2) + z \beta_3,\\
	\gamma_2 &= x \beta_4 + y \beta_5 + z (q_w^2 - q_x^2 - q_y^2 + q_z^2),
\end{align}

\begin{align}
	\frac{\partial \bp_{3d}^\prime}{\partial p_x} &= \begin{bmatrix} 1 \\ 0 \\ 0 \end{bmatrix}, &
	\frac{\partial \bp_{3d}^\prime}{\partial p_y} &= \begin{bmatrix} 0 \\ 1 \\ 0 \end{bmatrix}, &
	\frac{\partial \bp_{3d}^\prime}{\partial p_z} &= 
	\begin{bmatrix} -\gamma_0 / \lambda \\
		-\gamma_1 / \lambda \\
		1- \gamma_2 / \lambda
	\end{bmatrix},
\end{align}
\vspace{-2em}

\begin{align}
	\frac{\partial \bp_{3d}^\prime}{\partial q_x} &= 
	2\frac{d-p_z}{\lambda}
	\begin{bmatrix}
		\alpha_0 - \alpha_2 \gamma_0 / \lambda\\
		\alpha_1 - \alpha_2 \gamma_1 / \lambda\\
		\alpha_2 - \alpha_2 \gamma_2 / \lambda
	\end{bmatrix}, 
\end{align}

\begin{align}
	\frac{\partial \bp_{3d}^\prime}{\partial q_y} &= 
	2\frac{d-p_z}{\lambda}
	\begin{bmatrix}
		\alpha_4 + \alpha_3 \gamma_0  / \lambda\\
		\alpha_0 + \alpha_3 \gamma_1  / \lambda\\
		- \alpha_3 + \alpha_3 \gamma_2 / \lambda
	\end{bmatrix},
\end{align}
\begin{align}
	\frac{\partial \bp_{3d}^\prime}{\partial q_z} &= 
	2\frac{d-p_z}{\lambda}
	\begin{bmatrix}
		-\alpha_2 + \alpha_0 \gamma_0 / \lambda\\
		\alpha_3 - \alpha_0 \gamma_1 / \lambda\\
		\alpha_0 - \alpha_0 \gamma_2 / \lambda
	\end{bmatrix},
\end{align}
\begin{align}
	\frac{\partial \bp_{3d}^\prime}{\partial q_w} &= 
	2\frac{d-p_z}{\lambda}
	\begin{bmatrix}
		\alpha_3 - \alpha_4 \gamma_0 / \lambda\\
		\alpha_2 - \alpha_4 \gamma_1 / \lambda\\
		\alpha_4 - \alpha_4 \gamma_2 / \lambda
	\end{bmatrix}.
\end{align}

The Jacobian $\frac{\partial \bx_\frac{i \tau}{n-1}}{\partial \bp_{3d}^\prime} \in \nR^{2\times3}$ is related to the camera projection function. For a pinhole camera model with the intrinsic parameters $f_x, f_y, c_x, c_y$, it can be derived as 
\begin{equation}
	\frac{\partial \bx_\frac{i \tau}{n-1}}{\partial \bp_{3d}^\prime} =
	\begin{bmatrix}
		\frac{f_x}{\bp_{3d_z}^\prime} & 0 & -\frac{f_x \bp_{3d_x}^\prime}{(\bp_{3d_z}^\prime)^2} \\
		0 & \frac{f_y}{\bp_{3d_z}^\prime} & -\frac{f_y \bp_{3d_y}^\prime}{(\bp_{3d_z}^\prime)^2} 
	\end{bmatrix},
\end{equation}
where $\bp_{3d}^\prime = [\bp_{3d_x}^\prime, \bp_{3d_y}^\prime, \bp_{3d_z}^\prime]^T$.

Based on the above derivations, we implemented the forward and backpropagation processes by pure CUDA, enabling our tracker to accurately estimate the local camera motion trajectory in real-time.

\subsection{Motion Blur Aware Mapper}\label{sec:bad_mapping}
The motion blur image formation model can be seamlessly integrated with various scene representation methods. We implement this in two versions: an implicit Radiance Fields version~\cite{nerf} and an explicit Gaussian Splatting version~\cite{kerbl3Dgaussians}, which can be freely switched between.
\subsubsection{Preliminaries: 3D Scene Representations}\label{sec:bad_mapping_scene}
We review the rendering processes of NeRF~\cite{nerf} and 3DGS~\cite{kerbl3Dgaussians}, as our backend mapper can utilize either NeRF or 3DGS.

\PAR{Implicit Radiance Fields.} Following \cite{chan2022efficient, johari2023eslam}, we embrace implicit scene representation comprises appearance and geometry tri-planes along with corresponding decoders. Formally, the volumetric rendering~\cite{max1995optical, or2022stylesdf} process can be delineated as follows.

%
% $\bT_c^w$ with intrinsic $\bK$
Given an image pixel location $\bx$ with depth $\lambda$, we can sample a specific 3D point $\bX^w$ along the ray from camera pose, and its corresponding appearance features~$\boldsymbol{f_{a}}$\footnote{We abbreviate $\boldsymbol{f_{a}}(\bX^w)$ and $\boldsymbol{f_{g}}(\bX^w)$ as $\boldsymbol{f_{a}}$ and $\boldsymbol{f_{g}}$, respectively.} and geometry features~$\boldsymbol{f_{g}}$ can be retrieved as:
\begin{align}
	\boldsymbol{f_{a}} &= F_{a\text{-}xy}(\bX^w) + F_{a\text{-}xz}(\bX^w) + F_{a\text{-}yz}(\bX^w), \\
	\boldsymbol{f_{g}} &= F_{g\text{-}xy}(\bX^w) + F_{g\text{-}xz}(\bX^w) + F_{g\text{-}yz}(\bX^w),
\end{align}
where \{$F_{a\text{-}xy}$, $F_{a\text{-}xz}$, $F_{a\text{-}yz}$\} and \{$F_{g\text{-}xy}$, $F_{g\text{-}xz}$, $F_{g\text{-}yz}$\} are appearance and geometry tri-planes, respectively.

Then the raw color~$\boldsymbol{c}$, signed distance function~$s$ and volume density~$\sigma$ can be computed as:
\begin{align}\label{eqn:decoder_density}
	\boldsymbol{c} = h_{a}\left(\boldsymbol{f_{a}}\right), s = h_{g}\left(\boldsymbol{f_{g}}\right), \sigma = \beta \cdot \text{sigmoid} \left(-\beta \cdot s \right),
\end{align}
where $h_{a}$ and $h_{g}$ are appearance and geometry decoders, respectively, $\beta$ is a learnable parameter regulating the sharpness of the surface boundary. Following volume rendering principles \cite{max1995optical}, we can compute the pixel color and depth by sampling 3D points along the ray as follows:
\begin{align}
	w_{i} &= \exp( -\sum_{k=1}^{i-1} {\sigma}_{k})( 1 - \exp (-{\sigma}_{i})),\\
	\bI(\bx) &= \sum_{i=1}^{n} w_{i} \boldsymbol{c}_{i}, \quad
	\bD(\bx) = \sum_{i=1}^{n} w_{i} d_{i},
\end{align}
where $n$ is the number of sampled 3D points along the ray, both $\boldsymbol{c}_i$ and $\sigma_i$ are the predicted color and volume density of the $i^{th}$ sampled 3D point via \eqnref{eqn:decoder_density}, $d_i$ is the depth of the $i^{th}$ sampled point.

\PAR{Explicit Gaussian Splatting.} 3DGS~\cite{kerbl3Dgaussians} represents the scene with a series of Gaussians, where each Gaussian $\bG$, is parameterized by its mean position $\boldsymbol{\mu} \in \mathbb{R}^3$, 3D covariance $\mathbf{\Sigma} \in \mathbb{R}^{3 \times 3}$, opacity $o \in \mathbb{R} $ and color $\bc \in \mathbb{R}^3$. The distribution of each scaled Gaussian is defined as:
\begin{equation}
	\bG(\bx) = e^{-\frac{1}{2}(\bx-\bmu)^{\top}\mathbf{\Sigma}^{-1}(\bx-\bmu)}.
	\label{eq:gauss}
\end{equation}

To ensure the semi-definite physical property of 3D convariance $\mathbf{\Sigma}$ and enable differentiable rasterization, 3DGS~\cite{kerbl3Dgaussians} represents the 3D convariance $\mathbf{\Sigma}$ and 2D convariance $\mathbf{\Sigma^{\prime}} \in \mathbb{R}^{2 \times 2}$ as follows:
\begin{align}\label{eqn:covar}
	\mathbf{\Sigma} = \mathbf{RSS}^{T}\mathbf{R}^{T}, \quad \mathbf{\Sigma^{\prime}} = \mathbf{JR}_c\mathbf{\Sigma R}_c^T\mathbf{J}^{T},
\end{align}
where $\bS \in \mathbb{R}^3$ is the scale, $\bR \in \mathbb{R}^{3 \times 3}$ is the rotation matrix stored by a quaternion $\bq \in \mathbb{R}^4$, $\mathbf{J} \in \mathbb{R}^{2 \times 3}$ is the Jacobian of the affine approximation of the projective transformation and $\mathbf{R}_c$ is the rotation part of the rendering camera pose $\mathbf{T}_c = \{\mathbf{R}_c \in \mathbb{R}^{3 \times 3}, \mathbf{t}_c \in \mathbb{R}^3\}$.

Afterward, each pixel color is rendered by rasterizing these $N$ sorted 2D Gaussians based on their depths, following the formulation:
\begin{align}\label{eqn:render_gs}
	\bI(\bx) = \sum_{i}^{N} \bc_i \alpha_i \bT_i, \quad \bD(\bx) = \sum_{i}^{N} d_i \alpha_i \bT_i,
\end{align}
where $\bc_i$ and $d_i$ are the color and depth of each Gaussian, $\bT_i = \prod_j^{i-1}(1-\alpha_j)$ and $\alpha_i = o_i \cdot \exp(-\frac{1}{2} {\rm \Delta}_i^T \mathbf{\Sigma^{\prime}}^{-1} {\rm \Delta}_i)$.

\begin{figure*}[t]
	\begin{tabular}{ccc}
		\includegraphics[width=0.31\textwidth]{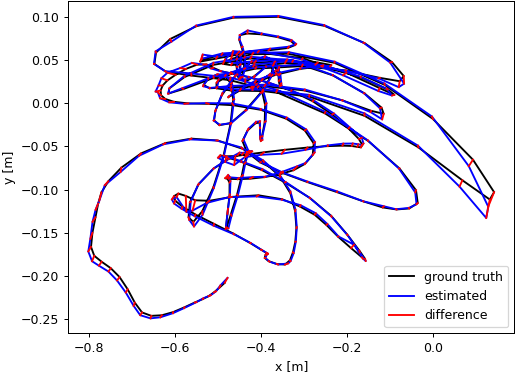} &
		\includegraphics[width=0.31\textwidth]{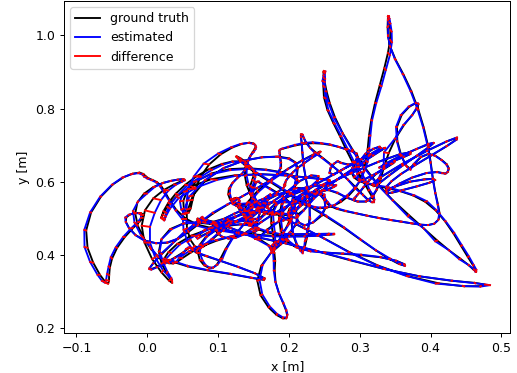} & 
		\includegraphics[width=0.31\textwidth]{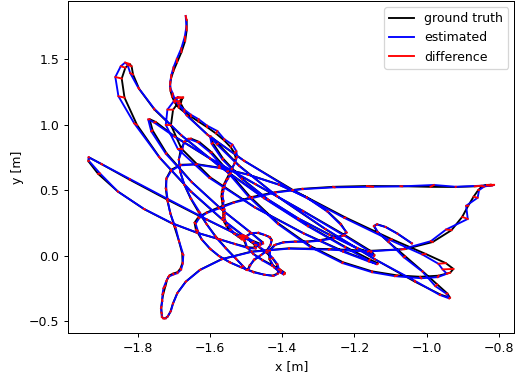} \\
		\scriptsize ArchViz-1 & \scriptsize ArchViz-2 & \scriptsize ArchViz-3
	\end{tabular}
	\vspace{-1em}
	\caption{\textbf{Estimated trajectories of \mbavo2 from the motion blurred image sequences of the ArchViz dataset.} It demonstrates that \mbavo2 can estimate accurate trajectories, although the camera motions are very challenging.}
	\label{fig:mba_vo_traj_archviz}
	\vspace{-1.0em}
\end{figure*}
% add blur model
\subsubsection{Blur Aware Mapper}\label{sec:bad_mapping_blur} 
Based on \eqnref{eq_blur_im_formation} and \eqnref{eq_trajectory}, we can also integrate the motion blur formation process to our mapper (either Radiance Fields~\cite{johari2023eslam} or 3DGS~\cite{kerbl3Dgaussians}), enabling the rendering of latent sharp images within the exposure time. If the mapper is Radiance Fields~\cite{johari2023eslam}, the learnable parameters are tri-planes ($F_a$, $F_g$),  decoders ($h_a$, $h_g$) and $\beta$; else if the mapper is 3DGS~\cite{kerbl3Dgaussians}, the learnable parameters are mean position $\boldsymbol{\mu}$, 3D covariance $\mathbf{\Sigma}$, opacity $o$ and color $\bc$. The objective of our blur aware mapper is now to estimate both $\bT_\mathrm{start}$ and $\bT_\mathrm{end}$ for each frame, alongside the mapper parameters. 

\subsubsection{Loss Functions}\label{sec:bad_mapping_loss} 
Given a batch of pixels $R$, the photo-metric \textbf{color loss} and geometric \textbf{depth loss} are:
\begin{align}
	\mathcal{L}_{c} &= \frac{1}{|R|} \sum_{\bx \in R} \norm{\bB(\bx) - \bB^{gt}(\bx)},\\
	\mathcal{L}_{d} &= \frac{1}{|R|} \sum_{\bx \in R} \norm{\bD(\bx) - \bD^{gt}(\bx)},
\end{align}
where $\bB(\bx)$ is the blurry color synthesized from $\bM_\theta$ using the above image formation model, which involves averaging all rendered colors along the camera trajectory, $\bB^{gt}(\bx)$ denotes the corresponding real captured blurry image. $\bD(\bx)$ is the rendered depth from the middle pose of the camera trajectory as $\bD^{gt}(\bx)$ typically denotes the measured depth.

Excluding the color and depth losses that are used in common, there are also some seperate losses used in Radiance Fields or Gaussian Splatting.

%
% \PAR{Radiance Fields Loss.}
\textit{If the scene representation is based on NeRF:} For rapid convergence and accurate geometry reconstruction, we apply the free space loss~$\mathcal{L}_{fs}$ and SDF loss~$\mathcal{L}_{sdf}$ to the sampled points, as the same with ESLAM~\cite{johari2023eslam}. We refer to prior works~\cite{johari2023eslam} for more details about these two losses.
% \begin{equation}
%     \mathcal{L}_{fs} = \frac{1}{|R|} \sum_{\bx \in R} \frac{1}{|P_{\bx}^{fs}|} \sum_{\bp \in P_{\bx}^{fs}} \norm{s(\bp) - 1},
% \end{equation}
% %
% \begin{equation}
%     \mathcal{L}_{sdf} = \frac{1}{|R|} \sum_{\bx \in R} \frac{1}{|P_{\bx}^{sdf}|} \sum_{\bp \in P_{\bx}^{sdf}} \norm{z(\bp) + s(\bp) \cdot T - D(\bp)},
% \end{equation}
%
%%where $T$ is the truncation, $P_{\bx}^{fs} = \{\bp~|~|d(\bp) - D(\bx)| > T\}$ and  $P_{\bx}^{sdf} = \{\bp~|~|d(\bp) - D(\bx)| \le T\}$. We apply $\mathcal{L}_{sdf}$ to all the points within the truncation region, with greater weight assigned to those closer to the surface (\ie~$P_{\bx}^{sdf-n} = \{\bp~|~|d(\bp) - D(\bx)| \le 0.4T\}$) and define $P_{\bx}^{sdf-f} = P_{\bx}^{sdf} -P_{\bx}^{sdf-n}$, then:
%\begin{align}
%        \mathcal{L}_{sdf-n} = \mathcal{L}_{sdf}(P_{\bx}^{sdf-n}), \\
%        \mathcal{L}_{sdf-f} = \mathcal{L}_{sdf}(P_{\bx}^{sdf-f}).
%\end{align}

The total loss function for Radiance Fields mapper is:
\begin{equation}
    \mathcal{L} = \lambda_{c}\mathcal{L}_{c} + \lambda_{d}\mathcal{L}_{d} + \lambda_{fs}\mathcal{L}_{fs} + \lambda_{sdf}\mathcal{L}_{sdf}.
\end{equation}
%The learnable parameters of mapper $\btheta$ (\ie~$F_a$, $F_g$, $h_a$, $h_g$ and $\beta$) and the camera trajectories (\ie~$\bT_\mathrm{start}$ and $\bT_\mathrm{end}$) are jointly estimated via backpropagation.
%

%
% \PAR{Gaussian Splatting Loss.}
\textit{If the scene representation is based on Gaussian Splatting:} To prevent the 3D Gaussian kernels from becoming overly skinny, we apply the scale regularization loss~\cite{xie2024physgaussian} to a batch of Gaussians $G$:
\begin{equation}
    \mathcal{L}_{reg} = \frac{1}{|G|} \sum_{\bg \in G} \max\{\max(\bS_g)/\min(\bS_g), r\} - r,
\end{equation}
where $\bS_g$ are the scales of 3DGS~\cite{kerbl3Dgaussians} in \eqnref{eqn:covar}. This loss constrains the ratio between the maximum and minor axis lengths to prevent the Gaussian from becoming too thin.

Combined SSIM loss $\rm{ssim}({\bB(\bx) - \bB^{gt}(\bx)})$, the final loss for Gaussian Splatting Mapper is:
\begin{equation}
    \mathcal{L} = \lambda_{c}\mathcal{L}_{c} + \lambda_{d}\mathcal{L}_{d} + \lambda_{ssim}\mathcal{L}_{ssim} + \lambda_{reg}\mathcal{L}_{reg}.
\end{equation}

We implemented Our-NeRF version approach based on implicit tri-plane based radiance fields~\cite{johari2023eslam}, and Our-GS version based on 3D Gaussian Splatting~\cite{kerbl3Dgaussians}.

\begin{table}[t]
	% \vspace{-2.5ex}
	\centering
	\vspace{-0ex}
	\caption{\add{\textbf{Tracking comparison} (ATE RMSE [cm]) of the proposed method vs. the SOTA methods on the synthetic \texttt{\#ArchViz dataset}. \ding{55} denotes that tracking fails while \ding{54} denotes the method can not successfully run on the ArchViz dataset due to code error.  The \colorbox{Green!30}{\textbf{best}} and \colorbox{Orange!30}{second-best} results are colored. \mbavo2 achieves best tracking performance in Radiance Fields and Gaussian Splatting approaches, respectively.}}
	\vspace{-1em}
	\label{tab:tracking_ArchViz}
	\resizebox{0.48\textwidth}{!}{
		\begin{tabular}{l|p{1.3cm}<{\centering}p{1.4cm}<{\centering}p{1.4cm}<{\centering}p{1.4cm}<{\centering}}
			\toprule
			Method      & \texttt{ArchViz-1}      & \texttt{ArchViz-2}     & \texttt{ArchViz-3}    & \texttt{\textbf{Avg.}} \\
			\midrule
			iMAP~\cite{sucar2021imap} & 255.75 & 186.87 & 756.55 & 399.72\\
			NICE-SLAM~\cite{zhu2022nice} & \ding{54} & \ding{54} & \ding{54} & - \\
			VoxFusion~\cite{yang2022vox} & 5.54 & 16.82  & 249.30 & 90.55\\
			CoSLAM~\cite{wang2023co} &  \snd 5.28 & \snd 4.67  & \snd 14.17 &  \snd 8.04\\
			ESLAM~\cite{johari2023eslam} & 20.12 & 12.61  & \ding{54} & -\\
			Point-SLAM~\cite{sandstrom2023point} & 289.56 & 181.27 & 596.25 &  355.69 \\
			Ours-NeRF & \fs 0.98 & \fs 1.13 & \fs 2.96  & \fs 1.69 \\
			\midrule
			\midrule
			SplaTAM~\cite{keetha2024splatam} & 36.88 & \ding{55} & 763.93 & -  \\
                \add{RTG-SLAM}~\cite{peng2024rtgslam} & \add{\ding{54}} & \add{17.49} & \add{26.78} & -  \\
                %
                % \add{GS-ICP-SLAM}~\cite{ha2024gsicp} & \add{x} & \add{x} & \add{x} & \add{x}  \\
                %
                \add{Photo-SLAM}~\cite{hhuang2024photoslam} & \add{4.57} & \snd \add{0.64} & \snd \add{2.01} & \snd \add{2.41} \\
        \add{MonoGS}~\cite{Matsuki:Murai:etal:CVPR2024} & \snd \add{1.92} & \add{2.96} & \add{38.37} & \add{14.42}  \\
			Ours-GS & \fs 0.75 & \fs 0.36 & \fs 1.41  & \fs 0.84 \\
			\bottomrule
		\end{tabular}
	}
	%	\vspace{-1ex}
\end{table}
\section{Experiments}
\label{sec:experiment}
The experimental setup across different datasets is described in Sec.~\ref{exp_setup}. We show that \mbavo2 outperforms state-of-the-art dense visual SLAM methods on blurred datasets in Sec.~\ref{exp_eval_blur} and sharp datasets in Sec.~\ref{exp_eval_sharp}. Runtime analysis and ablation studies are conducted in Sec.~\ref{exp_analysis} and Sec.~\ref{exp_ablation}.
\subsection{Experimental Setup}
\label{exp_setup}
\PAR{Baselines.} We compared our method with state-of-the-art NeRF-based approaches: iMAP~\cite{sucar2021imap}, NICE-SLAM~\cite{zhu2022nice}, Vox-Fusion~\cite{yang2022vox} CoSLAM~\cite{wang2023co}, ESLAM~\cite{johari2023eslam} and Point-SLAM~\cite{sandstrom2023point}. Further, we also report the results of Gaussian Splatting based visual SLAMs, \ie GS-SLAM~\cite{yan2024gs} and SplaTAM~\cite{keetha2024splatam}. As for Gaussian Splatting SLAMs, we extract the mesh from the estimated camera poses and rendered images and depths, following GS-SLAM~\cite{yan2024gs}.

\PAR{Datasets.} To evaluate the robustness and performance of \mbavo2, we conducted thorough assessments across a wide range of scenes sourced from diverse datasets, covering scenarios both with and without camera motion blur. \textbf{\textit{Motion blur data: }} Our evaluation encompasses synthetic blur datasets like ArchViz~\cite{mba-vo} that are rendered by the Unreal game engine\footnote{https://www.unrealengine.com} with a camera being quickly shaken back and forth, and 3 selected part sequence with motion blur from ScanNet~\cite{dai2017scannet} and TUM RGB-D~\cite{sturm2012benchmark}. Futher, we captured another 3 motion blur sequences with Realsense RGB-D camera. The groundtruth trajectory is provided by an indoor motion capturing system\footnote{https://www.vicon.com}. \textbf{\textit{Standard sharp data:}} Additionally, we incorporate three standard public 3D benchmarks commonly used in previous state-of-the-art methods: Replica~\cite{replica19arxiv}, ScanNet~\cite{dai2017scannet} and TUM RGB-D~\cite{sturm2012benchmark}.

\PAR{Metrics.} We utilize the widely adopted absolute trajectory error (\ie ATE RMSE)~\cite{sturm2012benchmark} to assess tracking performance. In evaluating reconstruction and geometric accuracy, we employ the culled mesh~\cite{lorensen1987marching} to compute the 2D Depth L1 metric (cm), as well as 3D metrics such as Precision (\%), Recall (\%) and F1-score ($<$ 1cm \%). To fairly compare these baselines, we run all the methods 5 times and report the average results. We further evaluate the rendering quality using classical signal-to-noise ratio (PSNR), SSIM and LPIPS~\cite{zhang2018unreasonable} metrics.

\PAR{Implementation Details.}~{\bf\textit{Tracker:}} To enhance efficiency, we sub-sample high-gradient pixels to derive sparse keypoints, ensuring uniform distribution across the image. Additionally, we define $9\times9$ local patches around each sampled sparse keypoint to facilitate better convergence. The energy function undergoes optimization in a coarse-to-fine manner, supplemented by the application of a robust huber loss function for improved robustness. The CUDA-accelerated backpropagation process implemented in our tracker is well-suited for real-time applications. ~{\bf\textit{Mapper:}} We implement the mapper using either Radiance Fields or 3DGS. \textit{If mapper is Radiance Fields:} We interpolate 7 virtual images between $\bT_{\mathrm{start}}$ and $\bT_{\mathrm{end}}$ ($n$ in Eq. \ref{eq_blur_im_formation}) following~\cite{wang2023badnerf}. Other details are same as~\cite{johari2023eslam}. \textit{If mapper is Gaussian Splatting:} We implement the GS version of \mbavo2 using PyTorch within the gsplat framework~\cite{ye2024gsplatopensourcelibrarygaussian}. The number of virtual images is set to 13. The coefficients and regularization $r$ for $\mathcal{L}_{reg}$ is $1e1$ and 1.0. Other configurations are as default in gsplat. As for Gaussians management, we add gaussians for new frame at those areas that alpha value is smaller than 0.5 and we adopt the default splitting, cloning, and pruning strategy in gsplat. For Gaussian color initialization, we directly utilize input blur color and depth information on these pixels to spawn newly added Gaussian primitives. Additionally, we have created interfaces in the tracker to facilitate PyTorch data transmission from the mapper to our CUDA-implemented tracker.
\begin{figure*}[t]
	\setlength\tabcolsep{1pt}
	\centering
	\begin{tabular}{cccccc}
		%		\rot{Input} &
		&\raisebox{-0.02in}{\rotatebox[origin=t]{90}{\scriptsize Input}}&\includegraphics[valign=m,width=0.46\textwidth]{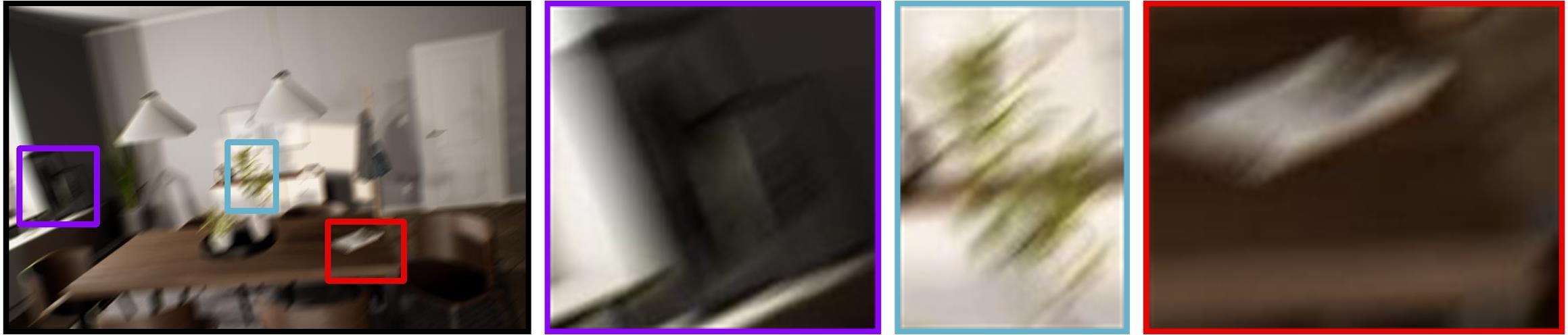}
		& &\raisebox{-0.035in}{\rotatebox[origin=t]{90}{\scriptsize Groundtruth}} &\includegraphics[valign=m,width=0.46\textwidth]{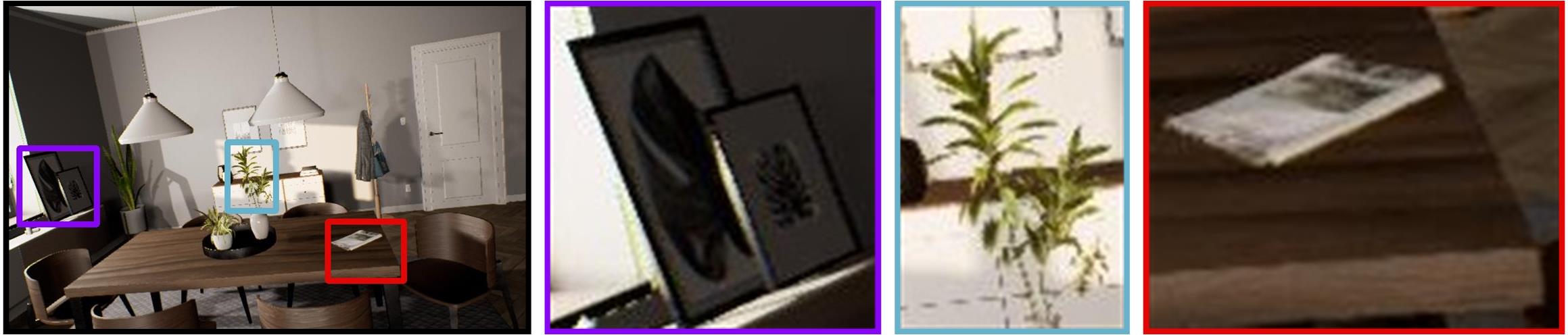}\\
        \specialrule{0em}{.1em}{.1em}
        &\raisebox{-0.02in}{\rotatebox[origin=t]{90}{\scriptsize CoSLAM\cite{wang2023co}}}&\includegraphics[valign=m,width=0.46\textwidth]{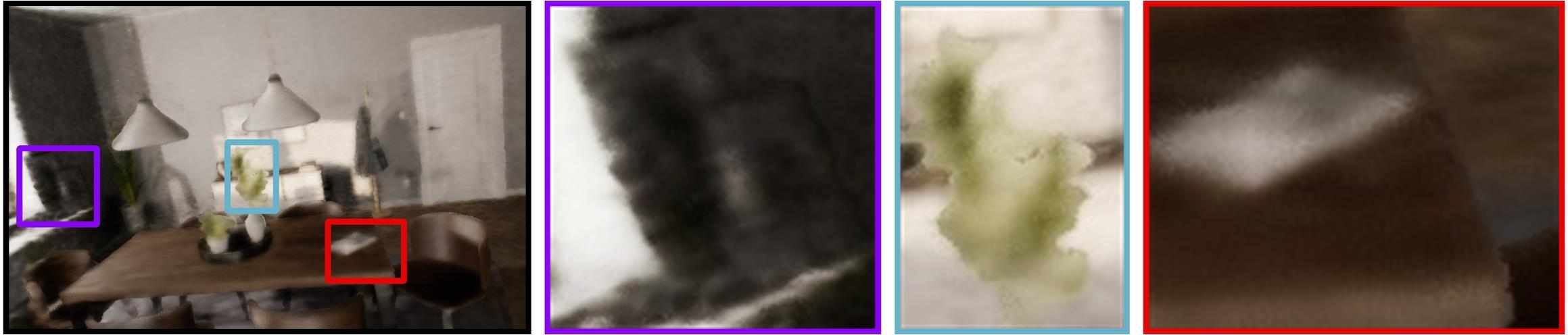}
		& \raisebox{-0.035in}{\rotatebox[origin=t]{90}{\scriptsize Spla}}
		&\raisebox{-0.035in}{\rotatebox[origin=t]{90}{\scriptsize TAM \cite{keetha2024splatam}}} &\includegraphics[valign=m,width=0.46\textwidth]{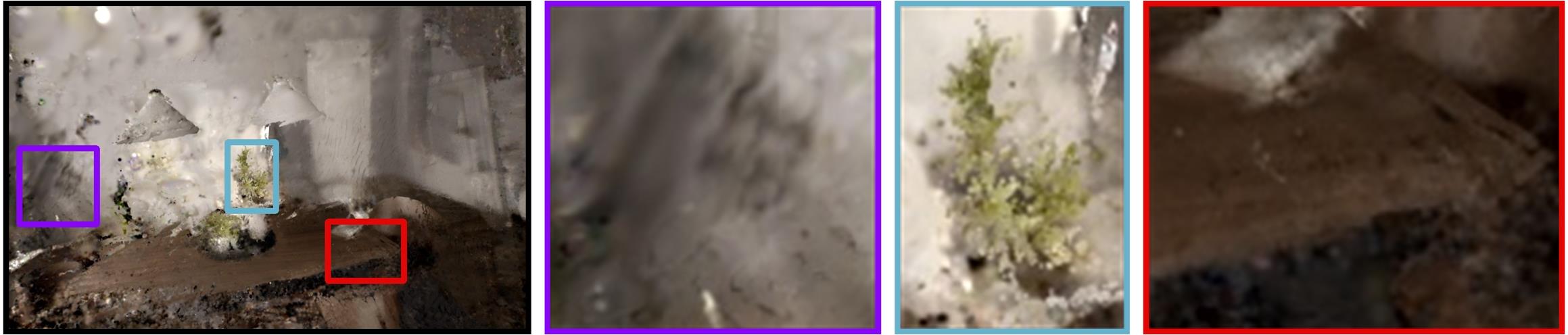}\\
        \specialrule{0em}{.1em}{.1em}
        &\raisebox{-0.02in}{\rotatebox[origin=t]{90}{\scriptsize ESLAM \cite{johari2023eslam}}}
		& \includegraphics[valign=m,width=0.46\textwidth]{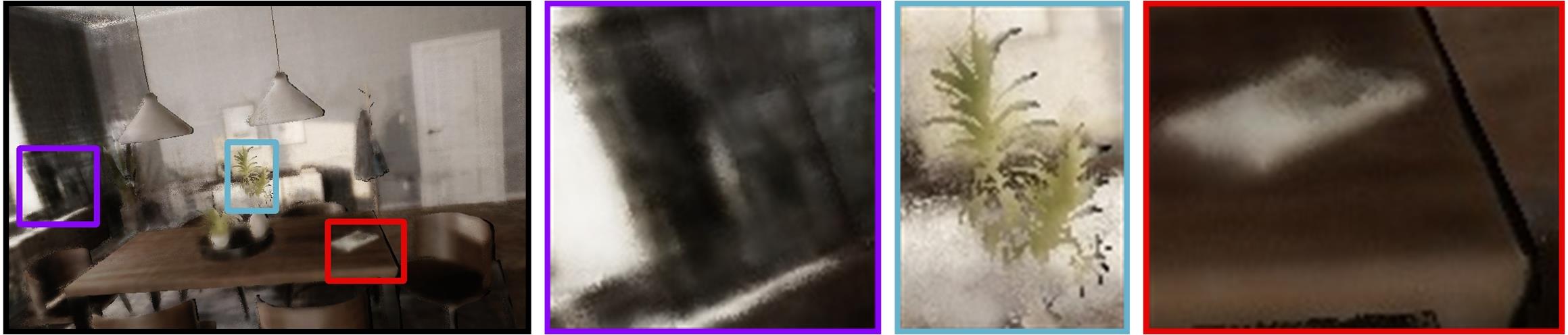}
		& & \raisebox{-0.035in}{\rotatebox[origin=t]{90}{\scriptsize \add{MonoGS}\cite{Matsuki:Murai:etal:CVPR2024}}} &\includegraphics[valign=m,width=0.46\textwidth]{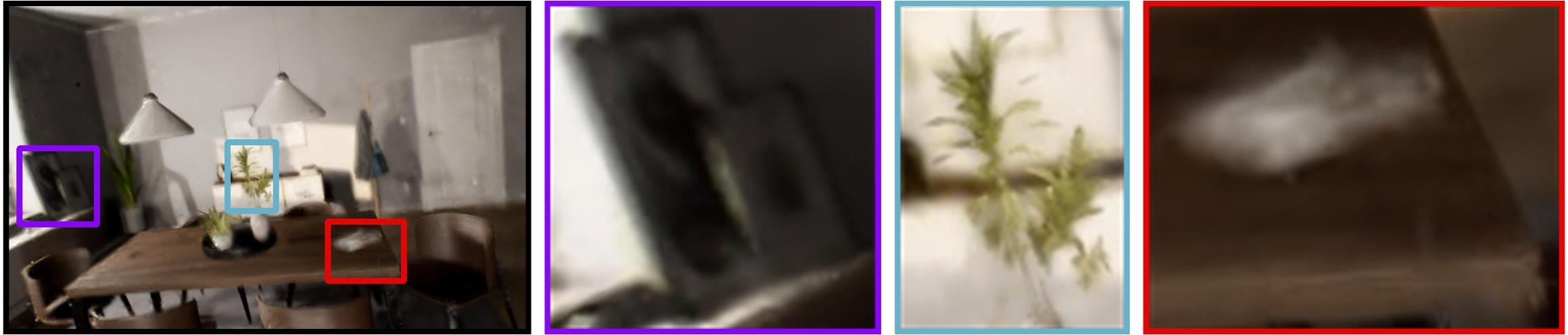}\\
		\specialrule{0em}{.1em}{.1em}
		\raisebox{-0.035in}{\rotatebox[origin=t]{90}{\scriptsize Point}}
		&\raisebox{-0.035in}{\rotatebox[origin=t]{90}{\scriptsize SLAM \cite{sandstrom2023point}}} &\includegraphics[valign=m,width=0.46\textwidth]{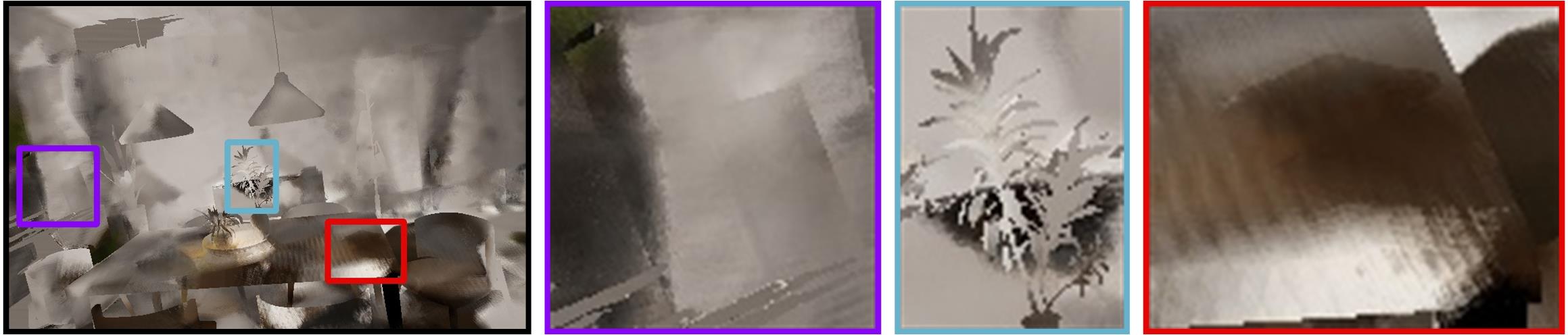}
		& \raisebox{-0.035in}{\rotatebox[origin=t]{90}{\scriptsize \add{Photo}}}
		&\raisebox{-0.035in}{\rotatebox[origin=t]{90}{\scriptsize \add{SLAM} \cite{hhuang2024photoslam}}} &\includegraphics[valign=m,width=0.46\textwidth]{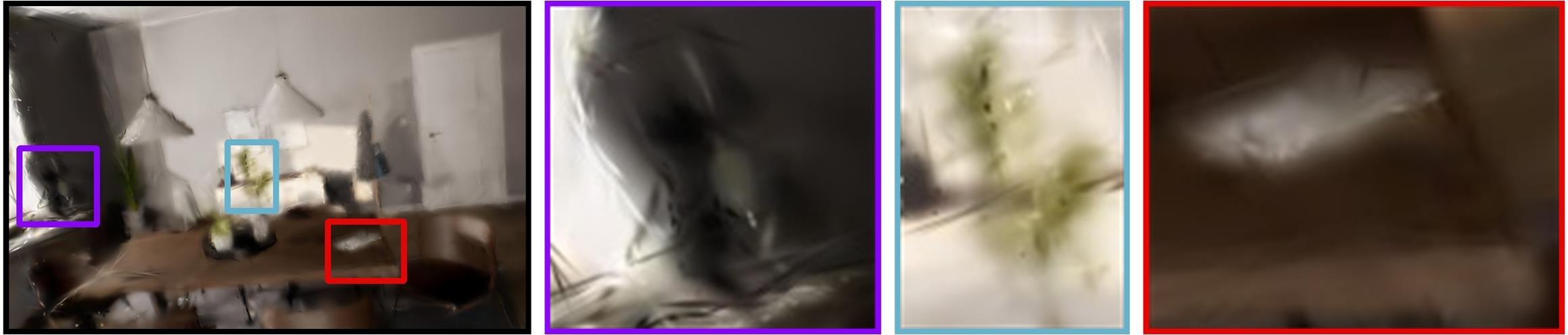}\\
        \specialrule{0em}{.1em}{.1em}
		&\raisebox{-0.035in}{\rotatebox[origin=t]{90}{\scriptsize Ours-NeRF}} &\includegraphics[valign=m,width=0.46\textwidth]{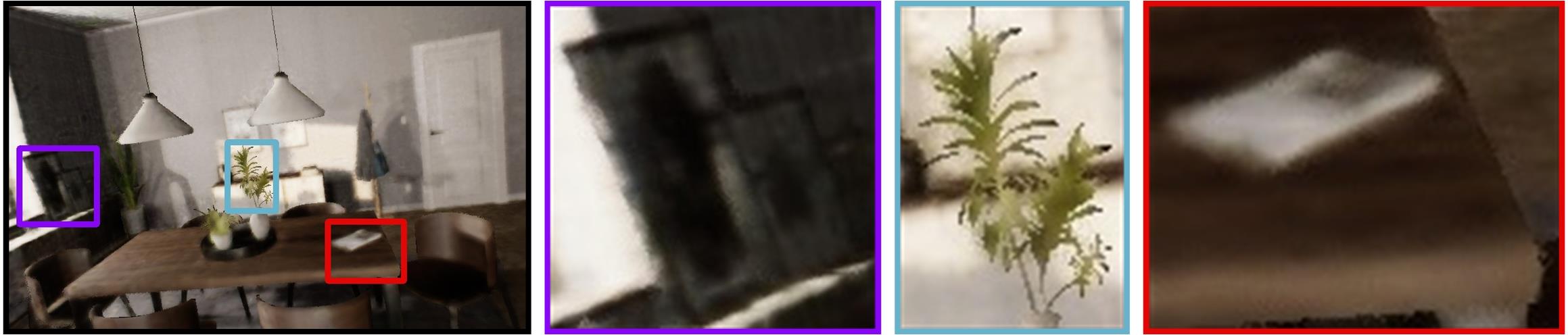}
		& &\raisebox{-0.035in}{\rotatebox[origin=t]{90}{\scriptsize Ours-GS}} &\includegraphics[valign=m,width=0.46\textwidth]{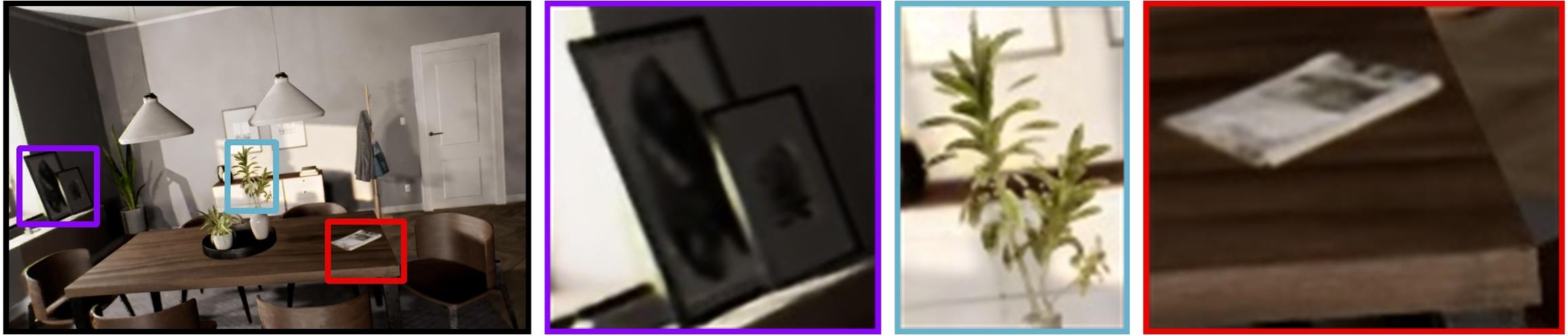}\\
	\end{tabular}
	% \vspace{-0.5em}
	\caption{\add{\textbf{Qualitative rendering results of different methods with synthetic ArchViz datasets.} It demonstrates that \mbavo2 can restore and render sharp images from blurry input and outperform other dense visual SLAMs. Best viewed in high resolution.}}
	\label{fig:mbavo_img}
	% \vspace{-1em}
\end{figure*}

\begin{table}[t]
	% \vspace{2ex}
	\centering
	\caption{\add{\textbf{Rendering and Reconstruction} comparison of the proposed method vs. the SOTA methods on the synthetic \texttt{\#ArchViz dataset}. \ding{55} denotes that rendering fails or unmeaningful rendering while \ding{54} denotes the method can not successfully run on the ArchViz dataset due to code error.  \mbavo2 achieves best tracking performance in Radiance Fields and Gaussian Splatting approaches, respectively.}}
	\label{tab:psnr_archviz}
    \vspace{-1em}
	\resizebox{0.48\textwidth}{!}{
		\setlength{\tabcolsep}{2pt}
		\renewcommand{\arraystretch}{1}
		\begin{tabular}{c|r|cccc|l|c}
			\toprule
			Method & Metric~~ & \texttt{Arch1}      & \texttt{Arch2}      & \texttt{Arch3} & \textbf{Avg.} & Metric-Rec. & \texttt{Arch1} \\
			% \midrule
			% \multirow{3}{*}{\begin{tabular}[l]{@{}c@{}}iMAP\\~\cite{sucar2021imap}\end{tabular}}
			% & PSNR $\uparrow$ & 17.83  & 15.78   & 13.08  & 15.56 & Precision $\uparrow$ & - \\
			% & SSIM $\uparrow$  & 0.622 & 0.451  & 0.442 & 0.505 & Recall $\uparrow$ & - \\
			% & LPIPS $\downarrow$ & 0.441 & 0.642  & 0.651 & 0.578 & F1 $\uparrow$ & - \\
			\midrule
			\multirow{3}{*}{\begin{tabular}[l]{@{}c@{}}NICESL\\AM~\cite{zhu2022nice}\end{tabular}}
			& PSNR $\uparrow$ &   &    &   & - & Precision $\uparrow$ & \\
			& SSIM $\uparrow$  & \ding{54} & \ding{54}  & \ding{54} & - & Recall $\uparrow$ & \ding{54}\\
			& LPIPS $\downarrow$ &  &   &  & - & F1 $\uparrow$ & \\
			\midrule
			\multirow{3}{*}{\begin{tabular}[l]{@{}c@{}}VoxFus\\ion~\cite{yang2022vox}\end{tabular}}
			& PSNR $\uparrow$ & 17.97   & 17.07 & 13.52 & 16.19 & Precision $\uparrow$ & 38.01 \\
			& SSIM $\uparrow$  & 0.519   & 0.447 & 0.392 & 0.452 & Recall $\uparrow$ & 51.44 \\
			& LPIPS $\downarrow$ & 0.511   & 0.593 & 0.790 & 0.631 & F1 $\uparrow$ & 43.71 \\
			\midrule
			\multirow{3}{*}{\begin{tabular}[l]{@{}c@{}}CoSLA\\M~\cite{wang2023co}\end{tabular}}
			& PSNR $\uparrow$ & \snd 22.97   & 21.98 & \snd 21.45 & \snd 22.13 & Precision $\uparrow$ & \snd 55.71 \\
			& SSIM $\uparrow$  & \snd 0.841   & 0.685 & \snd 0.744 & \snd 0.756 & Recall $\uparrow$ & \snd 61.59 \\
			& LPIPS $\downarrow$ & \snd 0.361   & 0.570 & \snd 0.477 & \snd 0.469 & F1 $\uparrow$ & \snd 58.50 \\
			\midrule
			\multirow{3}{*}{\begin{tabular}[l]{@{}c@{}}ESL\\AM~\cite{johari2023eslam}\end{tabular}}
			& PSNR $\uparrow$ & 21.07  & \snd 23.87   &   & - & Precision $\uparrow$ & 44.61 \\
			& SSIM $\uparrow$  & 0.766 & \snd 0.785  & \ding{54} & - & Recall $\uparrow$ & 54.88 \\
			& LPIPS $\downarrow$ & 0.446 & \snd 0.475  &  & - & F1 $\uparrow$ & 48.22 \\
			\midrule
			\multirow{3}{*}{\begin{tabular}[l]{@{}c@{}}Point\\SLAM~\cite{sandstrom2023point}\end{tabular}}
			& PSNR $\uparrow$ & 15.90   &       & 18.29  & - & Precision $\uparrow$ & 1.45 \\
			& SSIM $\uparrow$  & 0.493 & \ding{55}  & 0.563 & - & Recall $\uparrow$ & 18.21\\
			& LPIPS $\downarrow$ & 0.549 &   & 0.479 & - & F1 $\uparrow$ & 2.68 \\
			\midrule
			\multirow{3}{*}{\begin{tabular}[c]{@{}c@{}}Ours\\NeRF\end{tabular}}
			& PSNR $\uparrow$ & \fs 24.77   & \fs 26.35 & \fs 24.13 & \fs 25.08 & Precision $\uparrow$ & \fs 82.49 \\
			& SSIM $\uparrow$ & \fs 0.905   & \fs 0.875 & \fs 0.880 & \fs 0.887 & Recall $\uparrow$ & \fs 90.56 \\
			& LPIPS $\downarrow$ & \fs 0.312   & \fs 0.372 & \fs 0.370 & \fs 0.351 & F1 $\uparrow$ & \fs 86.95 \\
			\midrule
			\midrule
			\multirow{3}{*}{\begin{tabular}[l]{@{}c@{}}Spla\\TAM~\cite{keetha2024splatam}\end{tabular}}
			& PSNR $\uparrow$ & 17.69  &    &   & - & Precision $\uparrow$ & 15.57 \\
			& SSIM $\uparrow$  & 0.538 & \ding{55}  & \ding{55} & - & Recall $\uparrow$ & 28.49\\
			& LPIPS $\downarrow$ & 0.542 &   &  & - & F1 $\uparrow$ & 20.21 \\
			\midrule
            \multirow{3}{*}{\begin{tabular}[l]{@{}c@{}}\add{RTG}\\\add{SLAM}~\cite{peng2024rtgslam}\end{tabular}}
			& \add{PSNR} $\uparrow$ &   &  \add{20.08}  &  \add{17.99} & - & \add{Precision} $\uparrow$ &  \\
			& \add{SSIM} $\uparrow$  & \add{\ding{54}} & \add{0.558}  & \add{0.595} & - & \add{Recall} $\uparrow$ & \add{\ding{54}}\\
			& \add{LPIPS} $\downarrow$ &  & \add{0.551} & \add{0.538} & - & \add{F1} $\uparrow$ &  \\
			\midrule
            %
   %          \multirow{3}{*}{\begin{tabular}[l]{@{}c@{}}\add{GS-ICP}\\SLAM~\cite{ha2024gsicp}\end{tabular}}
			% & PSNR $\uparrow$ & -  &  -  &  - & - & Precision $\uparrow$ & - \\
			% & SSIM $\uparrow$  & - & -  & - & - & Recall $\uparrow$ & -\\
			% & LPIPS $\downarrow$ & - & - & - & - & F1 $\uparrow$ & - \\
   %          \midrule
            %
            \multirow{3}{*}{\begin{tabular}[l]{@{}c@{}}\add{Photo}\\\add{SLAM}~\cite{hhuang2024photoslam}\end{tabular}}
			& \add{PSNR}~$\uparrow$ & \add{\snd 23.45}  &  \add{\snd 27.49}  &  \add{\snd 23.70} & \add{\snd 24.88} & \add{Precision}~$\uparrow$ & \add{\snd 52.69} \\
			& \add{SSIM}~$\uparrow$  & \add{0.823} & \add{\snd 0.897}  & \add{\snd 0.846} & \add{\snd 0.855} & \add{Recall}~$\uparrow$ & \add{\snd 63.88}\\
			& \add{LPIPS}~$\downarrow$ & \add{0.387} & \add{0.363} & \add{\snd 0.396} & \add{0.382} & \add{F1}~$\uparrow$ & \add{\snd 53.76} \\
            \midrule
            \multirow{3}{*}{\begin{tabular}[l]{@{}c@{}}\add{Mono}\\\add{GS}~\cite{Matsuki:Murai:etal:CVPR2024}\end{tabular}}
			& \add{PSNR}~$\uparrow$ & \add{23.18} & \add{27.14} & \add{20.56} & \add{23.63} & \add{Precision}~$\uparrow$ & \add{44.43} \\
			& \add{SSIM}~$\uparrow$  & \add{\snd 0.852} & \add{0.884} & \add{0.696} & \add{0.811} & \add{Recall}~$\uparrow$ & \add{57.61}\\
			& \add{LPIPS}~$\downarrow$ & \add{\snd 0.337} & \add{\snd 0.329} & \add{0.427} & \add{\snd 0.364} & \add{F1}~$\uparrow$ & \add{50.17} \\
            \midrule
			\multirow{3}{*}{\begin{tabular}[c]{@{}c@{}}Ours-GS\end{tabular}}
			& PSNR $\uparrow$ & \fs 28.45   & \fs 30.16 & \fs 27.85 & \fs 28.82 & Precision $\uparrow$ & \fs 83.71 \\
			& SSIM $\uparrow$  & \fs 0.956   & \fs 0.951 & \fs 0.943 & \fs 0.950 & Recall $\uparrow$ & \fs 93.67 \\
			& LPIPS $\downarrow$ & \fs 0.148   & \fs 0.172 & \fs 0.171 & \fs 0.164 & F1 $\uparrow$ & \fs 88.41 \\
			\bottomrule
		\end{tabular}
	}
	\vspace{-1em}
\end{table}

\begin{figure*}
        % \vspace{-3ex}
	\setlength\tabcolsep{1pt}
	\centering
	\begin{tabular}{ccc}
		%		\rot{Input} & 
		\raisebox{-0.035in}{\rotatebox[origin=t]{90}{\scriptsize Point}}
		&\raisebox{-0.02in}{\rotatebox[origin=t]{90}{\scriptsize SLAM\cite{sandstrom2023point}}}
		& \includegraphics[valign=m,width=0.93\textwidth]{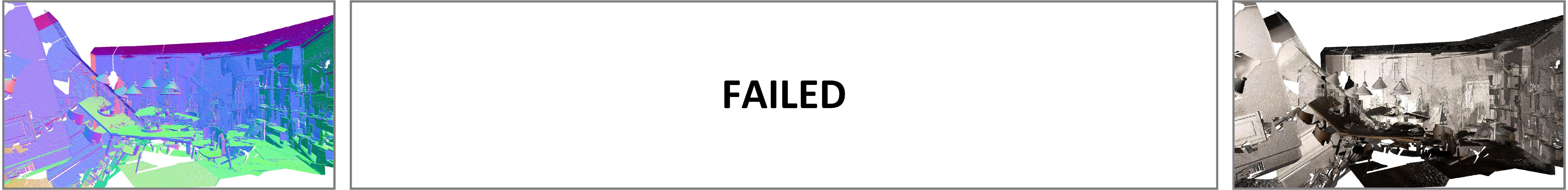}\\
		
		\specialrule{0em}{.1em}{.1em}
		\raisebox{-0.035in}{\rotatebox[origin=t]{90}{\scriptsize VoxFu}}
		&\raisebox{-0.02in}{\rotatebox[origin=t]{90}{\scriptsize sion\cite{yang2022vox}}}
		& \includegraphics[valign=m,width=0.93\textwidth]{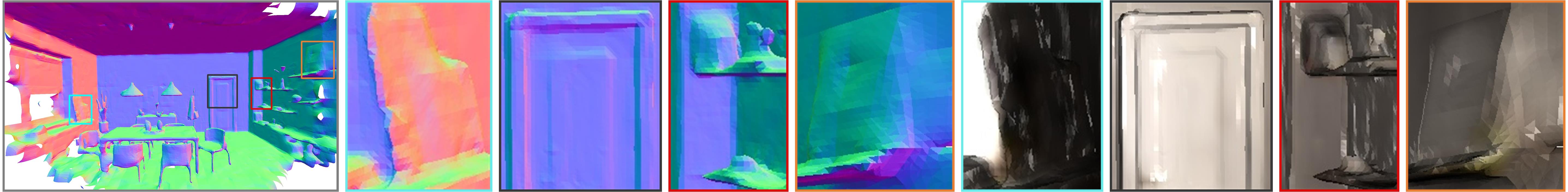}\\
		
		\specialrule{0em}{.1em}{.1em}
		&\raisebox{-0.02in}{\rotatebox[origin=t]{90}{\scriptsize CoSLAM\cite{wang2023co}}}
		& \includegraphics[valign=m,width=0.93\textwidth]{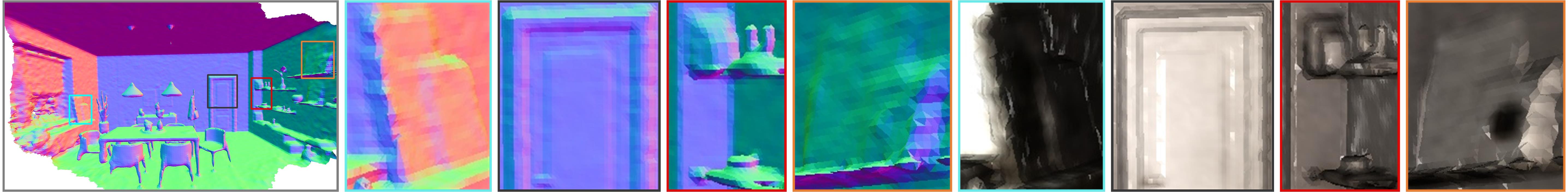}\\
		
		\specialrule{0em}{.1em}{.1em}
		&\raisebox{-0.02in}{\rotatebox[origin=t]{90}{\scriptsize ESLAM\cite{johari2023eslam}}}
		& \includegraphics[valign=m,width=0.93\textwidth]{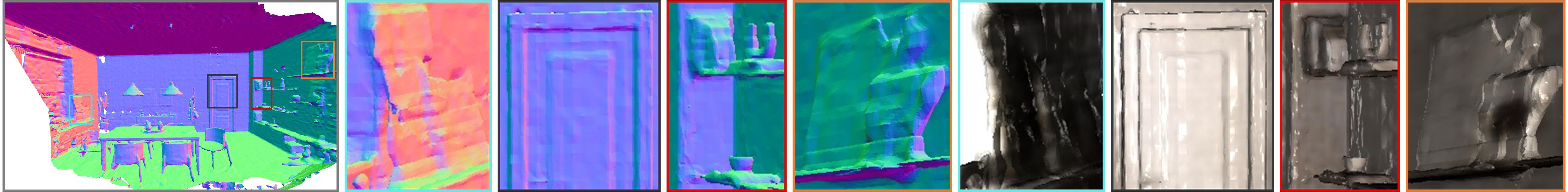}\\
		
		\specialrule{0em}{.1em}{.1em}
		&\raisebox{-0.02in}{\rotatebox[origin=t]{90}{\scriptsize Ours-NeRF}}
		& \includegraphics[valign=m,width=0.93\textwidth]{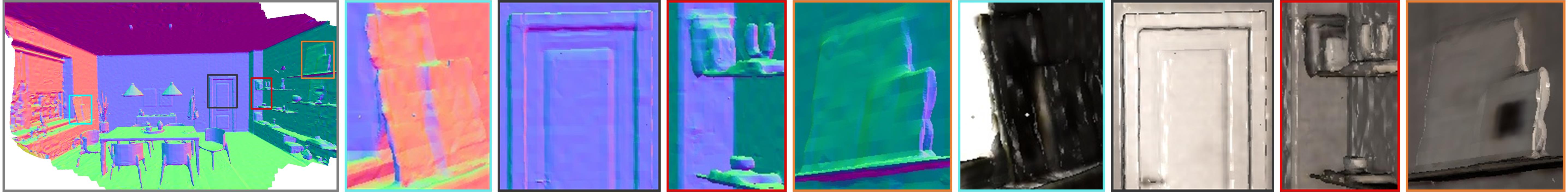}\\
		
		\specialrule{0em}{.1em}{.1em}
		\raisebox{-0.035in}{\rotatebox[origin=t]{90}{\scriptsize Spla}}
		&\raisebox{-0.02in}{\rotatebox[origin=t]{90}{\scriptsize TAM\cite{keetha2024splatam}}}
		& \includegraphics[valign=m,width=0.93\textwidth]{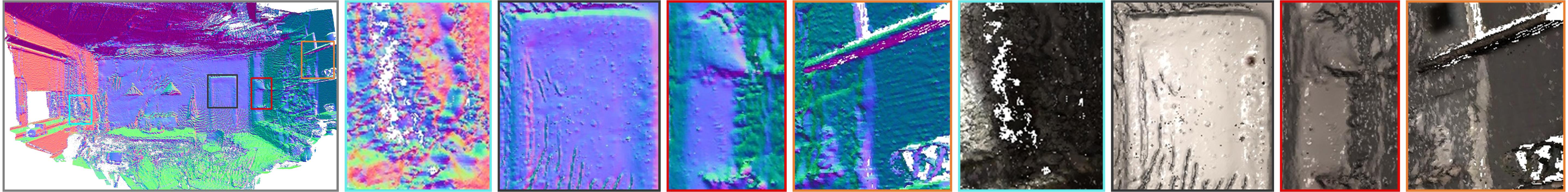}\\

        \specialrule{0em}{.1em}{.1em}
		\raisebox{-0.035in}{\rotatebox[origin=t]{90}{\scriptsize \add{Photo}}}
		&\raisebox{-0.02in}{\rotatebox[origin=t]{90}{\scriptsize \add{SLAM}\cite{hhuang2024photoslam}}}
		& \includegraphics[valign=m,width=0.93\textwidth]{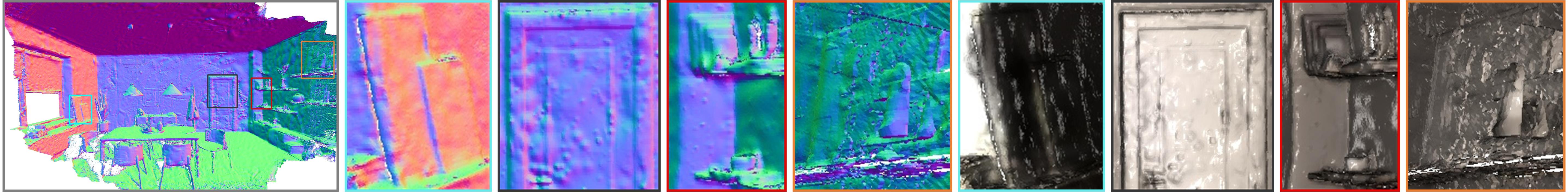}\\

        \specialrule{0em}{.1em}{.1em}
		&\raisebox{-0.02in}{\rotatebox[origin=t]{90}{\scriptsize \add{MonoGS}\cite{Matsuki:Murai:etal:CVPR2024}}}
		& \includegraphics[valign=m,width=0.93\textwidth]{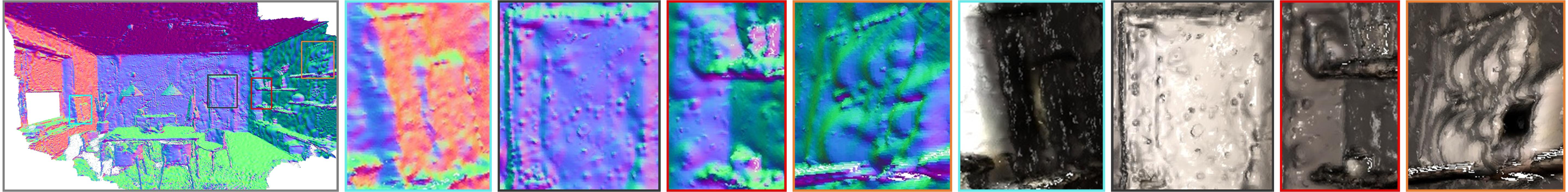}\\
		
		\specialrule{0em}{.1em}{.1em}
		&\raisebox{-0.02in}{\rotatebox[origin=t]{90}{\scriptsize Ours-GS}}
		& \includegraphics[valign=m,width=0.93\textwidth]{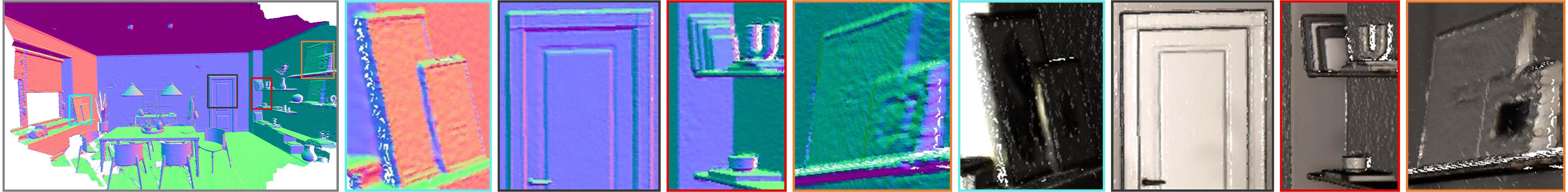}\\
		
		\specialrule{0em}{.1em}{.1em}
		&\raisebox{-0.02in}{\rotatebox[origin=t]{90}{\scriptsize Reference}}
		& \includegraphics[valign=m,width=0.93\textwidth]{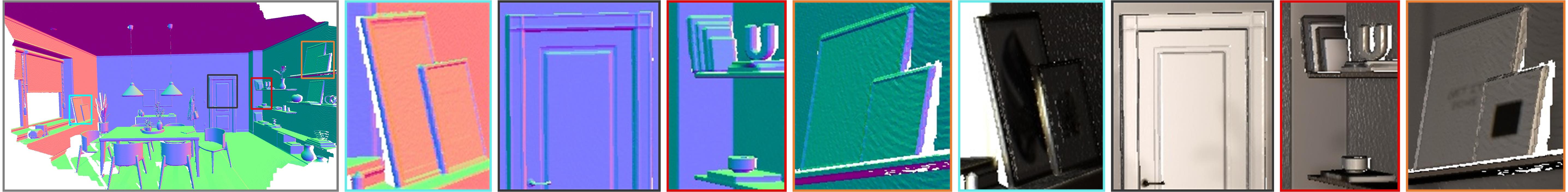}\\
		
	\end{tabular}
	% \vspace{-0.5em}
	\caption{\add{\textbf{Qualitative mesh visualization of different methods with ArchViz-1 datasets.} The result reveals that implicit Radiance Fields (\eg CoSLAM, ESLAM) deliver better reconstruction mesh performance than explicit point based methods (\ie Point-SLAM, SplaTAM). \mbavo2 always achieves best performance, no matter Ours-NeRF or Our-GS. RTG-SLAM fails to reconstruct mesh.}}
	\label{fig:mbavo_mesh}
	% \vspace{-1em}
\end{figure*}

\begin{table}[t] 
	% \vspace{-2.5ex}
	\centering
	\vspace{-0ex}
	\caption{\add{\textbf{Tracking comparison} (ATE RMSE [cm]) of the proposed method vs. the SOTA methods on the real-world \texttt{\#TUM, ScanNet: left} and real captured \texttt{\#Realsense: right} dataset. Our method achieves better tracking performance in real blurry dataset.}}
	\label{tab:tracking_real_blur}
    \vspace{-1em}
	\resizebox{0.48\textwidth}{!}{
		\begin{tabular}{l|cccc|cccc}
			\toprule
			Method                                    & \texttt{pub1}  & \texttt{pub2} & \texttt{pub3}     & \texttt{\textbf{Avg.}}  & \texttt{seq1}      & \texttt{seq2}      & \texttt{seq3} & \texttt{\textbf{Avg.}} \\
			\midrule
			NICE-SLAM~\cite{zhu2022nice} & 5.27 & 3.64 & 2.96 & 3.96 & 16.38 & 15.61 & 9.84  & 13.94 \\
			Vox-Fusion~\cite{yang2022vox} & 9.90 & 2.95 & 3.83 & 5.56 & 11.10 & 52.01 & 20.01 & 27.71 \\
			CoSLAM~\cite{wang2023co} & 6.32 & 1.58 & 2.75 & 3.55 & 12.97 & 7.63  & \snd 7.15  & 9.25  \\
			ESLAM~\cite{johari2023eslam} & \snd 4.71 & \fs 1.29 & \snd 2.44 & \snd 2.81 & 11.58 & \fs 6.49  & 8.06  & \snd 8.71  \\
			Point-SLAM~\cite{sandstrom2023point} & 5.13 & 3.16 & 2.56 & 3.62 & \snd 10.52 & 9.18  & 235.79  & 85.16  \\
			Ours-NeRF & \fs 3.53 & \snd 1.42 & \fs 2.08 & \fs 2.34 & \fs 9.75  & \snd 7.32  & \fs 5.54  & \fs 7.54  \\
			\midrule
			\midrule
			SplaTAM~\cite{keetha2024splatam} & 4.89 & 3.29 & 2.58 & 3.59 & 33.56 & 14.51 & 7.26 & 18.44 \\
            \add{RTG-SLAM}~\cite{peng2024rtgslam} & \add{21.83} & \add{15.23} & \add{-} & \add{-} & \add{8.97} & \add{\ding{54}} & \add{\ding{54}}  & \add{-} \\
            % \add{GS-ICP}~\cite{ha2024gsicp} & - & - & - & - & - & - & -  & - \\
            \add{Photo-SLAM}~\cite{hhuang2024photoslam} & \add{\fs 3.89} & \add{3.47} & \add{2.36} & \add{\snd 3.24} & \add{\fs 8.64} & \add{96.30} & \add{\fs 2.49}  & \add{35.81} \\
            \add{MonoGS}~\cite{Matsuki:Murai:etal:CVPR2024} & \add{5.52} & \add{\snd 3.07} & \add{\snd 2.28} & \add{3.62} & \add{8.92} & \add{\snd 6.13} & \add{9.94}  & \add{\snd 8.33} \\
			Ours-GS & \snd 4.16 & \fs 2.78 & \fs 2.16 & \fs 3.03 & \snd 8.79 & \fs 5.56 & \snd 3.27 & \fs 5.87 \\
			\bottomrule
		\end{tabular}
	}
	\vspace{-1ex}
\end{table}

\subsection{Evaluation on Blur Datasets}
\label{exp_eval_blur}

% \label{exp_ablation}

\begin{figure*}[!ht]
	\setlength\tabcolsep{1pt}
	\centering
	\begin{tabular}{ccc}
		%		\rot{Input} &
		&\raisebox{-0.02in}{\rotatebox[origin=t]{90}{\scriptsize Input}}
		& \includegraphics[valign=m,width=0.99\textwidth]{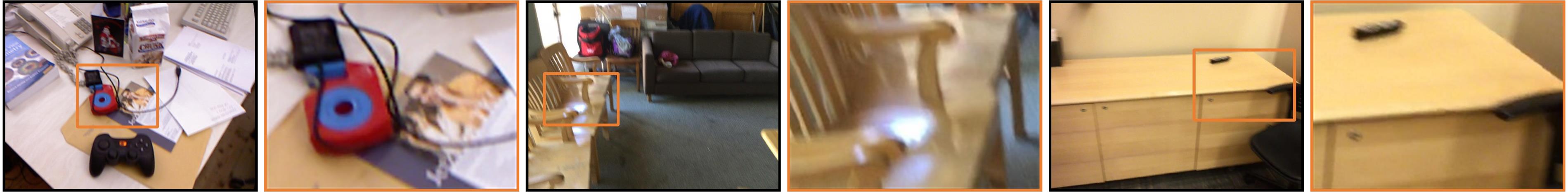}\\

        \specialrule{0em}{.1em}{.1em}
		&\raisebox{-0.02in}{\rotatebox[origin=t]{90}{\scriptsize CoSLAM\cite{wang2023co}}}
		& \includegraphics[valign=m,width=0.99\textwidth]{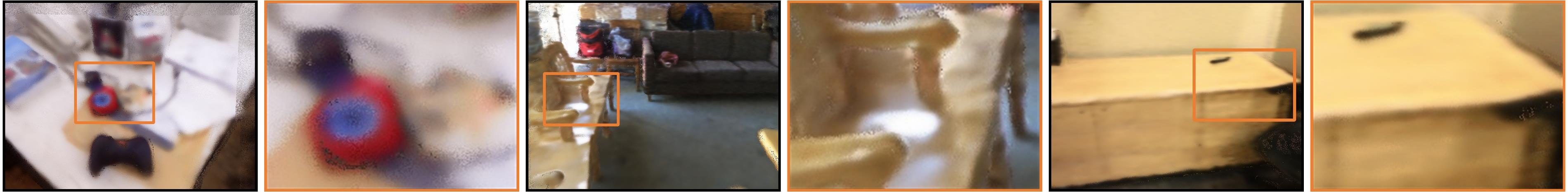}\\

        \specialrule{0em}{.1em}{.1em}
		&\raisebox{-0.02in}{\rotatebox[origin=t]{90}{\scriptsize ESLAM\cite{johari2023eslam}}}
		& \includegraphics[valign=m,width=0.99\textwidth]{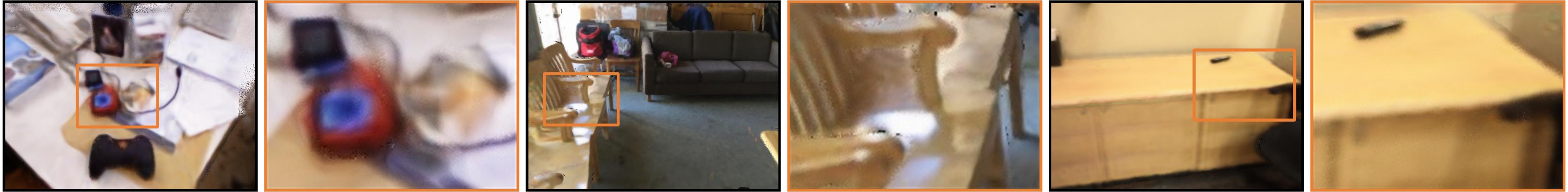}\\

        \specialrule{0em}{.1em}{.1em}
        \raisebox{-0.035in}{\rotatebox[origin=t]{90}{\scriptsize Point}}
		&\raisebox{-0.02in}{\rotatebox[origin=t]{90}{\scriptsize SLAM\cite{sandstrom2023point}}}
		& \includegraphics[valign=m,width=0.99\textwidth]{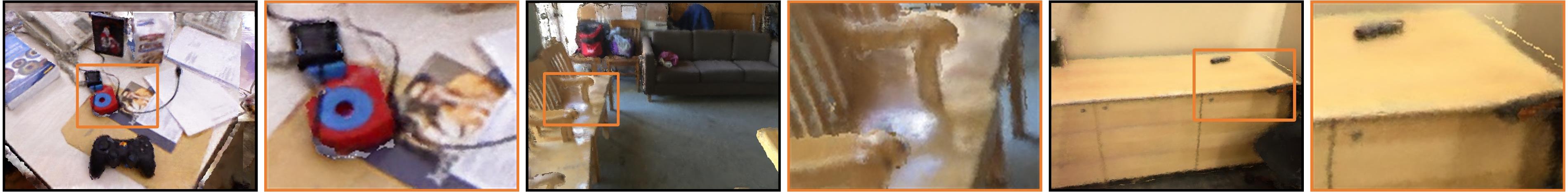}\\

        \specialrule{0em}{.1em}{.1em}
        \raisebox{-0.035in}{\rotatebox[origin=t]{90}{\scriptsize Spla}}
		&\raisebox{-0.02in}{\rotatebox[origin=t]{90}{\scriptsize TAM\cite{keetha2024splatam}}}
		& \includegraphics[valign=m,width=0.99\textwidth]{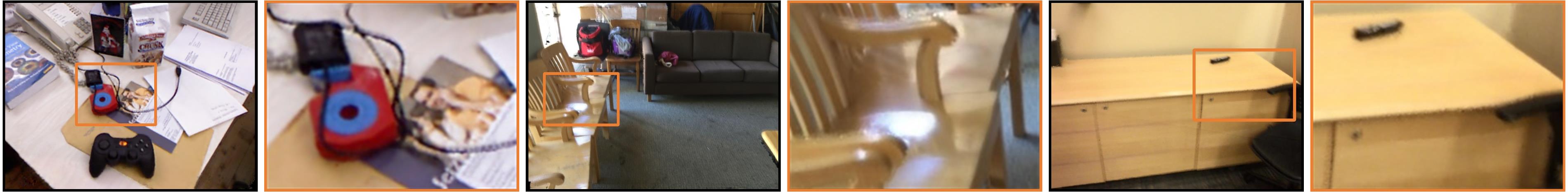}\\

        \specialrule{0em}{.1em}{.1em}
        \raisebox{-0.035in}{\rotatebox[origin=t]{90}{\scriptsize \add{RTG}}}
		&\raisebox{-0.02in}{\rotatebox[origin=t]{90}{\scriptsize \add{SLAM}\cite{peng2024rtgslam}}}
		& \includegraphics[valign=m,width=0.99\textwidth]{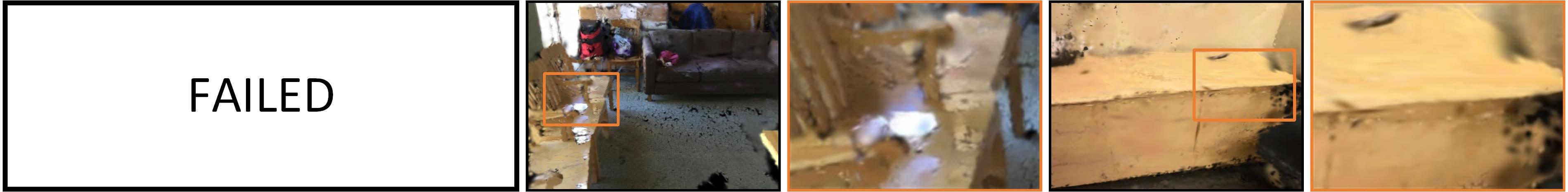}\\

        \specialrule{0em}{.1em}{.1em}
        \raisebox{-0.035in}{\rotatebox[origin=t]{90}{\scriptsize \add{Photo}}}
		&\raisebox{-0.02in}{\rotatebox[origin=t]{90}{\scriptsize \add{SLAM}\cite{hhuang2024photoslam}}}
		& \includegraphics[valign=m,width=0.99\textwidth]{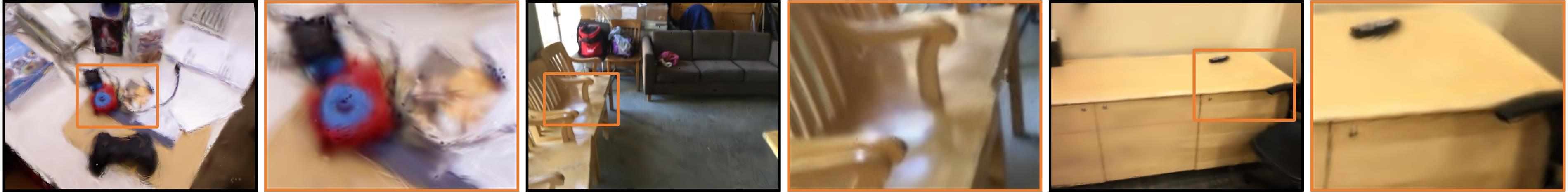}\\

        \specialrule{0em}{.1em}{.1em}
		&\raisebox{-0.02in}{\rotatebox[origin=t]{90}{\scriptsize \add{MonoGS}\cite{Matsuki:Murai:etal:CVPR2024}}}
		& \includegraphics[valign=m,width=0.99\textwidth]{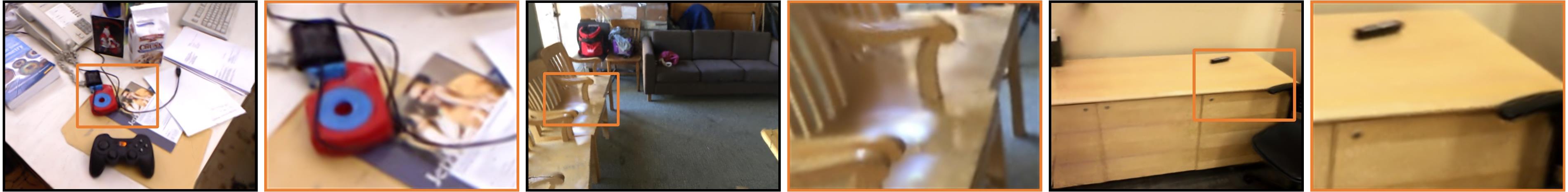}\\

        \specialrule{0em}{.1em}{.1em}
		&\raisebox{-0.02in}{\rotatebox[origin=t]{90}{\scriptsize Ours-GS}}
		& \includegraphics[valign=m,width=0.99\textwidth]{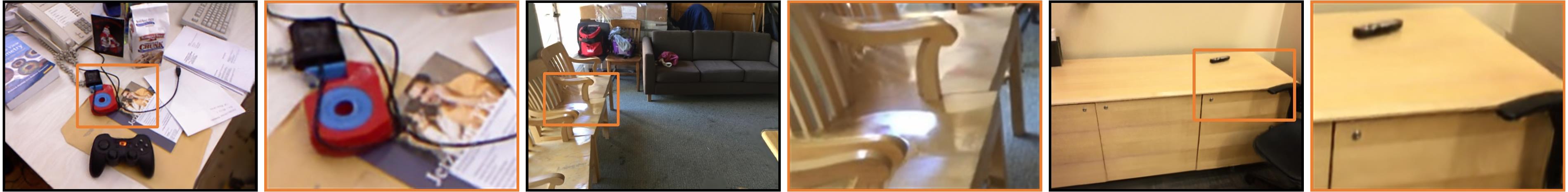}\\
		
	\end{tabular}
	% \vspace{-0.5em}
	\caption{\add{{\textbf{Qualitative rendering results of different methods with the real public ScanNet and TUM RGB-D datasets.}} The experimental results demonstrate that our method achieves superior performance over prior methods on the real public dataset.}}
	\label{fig:tum}
	% \vspace{-1ex}
\end{figure*}

\begin{figure*}[!ht]
	\setlength\tabcolsep{1pt}
	\centering
	\begin{tabular}{ccc}
		%		\rot{Input} &
		&\raisebox{-0.02in}{\rotatebox[origin=t]{90}{\scriptsize Input}}
		& \includegraphics[valign=m,width=0.99\textwidth]{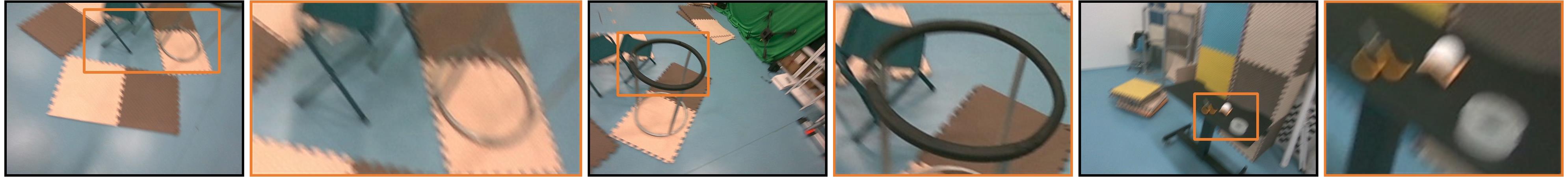}\\

        \specialrule{0em}{.1em}{.1em}
		&\raisebox{-0.02in}{\rotatebox[origin=t]{90}{\scriptsize CoSLAM\cite{wang2023co}}}
		& \includegraphics[valign=m,width=0.99\textwidth]{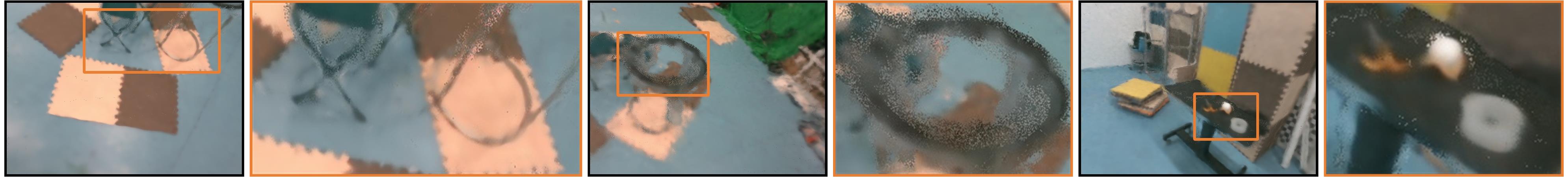}\\

        \specialrule{0em}{.1em}{.1em}
		&\raisebox{-0.02in}{\rotatebox[origin=t]{90}{\scriptsize ESLAM\cite{johari2023eslam}}}
		& \includegraphics[valign=m,width=0.99\textwidth]{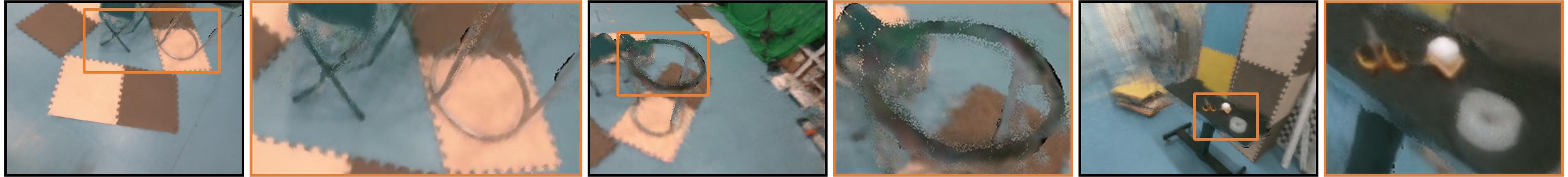}\\

        \specialrule{0em}{.1em}{.1em}
        \raisebox{-0.035in}{\rotatebox[origin=t]{90}{\scriptsize Point}}
		&\raisebox{-0.02in}{\rotatebox[origin=t]{90}{\scriptsize SLAM\cite{sandstrom2023point}}}
		& \includegraphics[valign=m,width=0.99\textwidth]{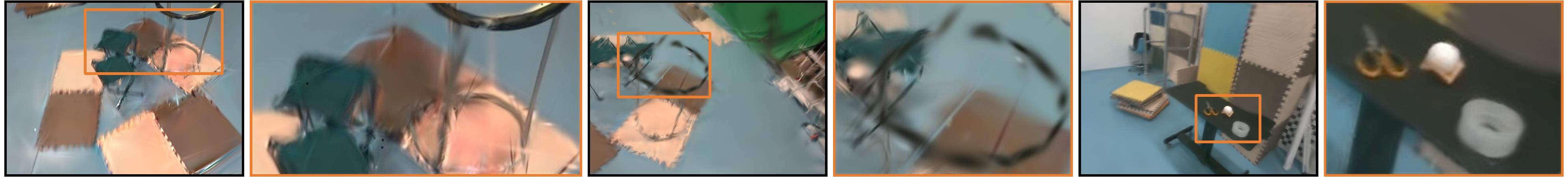}\\

        \specialrule{0em}{.1em}{.1em}
        \raisebox{-0.035in}{\rotatebox[origin=t]{90}{\scriptsize Spla}}
		&\raisebox{-0.02in}{\rotatebox[origin=t]{90}{\scriptsize TAM\cite{keetha2024splatam}}}
		& \includegraphics[valign=m,width=0.99\textwidth]{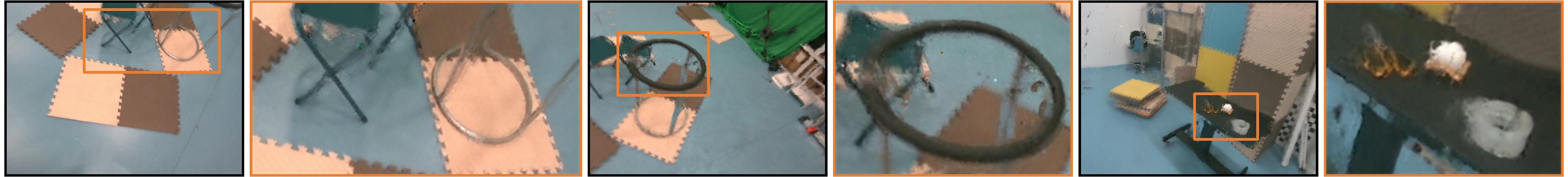}\\

        \specialrule{0em}{.1em}{.1em}
        \raisebox{-0.035in}{\rotatebox[origin=t]{90}{\scriptsize \add{RTG}}}
		&\raisebox{-0.02in}{\rotatebox[origin=t]{90}{\scriptsize \add{SLAM}\cite{peng2024rtgslam}}}
		& \includegraphics[valign=m,width=0.99\textwidth]{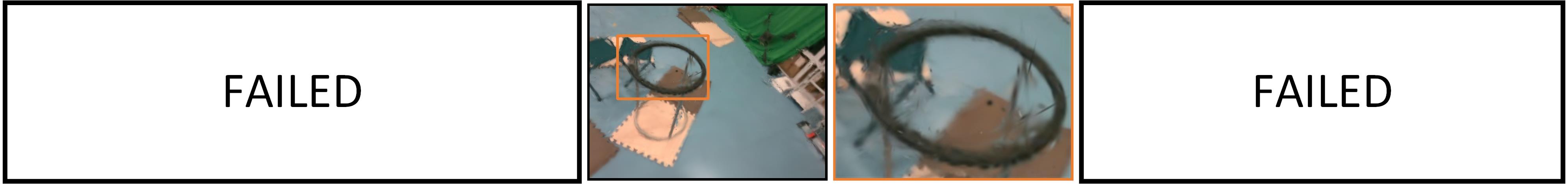}\\

        \specialrule{0em}{.1em}{.1em}
        \raisebox{-0.035in}{\rotatebox[origin=t]{90}{\scriptsize \add{Photo}}}
		&\raisebox{-0.02in}{\rotatebox[origin=t]{90}{\scriptsize \add{SLAM}\cite{hhuang2024photoslam}}}
		& \includegraphics[valign=m,width=0.99\textwidth]{figs/real_data/realsense/real_photo.jpg}\\

        \specialrule{0em}{.1em}{.1em}
		&\raisebox{-0.02in}{\rotatebox[origin=t]{90}{\scriptsize \add{MonoGS}\cite{Matsuki:Murai:etal:CVPR2024}}}
		& \includegraphics[valign=m,width=0.99\textwidth]{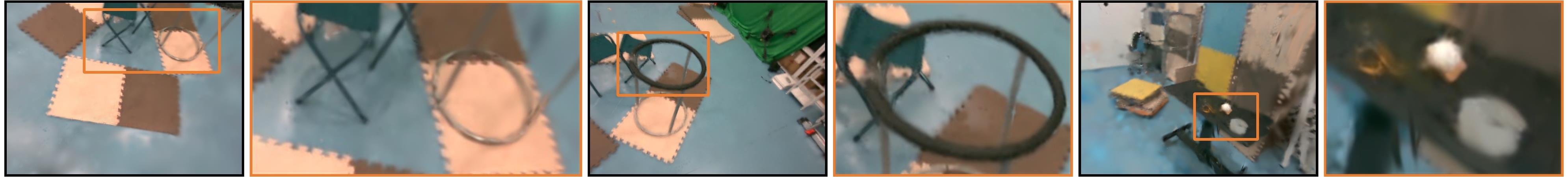}\\

        \specialrule{0em}{.1em}{.1em}
		&\raisebox{-0.02in}{\rotatebox[origin=t]{90}{\scriptsize Ours-GS}}
		& \includegraphics[valign=m,width=0.99\textwidth]{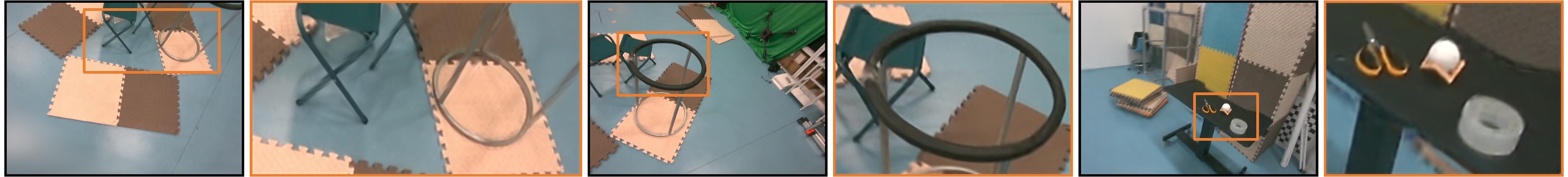}\\
		
	\end{tabular}
	% \vspace{-0.5em}
	\caption{\add{{\textbf{Qualitative rendering results of different methods with our real captured Realsense datasets.}} The experimental results demonstrate that our method achieves superior performance over prior methods on the real captured dataset as well. \add{Note that the first column shows an incorrect view rendered by Photo-SLAM, attributed to a failure in tracking.}}}
	\label{fig:realsense}
	\vspace{-1ex}
\end{figure*}

\PAR{Evaluation on Synthetic Blur Dataset: ArchViz.}

\textbf{\textit{Tracking:}} The results in \tabnref{tab:tracking_ArchViz} illustrate that the tracking performance of our method outperforms other state-of-the-art NeRF-SLAM and Gaussian-SLAM systems on the synthetic motion blur ArchViz datasets provided by~\cite{mba-vo}. Additionally, it reveals that pure implicit SLAMs, such as CoSLAM~\cite{wang2023co} and ESLAM~\cite{johari2023eslam}, perform better than methods based on explicit points or Gaussians, \ie~Point-SLAM~\cite{sandstrom2023point} and SplaTAM~\cite{keetha2024splatam}, when faced with challenging motion-blurred sequences. \figref{fig:mba_vo_traj_archviz} demonstrates the estimated trajectories of \mbavo2 on the motion-blurred sequences from the ArchViz dataset. Both quantitative and qualitative results highlight the effectiveness of our proposed algorithm for handling motion-blurred image sequences.

\textbf{\textit{Rendering:}} \tabnref{tab:psnr_archviz} presents the quantitative rendering results (in the left three columns) on the ArchViz dataset. The results demonstrate that our method significantly surpasses other state-of-the-art methods, thanks to the physical motion blur image formation model. The qualitative results shown in \figref{fig:mbavo_img} illustrate that our method can recover and render high-quality images from motion-blurred sequences, particularly in areas with rich textures (\eg~green leaves) and sharp edges.

\textbf{\textit{Reconstruction:}} Since the ArchViz dataset does not provide ground truth meshes, we utilize TSDF Fusion~\cite{curless1996volumetric} to extract reference meshes from the ground truth camera poses, sharp images, and depth data. We only report the reconstruction metrics for ArchViz1 as a reference (in the right column of \tabnref{tab:psnr_archviz}) because there are many unseen areas in ArchViz2 and ArchViz3, which should be excluded from the extracted mesh using the 'unseen points' file that we do not have when computing metrics with the ground truth mesh. The visualization results in \figref{fig:mbavo_mesh} demonstrate that \mbavo2 achieves a satisfying reconstructed mesh with clear boundaries and rich details.

\PAR{Evaluation on Real Blur Dataset.}

\textbf{\textit{Tracking:}} We conduct further experiments on real captured motion blur datasets, and the quantitative tracking results are presented in \tabnref{tab:tracking_real_blur}. `pub1-pub3' refer to the selected motion-blurred sequences from the public ScanNet~\cite{dai2017scannet} and TUM RGB-D~\cite{sturm2012benchmark} datasets, while `seq1-seq3' denote our collected datasets using a Realsense RGB-D camera. \mbavo2 achieves best tracking performance on average in both public and our captured datasets.

\textbf{\textit{Rendering:}} \figref{fig:tum} and \figref{fig:realsense} present qualitative examples of rendering from public datasets and our captured dataset. Although other state-of-the-art methods (e.g., Point-SLAM~\cite{sandstrom2023point} and SplaTAM~\cite{keetha2024splatam}) can achieve impressive performance on images that are not severely blurred, they struggle to restore sharp images and often introduce unpleasant artifacts in severely blurred cases. In contrast, our method consistently delivers robust performance regardless of the level of motion blur and effectively captures fine details (e.g., sofa, desk, chair legs, etc.).

\begin{table}[t]
	% \vspace{-2.5ex}
	% \aboverulesep=-
%	\centering
	\vspace{-0ex}
	\caption{\add{\textbf{Tracking comparison} (ATE RMSE [cm]) of the proposed method vs. the SOTA methods on the \texttt{\#Replica dataset}. The \colorbox{Green!30}{\textbf{best}} and \colorbox{Orange!30}{second-best} results are colored.}}
    \vspace{-1em}
	\label{tab:tracking_replica}
	\centering
	\resizebox{0.48\textwidth}{!}{
		\begin{tabular}{l|ccccccccc}
			\toprule
			Method                                    & \texttt{Rm0}      & \texttt{Rm1}      & \texttt{Rm2}      & \texttt{Of0}     & \texttt{Of1}     & \texttt{Of2}     & \texttt{Of3}     & \texttt{Of4}     & \texttt{\textbf{Avg.}} \\
			\midrule
			iMAP~\cite{sucar2021imap} & 3.12 & 2.54 & 2.31 & 1.69 & 1.03 & 3.99 & 4.05 & 1.93 & 2.58 \\
			NICE-SLAM~\cite{zhu2022nice} & 1.69 & 2.04 & 1.55 & 0.99 & 0.90 & 1.39 & 3.97 & 3.08 & 1.95 \\
			Vox-Fusion~\cite{yang2022vox} & 0.59 & 1.08 & 0.81 & 6.32 & 1.02 & 0.98 & \snd 0.67 & 0.95 & 1.55 \\
			CoSLAM~\cite{wang2023co} & 0.70 & 0.95 & 1.35 & 0.59 &  \snd  0.55 & 2.03 & 1.56 & 0.72 & 1.00 \\
			ESLAM~\cite{johari2023eslam} & 0.71 & 0.70 & \snd 0.52 & 0.57 &  \snd  0.55 & \snd 0.58 & 0.72 & \snd 0.63 & 0.63 \\
			Point-SLAM~\cite{sandstrom2023point} & \snd 0.56 & \snd 0.47 & \fs 0.30 & \snd 0.35 & 0.62 &  \fs 0.55 & 0.72 & 0.73 & \snd 0.54 \\
			Ours-NeRF & \fs 0.34 & \fs 0.43 & \fs 0.30 & \fs 0.25 & \fs 0.33 & 0.68 & \fs 0.48 & \fs 0.46 & \fs 0.41 \\
			\midrule
			\midrule
			GS-SLAM~\cite{yan2024gs} & 0.48 & 0.53 & 0.33 & 0.52 & 0.41 & 0.59 & 0.46 & 0.70 & 0.50 \\
			SplaTAM~\cite{keetha2024splatam} & 0.31 & 0.40 & 0.29 & 0.47 & \snd 0.27 & 0.29 & 0.32 & 0.55 & 0.36 \\
			\add{RTG-SLAM}~\cite{peng2024rtgslam} & \add{\fs 0.18} & \add{\snd 0.19} & \add{\fs 0.10} & \add{\fs 0.16} & \add{\fs 0.13} & \add{\fs 0.22} & \add{\snd 0.23} & \add{\fs 0.24} & \add{\fs 0.18} \\
            % \add{GS-ICP}~\cite{ha2024gsicp} & 0.18 & 0.21 & 0.16 & - &  0.15 &  0.18 & 0.19 & 0.24 & - \\
            \add{Photo-SLAM}~\cite{hhuang2024photoslam} & \add{0.33} & \add{0.34} & \add{\snd 0.19} & \add{0.43} & \add{0.33} & \add{0.87} & \add{0.40} & \add{\snd 0.53} & \add{0.43} \\
            \add{MonoGS}~\cite{Matsuki:Murai:etal:CVPR2024} & \add{0.44} & \add{0.32} & \add{0.31} & \add{0.44} & \add{0.52} & \add{\snd 0.23} & \add{\fs 0.17} & \add{2.25} & \add{0.58} \\
			Ours-GS &  \snd 0.25 & \fs 0.13 & 0.25 & \snd 0.32 & 0.30 & 0.74 & 0.26 & 0.58 & \snd 0.35 \\
			\bottomrule
		\end{tabular}
	    }
		% \vspace{-1.9em}
\end{table}

\begin{table}[t]
    % \vspace{-3ex}
    \centering
    \caption{\add{\textbf{Rendering Performance on Replica}. While Point-SLAM outperforms other NeRF-based SLAM methods, it suffers from significantly longer mapping optimization times — over 30 times slower than CoSLAM, ESLAM, and our NeRF-based system, as shown in~\tabnref{tab:memory_runtime}.}}
    \vspace{-1em}
    \label{tab:replica_rendering}
    \resizebox{0.48\textwidth}{!}{
        \setlength{\tabcolsep}{2pt}
        \renewcommand{\arraystretch}{1}
        \begin{tabular}{c|r|rrrrrrrrrr}
            \toprule
            Method & Metric~ & \texttt{Rm0}      & \texttt{Rm1}      & \texttt{Rm2}      & \texttt{Of0}     & \texttt{Of1}     & \texttt{Of2}     & \texttt{Of3}     & \texttt{Of4}     & \textbf{Avg.} & \textbf{FPS}\\
            \midrule
            \multirow{3}{*}{\begin{tabular}[l]{@{}c@{}}NICESL\\ AM~\cite{zhu2022nice}\end{tabular}}    & \scriptsize PSNR $\uparrow$ & 22.12 & 22.47 & 24.52 & 29.07 &  30.34 &  19.66 &  22.23 &  24.94 &  24.42 & \\
			& \scriptsize SSIM $\uparrow$ &  0.689 &  0.757 &  0.814 &  0.874 &  0.886 &  0.797 &  0.801 &  0.856 &  0.809 & 0.30\\
			& \scriptsize LPIPS $\downarrow$ &  0.330 &  0.271 & \snd 0.208 &  0.229 & \snd 0.181 &  \snd 0.235 &  0.209 & 0.198 & 0.233 & \\
            \midrule
            \multirow{3}{*}{\begin{tabular}[l]{@{}c@{}}VoxFus\\ ion~\cite{yang2022vox}\end{tabular}} & \scriptsize PSNR $\uparrow$ & 22.39 & 22.36 & 23.92 & 27.79 & 29.83 & 20.33 & 23.47 & 25.21 & 24.41 & \fs \\
		    & \scriptsize SSIM $\uparrow$ & 0.683 & 0.751 & 0.798 & 0.857 & 0.876 & 0.794 & 0.803 & 0.847 & 0.801 & \fs 3.88 \\
		    & \scriptsize LPIPS $\downarrow$ & 0.303 &  0.269 &  0.234 & 0.241 & 0.184 & 0.243 & 0.213 &  0.199  &  0.236 & \fs \\
            \midrule
            \multirow{3}{*}{\begin{tabular}[l]{@{}c@{}}CoSL\\ AM~\cite{wang2023co}\end{tabular}}    & \scriptsize PSNR $\uparrow$ & 27.27 & 28.45 & 29.06 & 34.14 & 34.87 & \snd 28.43 & 28.76 & 30.91 & 30.24 & \snd \\
			& \scriptsize SSIM $\uparrow$ & 0.910 & 0.909 & 0.932 & 0.961 & \snd 0.969 & 0.938 & 0.941 & 0.955 & 0.939 & \snd 3.68 \\
			& \scriptsize LPIPS $\downarrow$ & 0.324 & 0.294 & 0.266 &  0.209 & 0.196             & 0.258 & 0.229 & 0.236 & 0.252 & \snd \\
            \midrule
            \multirow{3}{*}{\begin{tabular}[l]{@{}c@{}}ESL\\AM~\cite{johari2023eslam}\end{tabular}}    & \scriptsize PSNR $\uparrow$ & 25.32 &  27.77 & 29.08 &  33.71 & 30.20 & 28.09 & 28.77 &  29.71 &  29.08 & \\
			& \scriptsize SSIM $\uparrow$ &  0.875 &  0.902 & 0.932 &  0.960 &  0.923 & \snd 0.943 & 0.948 &  0.945 &  0.929 & 2.82 \\
			& \scriptsize LPIPS $\downarrow$ &  0.313 &  0.298 & 0.248 & 0.184 & 0.228 & 0.241 & 0.196 & 0.204 & 0.336 & \\
            \midrule
            \multirow{3}{*}{\begin{tabular}[c]{@{}c@{}}\add{Point}\\\add{SLAM}~\cite{sandstrom2023point}\end{tabular}}
            & \add{\scriptsize PSNR}~$\uparrow$ & \add{\fs{32.40}} & \add{\fs{34.08}} & \add{\fs{35.50}} & \add{\fs{38.26}} & \add{\fs{39.16}} & \add{\fs{33.99}} & \add{\fs{33.48}} & \add{\fs{33.49}} & \add{\fs{35.05}} & \add{} \\
            & \add{\scriptsize SSIM}~$\uparrow$ & \add{\fs{0.974}} & \add{\fs{0.977}} & \add{\fs{0.982}} & \add{\fs{0.983}} & \add{\fs{0.986}} & \add{\fs{0.960}} & \add{\fs{0.960}} & \add{\fs{0.979}} & \add{\fs{0.975}} & \add{1.33} \\
            & \add{\scriptsize LPIPS}~$\downarrow$ & \add{\fs{0.113}} & \add{\fs{0.116}} & \add{\fs{0.111}} & \add{\fs{0.100}} & \add{\fs{0.118}} & \add{\fs{0.156}} & \add{\fs{0.132}} & \add{\fs{0.142}} & \add{\fs{0.124}} & \add{} \\
            \midrule
            \multirow{3}{*}{\begin{tabular}[c]{@{}c@{}}Ours\\NeRF\end{tabular}}
            & \scriptsize PSNR $\uparrow$ & \snd 28.20 & \snd 29.84 & \snd 29.92 & \snd 35.27 & \snd 35.03 & 27.03 & \snd 29.49 & \snd 32.01 & \snd 30.85 & \\
			& \scriptsize SSIM $\uparrow$ &  \snd 0.926 & \snd 0.926 & \snd 0.938 & \snd 0.969 &  0.968 & 0.926 & \snd 0.956 & \snd 0.963 & \snd 0.947 & 2.75 \\
			& \scriptsize LPIPS $\downarrow$ &  \snd 0.265 & \snd 0.262 & 0.244 & \snd 0.167 & 0.182 & 0.281 & \snd 0.182 & \snd 0.178 & \snd 0.220 & \\
            
            \midrule
            % \hdashline
            \midrule
            \multirow{3}{*}{\begin{tabular}[c]{@{}c@{}}GSSL\\AM~\cite{yan2024gs}\end{tabular}}
            & \scriptsize PSNR $\uparrow$ & 31.56 & 32.86 & 32.59 &  38.70 &  41.17 &  32.36  &  32.03 &  32.92 &  34.27 &  \\
			& \scriptsize SSIM $\uparrow$ &  0.968 &  0.973 & 0.971 &  0.986 &  \snd 0.993 &  0.978 &  0.970 &  0.968 &  0.975 &  386.90 \\
			& \scriptsize LPIPS $\downarrow$ &  0.094 &  0.075 & 0.093 &  \snd 0.050 &  \snd0.033 &  0.094 &  0.110 &  0.112 &  0.082 &  \\
            \midrule
            \multirow{3}{*}{\begin{tabular}[c]{@{}c@{}}SplaT\\ AM~\cite{keetha2024splatam}\end{tabular}}
            & \scriptsize PSNR $\uparrow$ &  32.34 &  33.66 &  35.34 & 38.39 & 39.33 & 32.18 & 30.27 & 32.37 & 34.24 &     \\
			& \scriptsize SSIM $\uparrow$ & \snd 0.974 & 0.969 &  0.983 & 0.982 & 0.983 & 0.968 & 0.953 & 0.950 & 0.970 & 234.83 \\
			& \scriptsize LPIPS $\downarrow$ &  0.073 & 0.097 &  \snd 0.069 & 0.080 & 0.091 & 0.096 & 0.119 & 0.148 & 0.097 & \\
            \midrule
            \multirow{3}{*}{\begin{tabular}[c]{@{}c@{}}\add{RTG}\\ \add{SLAM}~\cite{peng2024rtgslam}\end{tabular}}
            & \add{\scriptsize PSNR}~$\uparrow$ & \add{25.53} & \add{29.91} & \add{31.19} & \add{36.65} & \add{37.19} & \add{29.88} & \add{29.96} & \add{33.11} & \add{31.68} & \add{} \\
            & \add{\scriptsize SSIM}~$\uparrow$ & \add{0.900} & \add{0.919} & \add{0.937} & \add{0.984} & \add{0.985} & \add{0.958} & \add{0.959} & \add{0.952} & \add{0.949} & \add{456.89} \\
            & \add{\scriptsize LPIPS}~$\downarrow$ & \add{0.269} & \add{0.212} & \add{0.193} & \add{0.096} & \add{0.108} & \add{0.187} & \add{0.172} & \add{0.158} & \add{0.174} & \add{} \\
            \midrule
            %
   %          \multirow{3}{*}{\begin{tabular}[c]{@{}c@{}}\add{GS-ICP}\\SLAM~\cite{ha2024gsicp}\end{tabular}}
   %          & \scriptsize PSNR $\uparrow$ & 33.64 & 30.72 & 36.49 & \redtext{41.41} & 42.21 & 35.94 & 35.23 & 38.08 & 36.72 &     \\
			% & \scriptsize SSIM $\uparrow$ & 0.972 & 0.949 & 0.977 & \redtext{0.98}  & 0.995 & 0.982 & 0.987 & 0.975 & 0.977 & - \\
			% & \scriptsize LPIPS $\downarrow$ & 0.061 & 0.117 & 0.076 & \redtext{0.040} & 0.026 & 0.054 & 0.040 & 0.063 & 0.060 & \\
   %          \midrule
            %
            \multirow{3}{*}{\begin{tabular}[c]{@{}c@{}}\add{Photo}\\\add{SLAM}~\cite{hhuang2024photoslam}\end{tabular}}
            & \add{\scriptsize PSNR}~$\uparrow$ & \add{31.68} & \add{\snd 35.50} & \add{36.78} & \add{38.97} & \add{39.78} & \add{\snd 33.94} & \add{\snd 34.26} & \add{\fs 36.77} & \add{35.96} & \add{\fs}   \\
			& \add{\scriptsize SSIM}~$\uparrow$ & \add{0.948} & \add{0.970} & \add{0.987} & \add{0.984} & \add{0.988} & \add{0.975} & \add{0.973} & \add{0.969} & \add{0.974} & \add{\fs 1084.00} \\
			& \add{\scriptsize LPIPS}~$\downarrow$ & \add{0.100} & \add{0.080} & \add{0.061} & \add{0.076} & \add{0.079} & \add{0.126} & \add{0.095} & \add{0.078} & \add{0.087} & \add{\fs} \\
            \midrule
            \multirow{3}{*}{\begin{tabular}[c]{@{}c@{}}\add{Mono}\\\add{GS}~\cite{Matsuki:Murai:etal:CVPR2024}\end{tabular}}
            & \add{\scriptsize PSNR}~$\uparrow$ & \add{\snd{34.55}} & \add{\fs{36.72}} & \add{\snd{37.28}} & \add{\snd{40.06}} & \add{\snd{41.57}} & \add{\fs{35.13}} & \add{\fs{35.71}} & \add{36.03} & \add{\fs{37.13}} &  \add{\snd}   \\
			& \add{\scriptsize SSIM}~$\uparrow$ & \add{0.969} & \add{\snd{0.982}} & \add{\snd{0.989}} & \add{\snd{0.991}} & \add{0.989} & \add{\snd{0.990}} & \add{\snd{0.987}} & \add{\snd{0.982}} & \add{\snd{0.985}} & \add{\snd 769.00} \\
			& \add{\scriptsize LPIPS}~$\downarrow$ & \add{\snd{0.071}} & \add{\snd{0.069}} & \add{0.071} & \add{0.051} & \add{0.073} & \add{\snd{0.073}} & \add{\snd{0.079}} & \add{\snd{0.064}} & \add{\snd{0.069}} & \add{\snd} \\
            \midrule
            \multirow{3}{*}{\begin{tabular}[c]{@{}c@{}}Ours\\ GS\end{tabular}}  & \scriptsize PSNR $\uparrow$ & \fs 35.68 & 34.75 & \fs 38.63 & \fs 40.86 & \fs 42.26 & 33.26 & 33.69 & \snd 36.69 & \snd 36.98 &   \\
			& \scriptsize SSIM $\uparrow$ & \fs 0.989 & \fs 0.980 & \fs 0.994 & \fs 0.993 & \fs 0.995 & \fs 0.988 & \fs 0.990 & \fs 0.991 & \fs 0.990 & 497.63 \\
			& \scriptsize LPIPS $\downarrow$ & \fs 0.034 & \fs 0.064 & \fs 0.030 & \fs 0.025 & \fs 0.029 & \fs 0.074 & \fs 0.041 & \fs 0.049 & \fs 0.043 & \\
            \bottomrule
        \end{tabular}
    }
    \vspace{-1em}
\end{table}

\begin{table}[t]
    \vspace{-3.0ex}
    \centering
    \caption{\add{\textbf{Mesh reconstruction} comparison of the proposed method vs. the SOTA methods on \texttt{\#Replica} dataset.}}
    \vspace{-1em}
    \label{tab:recon_replica}
    \resizebox{0.48\textwidth}{!}{
        \setlength{\tabcolsep}{2pt}
        \renewcommand{\arraystretch}{1}
        \begin{tabular}{c|l|rrrrrrrrr}
            \toprule
            Method & Metric & \texttt{Rm0}      & \texttt{Rm1}      & \texttt{Rm2}      & \texttt{Of0}     & \texttt{Of1}     & \texttt{Of2}     & \texttt{Of3}     & \texttt{Of4}     & \textbf{Avg.} \\
            \midrule
            \multirow{4}{*}{\begin{tabular}[l]{@{}c@{}}NICESL\\AM~\cite{zhu2022nice}\end{tabular}}    & Depth L1 $\downarrow$ & 1.92  & 1.37   & 1.83  & 1.47  & 1.50  & 2.68  & 2.84  & 1.95 & 1.95 \\
            & Precision $\uparrow$  & 48.74 & 57.51  & 53.91 & 61.72 & 65.57 & 47.65 & 52.68 & 44.92 & 54.09 \\
            & Recall $\uparrow$ & 38.35 & 42.04 & 41.21 & 47.14 & 50.22 & 39.53 & 42.01 & 36.66 & 42.15 \\
            & F1 $\uparrow$ & 42.92 & 48.57  & 46.71 & 53.45 & 56.88 & 43.21 & 46.74 & 40.37 & 47.36 \\
            \midrule
            \multirow{4}{*}{\begin{tabular}[l]{@{}c@{}}VoxFus\\ion~\cite{yang2022vox}\end{tabular}}    & Depth L1 $\downarrow$ &   0.55  &  0.63  & 1.00 & 2.96  & 1.30   & \snd 1.25  &  1.37  &  0.76  & 1.23   \\
            & Precision $\uparrow$  &  \snd 95.03 &  92.65 & 85.76  & 46.04 & 83.88 &  87.98 &  90.10 & 86.37 & 83.48 \\
            & Recall $\uparrow$ & 75.58 & 69.41 & 68.37  & 41.64 & 65.22 & 68.47 & 71.10  & 66.42 & 65.78 \\
            & F1 $\uparrow$ & 84.59 & 79.37  & 76.09  & 43.13 & 73.39 &  77.01 & 79.83 & 75.09 & 73.56 \\
            \midrule
            \multirow{4}{*}{\begin{tabular}[l]{@{}c@{}}CoSLA\\M~\cite{wang2023co}\end{tabular}}    & Depth L1 $\downarrow$ & 0.87  & 0.91  & 2.26  & 1.17  & 1.25  & 2.13  & 2.30 & 1.51  & 1.55  \\
            & Precision $\uparrow$  & 90.30  & 83.18 & 80.68 & 85.81 &  92.84 & 67.64 & 69.23  &  87.77 & 82.18 \\
            & Recall $\uparrow$ & 74.60 & 65.59 & 67.30  & 73.78 & 76.12 & 55.80  & 58.55  & 70.20  & 67.74  \\
            & F1 $\uparrow$ & 81.70 & 73.35 & 73.38 & 79.34 & 83.65 & 61.15 & 63.45  & 77.98 & 74.25  \\
            \midrule
            \multirow{4}{*}{\begin{tabular}[l]{@{}c@{}}ESL\\AM~\cite{johari2023eslam}\end{tabular}}    & Depth L1 $\downarrow$ & 0.58  &   0.72  &  0.85  &  0.92  &   1.03 &   1.56  &  1.49  &  0.76  &  0.99 \\
            & Precision $\uparrow$  &  94.24 & 87.45 &  91.69 &  93.57 & 92.31 & \snd 90.83 &  90.23 & 87.71 &  91.00 \\
            & Recall $\uparrow$ &  \snd 84.31 &  82.15 &  82.02 &   87.23 &  81.68 &  \snd 79.43 &  78.39 &  77.26 &  81.56 \\
            & F1 $\uparrow$ & \snd 88.99 &  84.78 &  86.59 &  90.29 &  86.17 &  \snd 84.18 &  82.20 &  82.05 &  85.66 \\
            \midrule
            \multirow{4}{*}{\begin{tabular}[c]{@{}c@{}}\add{PointSL}\\\add{AM}~\cite{sandstrom2023point}\end{tabular}}    & \add{Depth L1}~$\downarrow$ & \add{\snd 0.53} & \add{\fs 0.22} & \add{\fs 0.46} & \add{\fs 0.30} & \add{\fs 0.57} & \add{\fs 0.49} & \add{\fs 0.51} & \add{\fs 0.46} & \add{\fs 0.44} \\
            & \add{Precision}~$\uparrow$ & \add{91.95} & \add{\fs 99.04} & \add{\fs 97.89} & \add{\fs 99.00} & \add{\fs 99.37} & \add{\fs 98.05} & \add{\fs 96.61} & \add{\fs 93.98} & \add{\fs 96.99} \\
            & \add{Recall}~$\uparrow$ & \add{82.48} & \add{\fs 86.43} & \add{\fs 84.64} & \add{\fs 89.06} & \add{\fs 84.99} & \add{\fs 81.44} & \add{\fs 81.17} & \add{\snd 78.51} & \add{\fs 83.59} \\
            & \add{F1}~$\uparrow$ & \add{86.90} & \add{\fs 92.31} & \add{\fs 90.78} & \add{\fs 93.77} & \add{\fs 91.62} & \add{\fs 88.98} & \add{\fs 88.22} & \add{\snd 85.55} & \add{\fs 89.77} \\
            \midrule
            \multirow{4}{*}{\begin{tabular}[c]{@{}c@{}}Ours\\NeRF\end{tabular}}    & Depth L1 $\downarrow$ & \fs 0.52  & \snd 0.37  & \snd 0.71  & \snd 0.56  & \snd 0.92  &  1.68  & \snd 1.12  & \snd 0.51  & \snd 0.80  \\
            & Precision $\uparrow$ & \fs 96.08 & \snd 96.31 & \snd 93.11 &  \snd 95.03 & \snd 95.20 &  87.34 & \snd 90.59 & \snd 93.66 & \snd 93.42  \\
            & Recall $\uparrow$ & \fs 85.41  & \snd 85.68 & \snd 82.13 & \snd 88.72 & \snd 83.72  &  74.60  & \snd 78.63 & \fs 80.55 & \snd 82.43 \\
            & F1 $\uparrow$ & \fs 90.42 & \snd 90.69 & \snd 87.28 & \snd 91.77 &  \snd 89.08 &  79.78 & \snd 84.10  & \fs 86.58 & \snd 87.46 \\
            
            \midrule
            % \hdashline
            \midrule
            \multirow{4}{*}{\begin{tabular}[c]{@{}c@{}}Spla\\TAM~\cite{keetha2024splatam}\end{tabular}}  & Depth L1 $\downarrow$ & \snd 0.49  & \fs 0.37  & \snd 0.59  & \snd 0.43  & \fs 0.63  & \fs 0.97  & \snd 1.15  & 0.71  & \fs 0.67  \\
            & Precision $\uparrow$  & \snd 96.08 & \fs 95.66 & \snd 95.58 & \snd 97.43 & \snd 98.03 & \fs 94.62 & 86.49 & 89.26 & \snd 94.14 \\
            & Recall $\uparrow$ & \snd 83.68 & 83.63 & \snd 83.16 & \snd 87.64 & \snd 83.27 & \fs 79.29 & 74.06 & 75.05 & \snd 81.22 \\
            & F1 $\uparrow$ & \snd 89.11 & \snd 88.64 & \snd 88.94 & \snd 92.27 & \snd 90.05 & \fs 86.55 & 80.22 & 81.54 & \snd 87.17 \\
            \midrule
            \multirow{4}{*}{\begin{tabular}[c]{@{}c@{}}\add{RTGS}\\\add{LAM}~\cite{peng2024rtgslam}\end{tabular}}    & \add{Depth L1}~$\downarrow$ & \add{0.79} & \add{0.85} & \add{1.49} & \add{0.98} & \add{1.12} & \add{1.97} & \add{1.78} & \add{0.92} & \add{1.24} \\
            & \add{Precision}~$\uparrow$ & \add{79.75} & \add{83.49} & \add{78.79} & \add{81.66} & \add{82.40} & \add{76.44} & \add{70.12} & \add{77.43} & \add{78.76}  \\
            & \add{Recall}~$\uparrow$ & \add{79.46} & \add{75.38} & \add{71.52} & \add{75.20} & \add{71.93} & \add{68.34} & \add{63.72} & \add{69.86} & \add{71.93} \\
            & \add{F1}~$\uparrow$ & \add{80.05} & \add{79.21} & \add{75.25} & \add{78.29} & \add{75.58} & \add{72.66} & \add{66.68} & \add{72.51} & \add{75.03} \\
            %
            %
            % \midrule
            % \multirow{4}{*}{\begin{tabular}[c]{@{}c@{}}\add{GS}\\ICP~\cite{ha2024gsicp}\end{tabular}}    & Depth L1 $\downarrow$ & -  & -  & -  & -  & -  & -  & -  & -  & - \\
            % & Precision $\uparrow$ & -  & -  & -  & -  & - & -  & -  & - & -  \\
            % & Recall $\uparrow$ & -  & -  & -  & -  & -  & -  & -  & - & -  \\
            % & F1 $\uparrow$ & -  & -  & -  & -  & -  & -  & -  &  - & -  \\
            %
            %
            \midrule
            \multirow{4}{*}{\begin{tabular}[c]{@{}c@{}}\add{PhotoSL}\\\add{AM}~\cite{hhuang2024photoslam}\end{tabular}}    & \add{Depth L1}~$\downarrow$ & \add{0.87}    & \add{0.41}    & \add{0.98}    & \add{0.45}    & \add{0.72}    & \add{\snd 1.21}    & \add{1.34}    & \add{\snd 0.64}    & \add{0.83} \\
            & \add{Precision}~$\uparrow$ & \add{85.89}   & \add{94.20}   & \add{91.59}   & \add{93.84}   & \add{96.04}   & \add{92.96}   & \add{88.07}   & \add{\snd 91.63}   & \add{91.78} \\
            & \add{Recall}~$\uparrow$ & \add{81.74}   & \add{\snd 83.71}   & \add{80.12}   & \add{87.09}   & \add{83.24}   & \add{77.82}   & \add{75.32}   & \add{\snd 75.86}   & \add{80.61} \\
            & \add{F1}~$\uparrow$ & \add{82.31}   & \add{89.23}   & \add{85.27}   & \add{91.98}   & \add{90.16}   & \add{84.72}   & \add{82.04}   & \add{\snd 82.45}   & \add{86.02}  \\
            \midrule
            \multirow{4}{*}{\begin{tabular}[c]{@{}c@{}}\add{Mono}\\\add{GS}~\cite{Matsuki:Murai:etal:CVPR2024}\end{tabular}}    & \add{Depth L1}~$\downarrow$ & \add{0.69}    & \add{\snd 0.39}    & \add{0.61}    & \add{0.55}    & \add{\snd 0.67}    & \add{1.39}    & \add{\fs 0.96}    & \add{0.68}    & \add{\snd 0.74} \\
            & \add{Precision}~$\uparrow$ & \add{88.24}   & \add{94.93}   & \add{95.31}   & \add{93.14}   & \add{97.75}   & \add{94.04}   & \add{\snd 90.17}   & \add{86.85}   & \add{92.55}  \\
            & \add{Recall}~$\uparrow$ & \add{81.56}   & \add{83.31}   & \add{82.93}   & \add{84.35}   & \add{83.04}   & \add{78.68}   & \add{\snd 76.09}   & \add{73.82}   & \add{80.47}  \\
            & \add{F1}~$\uparrow$ & \add{85.40}   & \add{88.74}   & \add{88.38}   & \add{88.53}   & \add{89.93}   & \add{\snd 86.12}   & \add{\snd 82.86}   & \add{79.81}   & \add{86.22}  \\
            \midrule
            \multirow{4}{*}{\begin{tabular}[c]{@{}c@{}}Ours\\GS\end{tabular}}  & Depth L1 $\downarrow$ & \fs 0.47  & \snd 0.39  & \fs 0.56  & \fs 0.35  & \fs 0.63  & 1.46  & 1.51 & \fs 0.55  & \snd 0.74  \\
            & Precision $\uparrow$  & \fs 96.63 & \snd 95.31 & \fs 96.64 & \fs 98.26 & \fs 98.37 & \snd 94.27 & \fs 94.17 & \fs 94.02 & \fs 95.96 \\
            & Recall $\uparrow$ & \fs 84.05 & \fs 83.96 & \fs 83.59 & \fs 88.27 & \fs 83.66 & \snd 79.07 & \fs 79.24 & \fs 78.81 & \fs 82.58 \\
            & F1 $\uparrow$ & \fs 89.90 & \fs 89.28 & \fs 89.64 & \fs 93.01 & \fs 90.42 & 86.00 & \fs 86.06 & \fs 85.74 & \fs 88.76 \\
            \bottomrule
        \end{tabular}
    }
    \vspace{-1ex}
\end{table}

\begin{figure*}
	\setlength\tabcolsep{1pt}
	\centering
	\begin{tabular}{ccc}
		%		\rot{Input} &
		\raisebox{-0.035in}{\rotatebox[origin=t]{90}{\scriptsize NICE}}
		&\raisebox{-0.02in}{\rotatebox[origin=t]{90}{\scriptsize SLAM\cite{zhu2022nice}}}
		& \includegraphics[valign=m,width=0.90\textwidth]{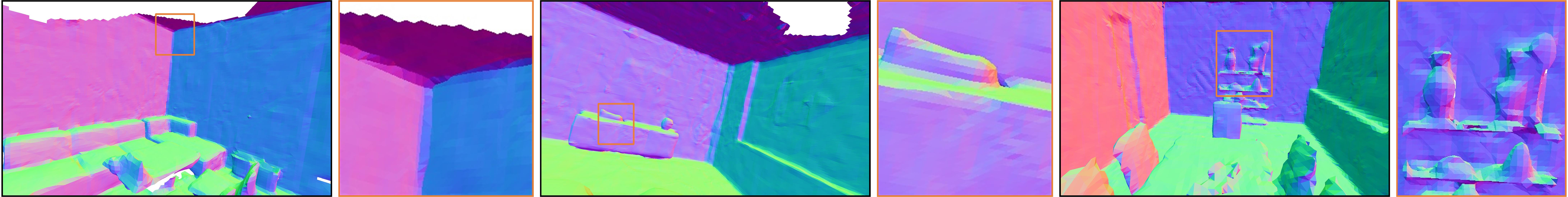}\\
		
		% \specialrule{0em}{.1em}{.1em}
		% \raisebox{-0.035in}{\rotatebox[origin=t]{90}{\scriptsize VoxFu}}
		% &\raisebox{-0.02in}{\rotatebox[origin=t]{90}{\scriptsize sion\cite{yang2022vox}}}
		% & \includegraphics[valign=m,width=0.86\textwidth]{figs/replica/mesh/vox.jpg}\\
		
		\specialrule{0em}{.1em}{.1em}
		&\raisebox{-0.02in}{\rotatebox[origin=t]{90}{\scriptsize CoSLAM\cite{wang2023co}}}
		& \includegraphics[valign=m,width=0.90\textwidth]{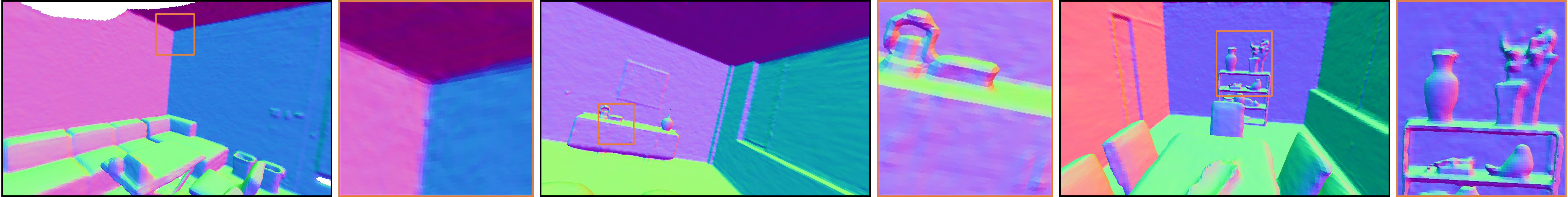}\\
		
		\specialrule{0em}{.1em}{.1em}
		&\raisebox{-0.02in}{\rotatebox[origin=t]{90}{\scriptsize ESLAM\cite{johari2023eslam}}}
		& \includegraphics[valign=m,width=0.90\textwidth]{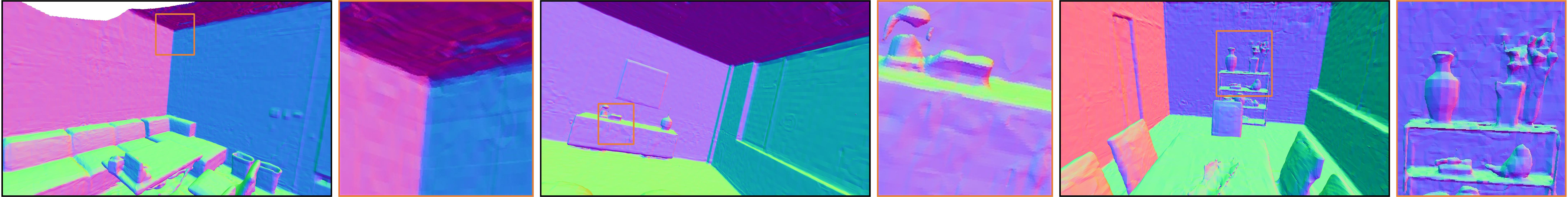}\\

        \specialrule{0em}{.1em}{.1em}
		\raisebox{-0.035in}{\rotatebox[origin=t]{90}{\scriptsize \add{Point}}}
		&\raisebox{-0.02in}{\rotatebox[origin=t]{90}{\scriptsize \add{SLAM}\cite{sandstrom2023point}}}
		& \includegraphics[valign=m,width=0.90\textwidth]{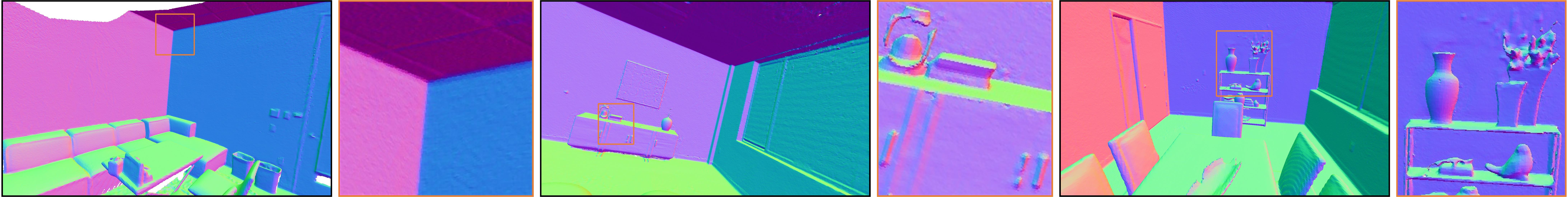}\\
		
		\specialrule{0em}{.1em}{.1em}
		&\raisebox{-0.02in}{\rotatebox[origin=t]{90}{\scriptsize Ours-NeRF}}
		& \includegraphics[valign=m,width=0.90\textwidth]{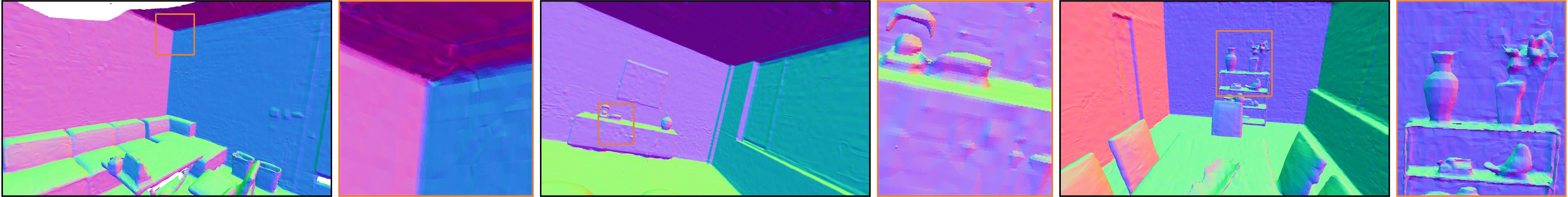}\\
		
		\specialrule{0em}{.1em}{.1em}
		&\raisebox{-0.02in}{\rotatebox[origin=t]{90}{\scriptsize SplaTAM\cite{keetha2024splatam}}}
		& \includegraphics[valign=m,width=0.90\textwidth]{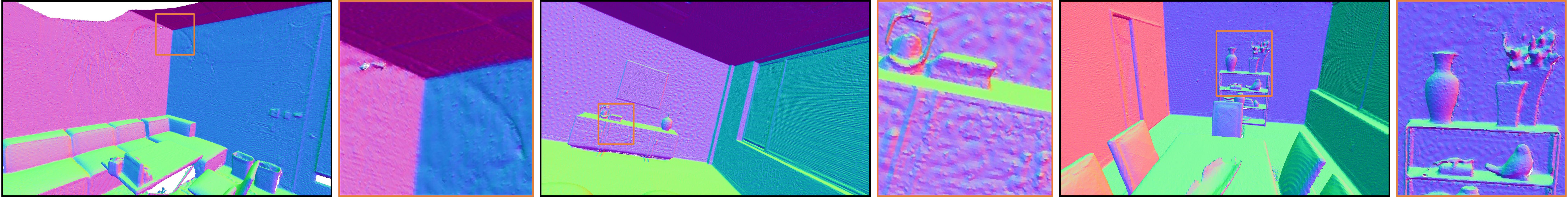}\\

        \specialrule{0em}{.1em}{.1em}
		\raisebox{-0.035in}{\rotatebox[origin=t]{90}{\scriptsize \add{RTG}}}
		&\raisebox{-0.02in}{\rotatebox[origin=t]{90}{\scriptsize \add{SLAM}\cite{peng2024rtgslam}}}
		& \includegraphics[valign=m,width=0.90\textwidth]{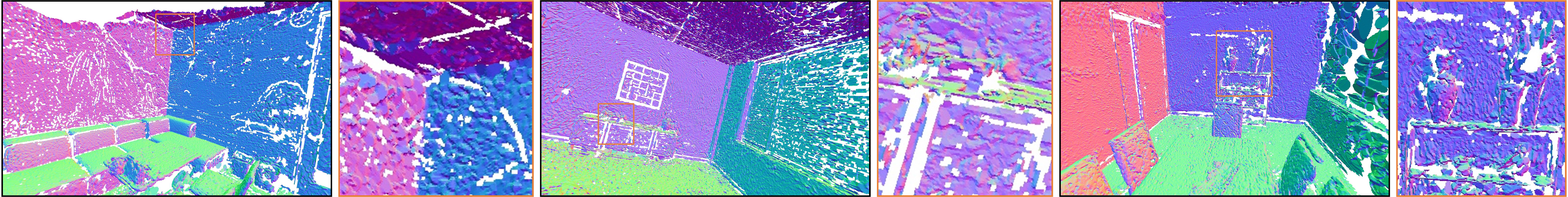}\\

        \specialrule{0em}{.1em}{.1em}
		\raisebox{-0.035in}{\rotatebox[origin=t]{90}{\scriptsize \add{Photo}}}
		&\raisebox{-0.02in}{\rotatebox[origin=t]{90}{\scriptsize \add{SLAM}\cite{hhuang2024photoslam}}}
		& \includegraphics[valign=m,width=0.90\textwidth]{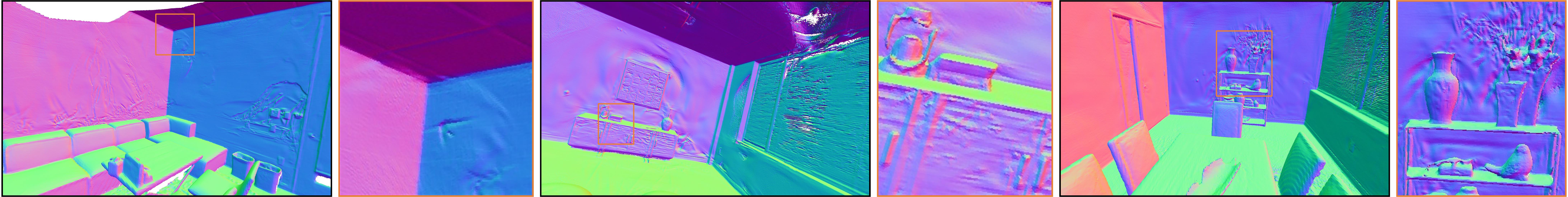}\\

        \specialrule{0em}{.1em}{.1em}
		&\raisebox{-0.02in}{\rotatebox[origin=t]{90}{\scriptsize \add{MonoGS}\cite{Matsuki:Murai:etal:CVPR2024}}}
		& \includegraphics[valign=m,width=0.90\textwidth]{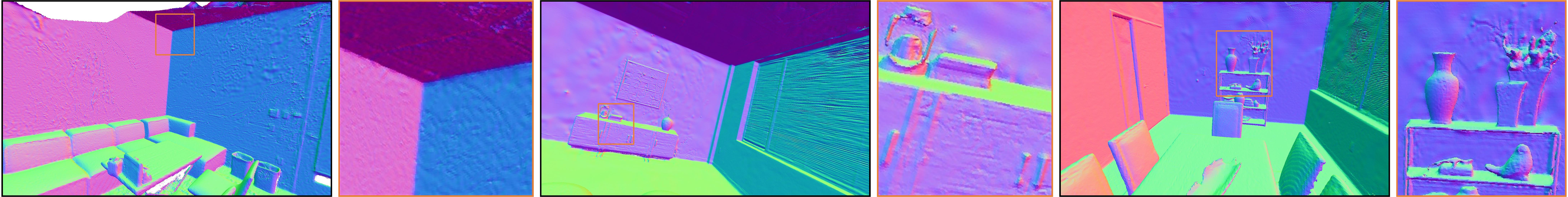}\\
		
		\specialrule{0em}{.1em}{.1em}
		&\raisebox{-0.02in}{\rotatebox[origin=t]{90}{\scriptsize Ours-GS}}
		& \includegraphics[valign=m,width=0.90\textwidth]{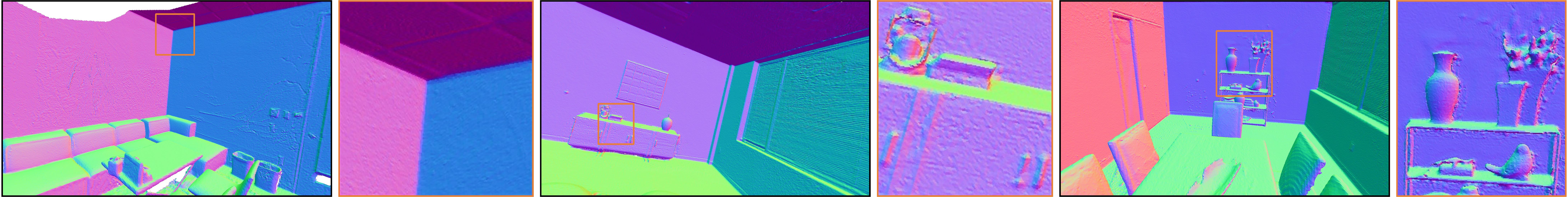}\\
		
		\specialrule{0em}{.1em}{.1em}
		&\raisebox{-0.02in}{\rotatebox[origin=t]{90}{\scriptsize Groundtruth}}
		& \includegraphics[valign=m,width=0.90\textwidth]{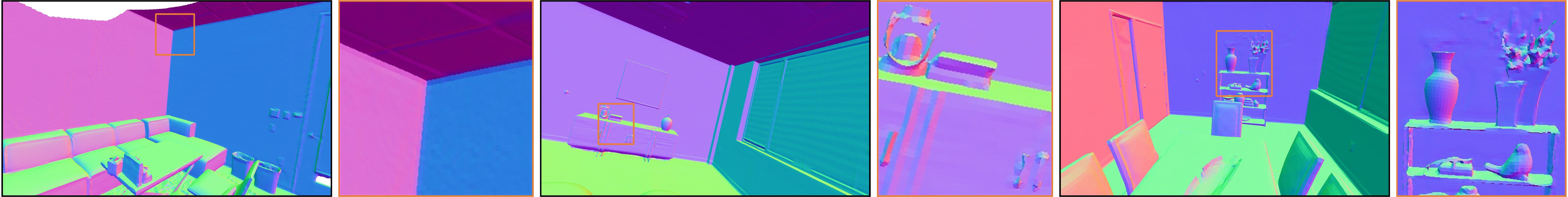}\\
		
	\end{tabular}
	% \vspace{-0.5em}
	\caption{\add{\textbf{Qualitative mesh visualization results of different methods with Replica datasets.} It demonstrates \mbavo2 surpasses other state-of-the-art dense visual SLAMs even on standard sharp datasets. Note that RTG-SLAM produces incomplete meshes due to the presence of numerous holes in the rendered depth maps.}}
	\label{fig:replica_mesh}
	\vspace{-1ex}
\end{figure*}

%
% \begin{table}[t] 
% 	% \vspace{-2.5ex}
% 	\centering
% 	\vspace{-0ex}
% 	\caption{Tracking comparison (ATE RMSE [cm]) of the proposed method vs. the SOTA methods on the ScanNet dataset.}
% 	\label{tab:tracking_scannet}
% 	\resizebox{0.48\textwidth}{!}{
% 		\begin{tabular}{l|ccccccc}
% 			\toprule
% 			Method                                    & \texttt{0000}      & \texttt{0059}      & \texttt{0106}      & \texttt{0169}     & \texttt{0181}     & \texttt{0207}    & \texttt{\textbf{Avg.}} \\
% 			\midrule
% 			iMAP~\cite{sucar2021imap} & 55.95 & 32.06 & 17.50 & 70.51 & 32.10 & 11.91 & 36.67 \\
% 			%
% 			NICE-SLAM~\cite{zhu2022nice} & 8.64 & 12.25 & 8.09 & 10.28 & 12.93 &  5.59 & 9.63 \\
% 			%
% 			Vox-Fusion~\cite{yang2022vox} & 8.39 & 8.95 & 8.41 & 9.50 & 12.20 & 6.43 & 8.98 \\
% 			%
% 			CoSLAM~\cite{wang2023co} &  7.18 & 12.29 & 9.57 &  6.62 &    13.43 & 7.13 & 9.37 \\
% 			%
% 			ESLAM~\cite{johari2023eslam} &  7.84 & 9.24 &  7.82 & 6.78 &     9.35 & 5.88 &  7.82 \\
% 			%
% 			Point-SLAM~\cite{sandstrom2023point} & 10.24 &  7.81 & 8.65 &  22.16 & 14.77 & 9.54 & 12.19 \\
% 			%
% 			\midrule
% 			SplaTAM~\cite{keetha2024splatam} & 12.83 & 10.10 & 17.72 & 12.08 & 11.10 & 7.46 & 11.88 \\
% 			\midrule
% 			Ours & 9.23 &  6.68 &  7.54 &  6.22 &   9.62 &  4.93 &  7.37 \\
% 			\bottomrule
% 		\end{tabular}
% 	}
% 	\vspace{-3ex}
% \end{table}

\begin{table}[t] 
	% \vspace{-1.0ex}
	\centering
	% \vspace{-2.5ex}
	\caption{\add{\textbf{Tracking comparison} (ATE RMSE [cm]) of the proposed method vs. the SOTA methods on the real-world \texttt{\#TUM: left} and \texttt{\#ScanNet: right} dataset. \mbavo2 achieves better tracking performance. "-" indicates GS-SLAM is not open sourced and we can not get the results.}}
    \vspace{-1em}
	\label{tab:tracking_scannet}
	\resizebox{0.48\textwidth}{!}{
		\begin{tabular}{l|cccc|cccc}
			\toprule
			Method                                    & \texttt{desk}  & \texttt{xyz} & \texttt{off}     & \texttt{\textbf{Avg.}}  & \texttt{0059}      & \texttt{0106}      & \texttt{0169} & \texttt{\textbf{Avg.}} \\
			\midrule
			NICE-SLAM~\cite{zhu2022nice} & 4.26 & 31.73 & 3.87 & 13.29 & 12.25 &  8.09 & 10.28 & 10.21  \\
			Vox-Fusion~\cite{yang2022vox} & 3.52 & 1.49 & 26.01 & 10.34 & \snd 8.95 & 8.41 & 9.50  &  8.95  \\
			CoSLAM~\cite{wang2023co} & 2.70 & 1.90 & 2.60 & 2.40 & 12.29 & 9.57 & \snd 6.62 & 9.49    \\
			ESLAM~\cite{johari2023eslam} & \fs 2.47 & \snd 1.11 & \fs 2.42 &  \fs 2.00 & 9.24 & \snd 7.82 &  6.78 & \snd 7.95  \\
			Point-SLAM~\cite{sandstrom2023point} & 4.34 & 1.31 & 3.48 & 3.04 &  7.81 & 8.65 &  22.16 & 12.87  \\
                Ours-NeRF & \snd 2.49 & \fs 1.05 & \snd 2.58 & \snd 2.04 & \fs 6.68 &  \fs 7.54 & \fs 6.22 &  \fs 6.81  \\
                \midrule
			\midrule
                GS-SLAM~\cite{yan2024gs} & 3.20 & 1.30 & 6.60 & 3.70 & - & - & - & - \\
			SplaTAM~\cite{keetha2024splatam} & 3.35 &  1.24 &  5.16 &  3.25 &  10.10 &  17.72 & \snd 12.08 & 13.30  \\
            \add{RTG-SLAM}~\cite{peng2024rtgslam} & \add{\snd 1.66} & \add{\snd 0.38} & \add{\snd 1.13} & \add{\fs 1.06} & \add{11.57} & \add{14.85} & \add{19.34} & \add{15.25}  \\
            %
            % \add{GS-ICP}~\cite{ha2024gsicp} & 2.56 & 1.79 & 2.68 & 2.34 & - & - & - & -  \\
            %
            \add{Photo-SLAM}~\cite{hhuang2024photoslam} & \add{2.53} & \add{\fs 0.32} & \add{\fs 0.99} & \add{\snd 1.28} & \add{\fs 7.12} & \add{\fs 6.96} & \add{\fs 8.05} & \add{\fs 7.38}  \\
            \add{MonoGS}~\cite{Matsuki:Murai:etal:CVPR2024} & \add{\fs 1.50} & \add{1.44} & \add{1.49} & \add{1.47} & \add{7.76} & \add{\snd 10.91} & \add{15.58} & \add{11.42}  \\
                Ours-GS & 2.11 & 1.37 & 2.46 & 1.98 & \snd 7.57 & 11.71 & 13.54 & \snd 10.94 \\
			\bottomrule
		\end{tabular}
	}
	\vspace{-3ex}
\end{table}

\subsection{Evaluation on Standard Sharp Datasets}
\label{exp_eval_sharp}
To demonstrate that our method can also handle commonly used standard sharp datasets, we conduct the following experiments without modeling the blur process in either our tracker or mapper, following the exact same settings as in prior works by setting the number of virtual images in \eqnref{eq_blur_im_formation} to 1.

\PAR{Evaluation on Synthetic Dataset: Replica.}

\textbf{\textit{Tracking}}: The quantitative tracking results are presented in \tabnref{tab:tracking_replica}. \add{Our method outperforms most previous state-of-the-art approaches, including both NeRF-based SLAMs (e.g., NICE-SLAM~\cite{zhu2022nice}, CoSLAM~\cite{wang2023co}, ESLAM~\cite{johari2023eslam}, and Point-SLAM~\cite{sandstrom2023point}) and Gaussian Splatting-based SLAMs (e.g., GS-SLAM~\cite{yan2024gs}, SplaTAM~\cite{keetha2024splatam}, Photo-SLAM~\cite{hhuang2024photoslam} and MonoGS~\cite{Matsuki:Murai:etal:CVPR2024}, except RTG-SLAM~\cite{peng2024rtgslam}), in terms of both tracking accuracy and running speed (as shown in \tabnref{tab:memory_runtime})}. Notably, our CUDA-implemented tracker operates in real time, with the running speed primarily limited by the back-end mapper. The tracking results for \texttt{office2} in \tabnref{tab:tracking_replica} are not as strong as those for the other sequences due to the reduced number of features in \texttt{office2}, which hampers our feature-based tracker.

\textbf{\textit{Rendering}}: We also compare the rendering performance of \mbavo2 with dense visual SLAMs in \tabnref{tab:replica_rendering}. \add{The results demonstrate that \mbavo2 achieves the second-best performance after Point-SLAM, outperforming the third-best implicit dense visual SLAM by 0.6 PSNR in the Our-NeRF version and delivering superior or comparable results to other 3DGS-SLAM approaches.}

\textbf{\textit{Reconstruction:}} The quantitative evaluations of 3D mesh reconstruction against other state-of-the-art dense visual SLAMs are presented in \tabnref{tab:recon_replica}. \add{\mbavo2 surpasses all other SLAM systems except Point-SLAM in Precision, Recall, and F1 metrics on average, achieving comparable Depth L1 performance with SplaTAM~\cite{keetha2024splatam} and MonoGS~\cite{Matsuki:Murai:etal:CVPR2024}}. Qualitative mesh visualizations are shown in \figref{fig:replica_mesh}, where we observe that CoSLAM~\cite{wang2023co} and ESLAM~\cite{johari2023eslam} produce plausible geometry, but lack detail, resulting in over-smoothed mesh outcomes. \add{In contrast, SplaTAM~\cite{keetha2024splatam}, Photo-SLAM~\cite{hhuang2024photoslam} and MonoGS~\cite{Matsuki:Murai:etal:CVPR2024} deliver more details, albeit with increased artifacts and protrusions.} Compared to these methods, \mbavo2 provides high-fidelity reconstruction results with more accurate geometric details and fewer artifacts.

\PAR{Evaluation on Real Datasets: ScanNet and TUM RGB-D.}

\textbf{\textit{Tracking}}: \add{The quantitative tracking results in \tabnref{tab:tracking_scannet} demonstrate that \mbavo2 achieves average camera localization performance comparable to other state-of-the-art methods. Furthermore, our NeRF version outperforms all other approaches on the ScanNet dataset.} The “-” symbol indicates that GS-SLAM is not open-sourced, so we are unable to report the results.

\textbf{\textit{Rendering}}: Since the public ScanNet and TUM RGB-D datasets are captured with motion blur, we conduct rendering experiments while modeling blur, as shown in \figref{fig:tum}.

% runtime
\begin{table}[t]
    \vspace{-1.5ex}
    \centering
    %\footnotesize
    \caption{\add{\textbf{Runtime and memory} usage on Replica \texttt{\#Room0}. The decoder parameters and embedding denote the parameter number of MLPs and the memory usage of the scene representation. Ours-NeRF achieves 22.57 FPS, while the speed of Ours-GS is slower than GS-SLAM but faster than SlaTAM. Our tracking achieves best performance in Radiance Fields and Gaussian Splatting approaches, respectively.}}
    % Note that Point-SLAM uses extra memory dynamic radius to improve performance (mark as $^\dagger$).
    \vspace{-2.5ex}
    \scriptsize
    \setlength{\tabcolsep}{5pt}
    \resizebox{\columnwidth}{!}{
        \begin{tabular}{l|ccccccccc}
            \toprule
            \multirow{2}{*}{Method} & Tracking & Mapping & Keyframe & System & ATE & Scene \\
            & [ms$\times$it] $\downarrow$ & [ms$\times$it] $\downarrow$ & Number $\downarrow$ & FPS~$\uparrow$ & [cm]~$\downarrow$ & Embedding$\downarrow$ \\
            \midrule
            NICE-SLAM~\cite{zhu2022nice} & 6.64 $\times$ 10 & 28.63 $\times$ 60 & \snd 400 & 2.91 & 1.69 & 48.48 MB \\
            % Vox-Fusion~\cite{yang2022vox} & 0.03 $\times$ 30  & 66.53 $\times$ 10 & 1.28 & 1.28 & 0.59 & 1.49 MB \\
            CoSLAM~\cite{wang2023co} & 6.01 $\times$ 10 & \fs 13.18 $\times$ 10    & \snd 400 & \fs 16.64 & 0.70 & \fs 6.36 MB \\
            ESLAM~\cite{johari2023eslam} & \snd 6.85 $\times$ 8 & \snd 19.87 $\times$ 15 & 500 & 13.42 & 0.71 & \snd 27.12 MB \\
            Point-SLAM~\cite{sandstrom2023point} & 4.36 $\times$ 40            & 34.81 $\times$ 300 & 417 & 0.42 & \snd 0.56 & 12508.62 MB \\
            Ours-NeRF & \fs 34.48 $\times$ 1 & 19.96 $\times$ 15 & \fs 296 & \fs 22.57 & \fs 0.34 & \snd 27.12 MB \\
            \midrule
            GS-SLAM~\cite{yan2024gs} & 11.9 $\times$ 10  & \snd 12.8 $\times$ 100   & -- & \snd 8.34 & 0.48 & 198.04 MB \\
            SplaTAM~\cite{keetha2024splatam} & 36.25 $\times$ 40  & 35.33 $\times$ 60 & 2000 & 0.47 & 0.34 & 253.34 MB \\
            % \add{Photo-SLAM}~\cite{hhuang2024photoslam} & 54.17 $\times$ 1  & 26.56 $\times$ 100 & 308 & 1.63 & 0.42 & 193.71 MB \\
            \add{RTG-SLAM}\cite{peng2024rtgslam} & \add{60.93~$\times$~1} & \add{\fs 4.88~$\times$~100} & \add{400} & \add{\fs 9.90} & \add{\fs 0.18} & \add{\fs 52.26~MB} \\
            \add{Photo-SLAM}~\cite{Matsuki:Murai:etal:CVPR2024} & \add{\snd 41.18~$\times$~1} & \add{19.40~$\times$~100} & \add{\snd 156} & \add{6.61} & \add{0.33} & \add{\snd 59.88~MB}\\
            \add{MonoGS}~\cite{Matsuki:Murai:etal:CVPR2024} & \add{2.54~$\times$~100}  & \add{26.56~$\times$~100} & \add{308} & \add{1.63} & \add{0.42} & \add{193.71~MB} \\
            Ours-GS & \fs 34.23 $\times$ 1 & 75.63 $\times$ 100 & \fs 134 &  1.97 & \snd 0.25 & 242.08 MB \\
            \bottomrule
        \end{tabular}
    }
    \label{tab:memory_runtime}
\end{table}

\begin{table}[t]
    \vspace{-1.5ex}
    \centering
    \caption{\add{\textbf{Runtime and performance} on blur dataset \texttt{\#pub3} and \texttt{\#ArchViz-1}. \ding{55} denotes failed running due to code bugs. Our-NeRF* and Ours-GS* indicate we do not model blur formation process, as other methods.}}
    %}Note that Point-SLAM uses extra memory dynamic radius to improve performance (mark as $^\dagger$).}
    %\footnotesize
    \vspace{-2.0ex}
    \scriptsize
    \setlength{\tabcolsep}{2pt}
    \resizebox{\columnwidth}{!}{
        \begin{tabular}{l|cccc|ccccc}
            \toprule
            \multirow{3}{*}{Method} & \multicolumn{4}{c|}{\texttt{pub3}} & \multicolumn{5}{c}{\texttt{ArchViz-1}}\\
            % \multirow{3}{*}{Method} & & & \texttt{fr1\_desk} & 
            % & & & \texttt{ArchViz-1} & & \\
            %% \cline{2-3} \cline{5-6} 
            % & \multirow{2}{*}[0.5ex]{FPS~$\uparrow$}   & \multirow{2}{*}[0.5ex]{Memory~$\downarrow$} &                    & \multirow{2}{*}[0.5ex]{FPS~$\uparrow$} & \multirow{2}{*}[0.5ex]{Memory~$\downarrow$} \\
            & Tracking & Mapping & System & ATE & Tracking & Mapping & System & ATE & Image \\
            & [ms$\times$it]~$\downarrow$ & [ms$\times$it]~$\downarrow$ & FPS~$\uparrow$ & [cm]$\downarrow$ & [ms$\times$it]~$\downarrow$ & [ms$\times$it]~$\downarrow$ & FPS~$\uparrow$ & [cm]$\downarrow$ & PSNR$\uparrow$ \\
            \midrule
            NICE-SLAM~\cite{zhu2022nice} & 32.72$\times$200 & 49.76$\times$60 & 0.15 & 2.96 & & & \ding{55} & & \\
            CoSLAM~\cite{wang2023co} & 3.89$\times$20 & \fs 15.81$\times$20 & \fs 6.33 & 2.75 & 8.36$\times$50 & 31.00$\times$40 & \snd 2.39 & 5.28 & 22.97\\
            ESLAM~\cite{johari2023eslam} & 16.09$\times$200 & 32.49$\times$60 & 0.31 & \snd 2.44 & 10.32$\times$40 & \snd 36.24$\times$30 & \fs 2.42 & 20.21 & 21.07\\
            PointSLAM~\cite{johari2023eslam} & 29.91$\times$200 & 27.84$\times$150 & 0.17 & 2.56 & 30.99$\times$60 & 67.22$\times$300 & 0.20 & 289.56 & 15.90\\
            Ours-NeRF* & \fs 16.56$\times$1 & \snd 30.07$\times$60 & \snd 2.22 & 2.45 & \fs 10.73$\times$1 & \fs 13.82$\times$50 & 1.99 & \snd 1.77 & \snd 23.51\\
            Ours-NeRF & \snd 39.64$\times$1 & 44.74$\times$60 & 1.49 & \fs 2.08 & \snd 35.91$\times$1 & 36.23$\times$50 & 0.76 & \fs 0.98 & \fs 24.77\\
            \midrule
            SplaTAM~\cite{keetha2024splatam} & 10.59$\times$200 & \fs 12.06$\times$30 & 0.47 & 2.58 & 18.46$\times$60 & 22.34$\times$80 & 0.56 & 36.88 & 17.69\\
            \add{RTG-SLAM}~\cite{peng2024rtgslam} & \add{} & \add{\ding{55}} & \add{} & \add{} & \add{43.92~$\times$~1} & \add{\fs 5.12$\times$100} & \add{\fs 7.04} & \add{\ding{55}} & \add{\ding{55}} \\
            \add{Photo-SLAM}~\cite{Matsuki:Murai:etal:CVPR2024} & \add{36.35$\times$1} & \add{13.80$\times$100} & \add{\fs 3.81} & \add{2.36} & \add{31.32$\times$1} & \add{16.4$\times$100} & \add{\snd 1.13} & \add{4.57} & \add{\snd 23.45} \\
            \add{MonoGS}~\cite{Matsuki:Murai:etal:CVPR2024} & \add{4.39$\times$100} & \add{\snd 64.65$\times$10} & \add{\snd 2.31} & \add{\snd 2.28} & \add{5.33$\times$100} & \add{19.7$\times$100} & \add{1.05} & \add{\snd 1.92} & \add{23.18} \\
            Ours-GS* & \fs 17.15$\times$1 & 29.87$\times$100 & 1.34 & 2.53 & \fs 22.14$\times$1 & \snd 15.29$\times$100 & 0.90 & 4.58 &  22.32\\
            Ours-GS & \snd 19.10$\times$1 & 192.71$\times$160 & 0.13 & \fs 2.16 & \snd 23.91$\times$1 & 54.95$\times$160 & 0.17 & \fs 0.75 & \fs 28.44\\
            \bottomrule
        \end{tabular}
    }
    \vspace{-1em}
    \label{tab:memory_runtime_blur}
\end{table}

\begin{table}[h]
    \centering
    \caption{\add{Ablation studies on the \textbf{sharp datasets} \texttt{\#Room0} and \texttt{\#Office0} compare custom NeRF and 3DGS-based frame-to-map (f2m) tracking with our proposed frame-to-frame (f2f) tracking.}}
    \vspace{-2.5ex}
    % \scriptsize
    \setlength{\tabcolsep}{2pt}
    \resizebox{\columnwidth}{!}{
        \begin{tabular}{l|cccc|cccc|c}
            \toprule
            \multirow{2}{*}{\add{Method}} & \multicolumn{4}{c|}{\add{\texttt{Room0}}} & \multicolumn{4}{c|}{\add{\texttt{Office0}}} & \add{Tracking}\\
            & \add{ATE}$\downarrow$ & \add{PSNR}$\uparrow$ & \add{SSIM}$\uparrow$ & \add{LPIPS}$\downarrow$ & \add{ATE}$\downarrow$ & \add{PSNR}$\uparrow$ & \add{SSIM}$\uparrow$ & \add{LPIPS}$\downarrow$ & \add{FPS}$\uparrow$\\
            \midrule
            \add{COSLAM (f2m)~\cite{wang2023co}} &  \add{0.70} & \add{27.27} & \add{0.910} & \add{0.324} & \add{0.59} & \add{34.14} & \add{0.961} & \add{0.209} & \add{16.64}\\
            \add{COSLAM-ours (f2f)} & \add{\fs 0.32} & \add{\fs 28.86} & \add{\fs 0.931} & \add{\fs 0.272} & \add{\fs 0.34} & \add{\fs 35.45} & \add{\fs 0.969} & \add{\fs 0.171} & \add{\fs 29.06}\\
            \midrule
            \add{ESLAM (f2m)~\cite{johari2023eslam}} & \add{0.71} & \add{25.32} & \add{0.875} & \add{0.313} & \add{0.57}	& \add{33.71} & \add{0.960} & \add{0.184} & \add{18.25}\\
            \add{ESLAM-ours (f2f)} & \add{\fs 0.34} & \add{\fs 28.20} & \add{\fs 0.916} & \add{\fs 0.265} & \add{\fs 0.25} & \add{\fs 35.27} & \add{\fs 0.969} & \add{\fs 0.167} & \add{\fs 29.00}\\
            \midrule
            \add{MonoGS (f2m)~\cite{Matsuki:Murai:etal:CVPR2024}} & \add{0.42} & \add{34.55} & \add{0.969} & \add{0.071} & \add{0.45} & \add{40.06} & \add{0.991} &	\add{0.051} & \add{3.94} \\
            \add{MonoGS-ours (f2f)} & \add{\fs 0.31} & \add{\fs 35.07} & \add{\fs 0.976} & \add{\fs 0.066} & \add{\fs 0.36} & \add{\fs 40.23} & \add{\fs 0.993} & \add{\fs 0.049} & \add{\fs 29.21}\\
            \bottomrule
        \end{tabular}
    }
    \vspace{-1em}
    \label{tab:ablation_tracker_sharp}
\end{table}
\begin{table}[h]
    \centering
    \caption{\add{Ablation studies on the \textbf{blur datasets} \texttt{\#ArchViz-1} and \texttt{\#ArchViz-2} compare custom NeRF and 3DGS-based frame-to-map (f2m) tracking with our frame-to-frame (f2f) tracking.}}
    \vspace{-2.5ex}
    % \scriptsize
    \setlength{\tabcolsep}{2pt}
    \resizebox{\columnwidth}{!}{
        \begin{tabular}{l|cccc|cccc|c}
            \toprule
            \multirow{2}{*}{\add{Method}} & \multicolumn{4}{c|}{\add{\texttt{ArchViz-1}}} & \multicolumn{4}{c|}{\add{\texttt{ArchViz-2}}} & \add{Tracking}\\
            & \add{ATE}$\downarrow$ & \add{PSNR}$\uparrow$ & \add{SSIM}$\uparrow$ & \add{LPIPS}$\downarrow$ & \add{ATE}$\downarrow$ & \add{PSNR}$\uparrow$ & \add{SSIM}$\uparrow$ & \add{LPIPS}$\downarrow$ & \add{FPS}$\uparrow$ \\
            \midrule
            \add{COSLAM (f2m)} &  \add{5.28}  & \add{22.97} & \add{0.841} & \add{0.361} & \add{4.67}  & \add{21.98} & \add{0.685} & \add{0.570} & \add{\snd 2.39}\\
            \add{COSLAM-blur (f2m)} & \add{\snd 0.87}  & \add{\snd 25.08} & \add{\snd 0.904} & \add{\snd 0.292} & \add{\snd 1.09}  & \add{\snd 26.03} & \add{\snd 0.866} & \add{\snd 0.359} & \add{0.61}\\
            \add{COSLAM-blur-ours (f2f)} & \add{\fs 0.61}  & \add{\fs 25.43} & \add{\fs 0.922} & \add{\fs 0.265} & \add{\fs 0.71}  & \add{\fs 26.72} & \add{\fs 0.893} & \add{\fs 0.314} & \add{\fs 29.16}\\
            \midrule
            \add{ESLAM (f2m)} & \add{20.12} & \add{21.07} & \add{0.766} & \add{0.446} & \add{12.61} & \add{23.87} & \add{0.785} & \add{0.475} & \add{\snd 2.42} \\
            \add{ESLAM-blur (f2m)} & \add{\snd 2.46}  & \add{\snd 23.96} & \add{\snd 0.873} & \add{\snd 0.328} & \add{\snd 3.59}  & \add{\snd 25.23} & \add{\snd 0.831} & \add{\snd 0.406} & \add{0.57}\\
            \add{ESLAM-blur-ours (f2f)} & \add{\fs 0.98}  & \add{\fs 24.77} & \add{\fs 0.905} & \add{\fs 0.312} & \add{\fs 1.13}  & \add{\fs 26.35} & \add{\fs 0.875} & \add{\fs 0.372} & \add{\fs 27.85} \\
            \midrule
            \add{MonoGS (f2m)} & \add{1.92}  & \add{23.18} & \add{0.852} & \add{0.337} & \add{2.96}  & \add{27.14} & \add{0.884} & \add{0.329} & \add{\snd 1.88}\\
            \add{MonoGS-blur (f2m)} & \add{\snd 1.06}  & \add{\snd 27.30} & \add{\snd 0.924} & \add{\snd 0.214} & \add{\snd 0.52}  & \add{\snd 29.92} & \add{\snd 0.948} & \add{\snd 0.208} & \add{0.41}\\
            \add{MonoGS-blur-ours (f2f)} & \add{\fs 0.68}  & \add{\fs 28.27} & \add{\fs 0.953} & \add{\fs 0.148} & \add{\fs 0.30}  & \add{\fs 30.51} & \add{\fs 0.960} & \add{\fs 0.166} & \add{\fs 41.23}\\
            \bottomrule
        \end{tabular}
    }
    \vspace{-1.0em}
    \label{tab:ablation_tracker_blur}
\end{table}

\begin{table}[!h]
    % \vspace{-1.5ex}
    \centering
    \caption{The effect of the number of interpolated virtual images on dataset \texttt{\#ArchViz-1}. It demonstrates that the performance saturates as the number increases.}
    \vspace{-1.5ex}
    % \scriptsize
    \setlength{\tabcolsep}{2pt}
    \resizebox{\columnwidth}{!}{
        \begin{tabular}{c|ccccc|ccccc}
            \toprule
            \multirow{3}{*}{$n$} & \multicolumn{5}{c|}{\texttt{ArchViz-1}} & \multicolumn{5}{c}{\texttt{ArchViz-3}}\\
            & ATE & Mapping & \multicolumn{3}{c|}{Image} & ATE & Mapping & \multicolumn{3}{c}{Image} \\
            & [cm]$\downarrow$ & Frame/s$\downarrow$ & PSNR$\uparrow$ & SSIM$\uparrow$ & LPIPS$\downarrow$ & [cm]$\downarrow$ & Frame/s$\downarrow$ & PSNR$\uparrow$ & SSIM$\uparrow$ & LPIPS$\downarrow$ \\
            \midrule
            7 & 1.024 & \fs5.931 & 27.89 & 0.948 & 0.171 & 1.810 & \fs6.512 & 27.45 & 0.933 & 0.188\\
            9 & 0.909 & \snd 6.730 & 27.85 & 0.947 & 0.158 & 1.689 & \snd 7.439 & 27.36 & 0.932 & 0.191\\
            11 & 0.771 & 7.899 & 28.29 & 0.953 & 0.150 & 1.697 & 8.612 & 27.54 & 0.940 & 0.177\\
            13 & 0.749 & 8.793  & \snd 28.45 & \fs 0.956 & \snd0.148 & 1.413 & 9.493 & \snd 27.85 & \snd 0.943 & \snd 0.171\\
            15 & \fs 0.714 & 9.918 & \fs 28.49 & \snd0.955 & \fs0.147 & \snd 1.257 & 10.515 & 27.84 & 0.942 & \fs 0.170\\
            17 & \snd 0.723 & 10.516 & 28.37 & \snd0.955 & 0.149 & \fs 1.223 & 11.355 & \fs 27.96 & \fs0.944 & \snd 0.171\\
            \bottomrule
        \end{tabular}
    }
    \vspace{-1.8em}
    \label{tab:ablation_spline}
\end{table}

\subsection{Efficiency Evaluations}
\label{exp_analysis}
\tabnref{tab:memory_runtime} and \tabnref{tab:memory_runtime_blur} report the runtime of \mbavo2 and state-of-the-art methods on the standard Replica and blurred ArchViz datasets. We also report the scene representation memory usage on Replica. The results in \tabnref{tab:memory_runtime} indicate that our NeRF version SLAM system achieves near real-time speed at 22.57 FPS compared to other NeRF-based SLAMs, while our Gaussian Splatting version achieves 1.97 FPS, which is slower than GS-SLAM~\cite{yan2024gs} but faster than SplaTAM~\cite{keetha2024splatam}. Furthermore, our NeRF and Gaussian Splatting versions achieve the best tracking performance among Radiance Fields based and Gaussian Splatting based SLAMs, respectively.
For runtime on the blurred dataset presented in \tabnref{tab:memory_runtime_blur}, we additionally report Our-NeRF* and Our-GS*, which indicate that we do not model the blur formation process in our tracking and mapping. This results in faster FPS due to reduced mapping time, but it leads to a significant decrease in tracking accuracy compared to modeling the blur process (\ie, Our-NeRF and Our-GS). The presence of motion blurred images complicates accurate camera pose estimation, highlighting the robustness of our approach. Without modeling blur, \mbavo2 achieves comparable or better performance in tracking accuracy and system FPS compared with other methods, except for CoSLAM FPS in the \texttt{pub3} sequence. When modeling the blur process, \mbavo2 demonstrates improved performance in tracking ATE and rendering (\eg, \figref{fig:mbavo_img}, \figref{fig:tum} and \figref{fig:realsense}), albeit with slower system FPS due to the increased mapping time required. Despite this, \mbavo2 consistently provides the capability for real-time tracking.
\subsection{Ablation Study}
\label{exp_ablation}
\PAR{Effect of Our Tracker.}
\add{\tabnref{tab:ablation_tracker_sharp} and \tabnref{tab:ablation_tracker_blur} present the ablation study of integrating our tracker on the Replica \texttt{Room0, office0} and ArchViz \texttt{ArchViz-1, ArchViz-2} datasets, compared with the frame-to-frame tracking methods: hash-grid based CoSLAM, tri-plane-based ESLAM and 3DGS-based MonoGS. The results demonstrate that our fully CUDA-implemented frame-to-frame tracker significantly enhances both tracking performance (in terms of ATE and speed) and mapping quality (e.g., PSNR) on both sharp and blurred datasets, compared with NeRF/3DGS-based frame-to-map tracking methods.}
\add{Explanation of~\tabnref{tab:ablation_tracker_sharp} and \tabnref{tab:ablation_tracker_blur}: CoSLAM, ESLAM, and MonoGS are the original frame-to-map models. In CoSLAM-ours, ESLAM-ours, and MonoGS-ours, we replace the tracking modules with our frame-to-frame approach while retaining the original mapping. For fair comparison, we add a physical motion blur model to both tracking and mapping, resulting in COSLAM-blur, ESLAM-blur, and MonoGS-blur. Correspondingly, COSLAM-blur-ours, ESLAM-blur-ours, and MonoGS-blur-ours are the blur-aware variants that use our frame-to-frame tracking while preserving the blur-aware mapping components.}
%
% ``Model Blur" refers to modeling the blur formation process in both tracking and mapping. The results demonstrate that our tracker significantly enhances tracking performance, thereby improving mapping and rendering performance on both the sharp \texttt{Room0} and blurred \texttt{ArchViz-1} datasets.
%
Furthermore, by integrating our CUDA-implemented tracker, the system can achieve real-time tracking speeds, as shown in \tabnref{tab:memory_runtime} and \tabnref{tab:memory_runtime_blur}.

\PAR{Number of Virtual Images.}
We evaluate the effect of the number of interpolated virtual images (\ie~$n$ in \eqnref{eq_blur_im_formation}) within the exposure time. For this experiment, we select two sequences from the synthetic ArchViz dataset: \texttt{ArchViz-1} and \texttt{ArchViz-3}, representing sequences with low and high levels of motion blur, respectively. The experiments are conducted using our Gaussian Splatting version system with a varying number of interpolated virtual images. The tracking ATE, mapping time, and image quality are reported in \tabnref{tab:ablation_spline}. The results show that as the number of virtual images increases, the tracking and rendering performance (i.e., PSNR, SSIM, LPIPS) tend to saturate, while the mapping time continues to increase. Therefore, after considering the trade-off between system performance and processing speed, we choose to use 13 virtual images for our experiments.

\PAR{\add{NeRF or 3DGS SLAM.}}
\add{Our NeRF-based SLAM achieves substantially faster training and inference by sampling only a subset of image pixels during both tracking and mapping. This efficiency, however, comes at the cost of reduced rendering fidelity. In contrast, 3DGS-based SLAM employs full-image rendering with Gaussian representations, leading to slower training but higher rendering quality and faster test-time rendering. As shown in~\figref{fig:mbavo_img}, \tabnref{tab:psnr_archviz}, and \tabnref{tab:memory_runtime_blur}, NeRF-SLAM is preferable for speed-critical applications, while 3DGS-SLAM is more suitable for tasks requiring high-fidelity reconstruction.}

%\textcolor[RGB]{216, 82, 24}{END:}
%Enumeration of section headings is desirable, but not required. When numbered, please be consistent throughout the article, that is, all headings and all levels of section headings in the article should be enumerated. Primary headings are designated with Roman numerals, secondary with capital letters, tertiary with Arabic numbers; and quaternary with lowercase letters. Reference and Acknowledgment headings are unlike all other section headings in text. They are never enumerated. They are simply primary headings without labels, regardless of whether the other headings in the article are enumerated.

\section{Conclusion}
In this paper, we introduce a novel framework \mbavo2 for robust dense visual RGB-D SLAM, implementing both an implicit Radiance Fields version and an explicit Gaussian Splatting version. With our physical motion blur image formation model, highly CUDA-optimized blur-aware tracker and deblurring mapper, our \mbavo2 can track accurate camera motion trajectories within exposure time and reconstructs a sharp and photo-realistic map given severely-blurred video sequence input. We also propose a real-world motion-blurred SLAM dataset with motion-captured groundtruth camera poses that can be useful to the community. Through extensive experiments, we demonstrate that our method performs state-of-the art on both existing and our real-world datasets.

\section*{Acknowledgments}
This work was supported by National Natural Science Foundation of China (62202389). 

% {\appendix[Proof of the Zonklar Equations]
% Use $\backslash${\tt{appendix}} if you have a single appendix:
% Do not use $\backslash${\tt{section}} anymore after $\backslash${\tt{appendix}}, only $\backslash${\tt{section*}}.
% If you have multiple appendixes use $\backslash${\tt{appendices}} then use $\backslash${\tt{section}} to start each appendix.
% You must declare a $\backslash${\tt{section}} before using any $\backslash${\tt{subsection}} or using $\backslash${\tt{label}} ($\backslash${\tt{appendices}} by itself
%  starts a section numbered zero.)}


\bibliographystyle{IEEEtran}
\bibliography{reference}

% \newpage



\vfill

\end{document}